\documentclass[runningheads]{llncs}
\usepackage[T1]{fontenc}
\usepackage[misc]{ifsym}
\newcommand{\auc}{\text{AUROC$\uparrow$}}
\newcommand{\fpr}{\text{FPR$\downarrow_{95\%}$}}
\usepackage{textcomp}
\usepackage{tikz}
\usepackage{subcaption}
\usepackage{graphicx}
\usepackage{amsmath}
\usepackage{amssymb}
\usepackage{booktabs}
\usepackage{enumitem}
\usepackage{adjustbox}

\usepackage{stackengine}
\usepackage{dsfont}
\def\E{\mathbb{E}} 
\DeclareMathOperator*{\argmax}{arg\,max}
\DeclareMathOperator*{\argmin}{arg\,min}

\usepackage{multirow}
\usepackage[normalem]{ulem}
\useunder{\uline}{\ul}{}
\usepackage[short]{optidef}
\newcommand{\mead}{\textsf{{MEAD}}}

\usepackage[pagebackref,breaklinks,colorlinks]{hyperref}
\usepackage[capitalize]{cleveref}
\crefname{section}{Sec.}{Secs.}
\Crefname{section}{Section}{Sections}
\Crefname{table}{Table}{Tables}
\crefname{table}{Tab.}{Tabs.}
\usepackage{xcolor}

\newcommand{\method}[1]{\textsl{#1}}

\begin{document}
\title{\textsc{MEAD}: A Multi-Armed Approach for Evaluation of Adversarial Examples  Detectors}

\titlerunning{\textsc{MEAD}}
%
\author{Federica Granese (\Letter)\inst{1}\thanks{\footnotesize{These authors contributed equally to this work.}}
\and
Marine Picot\inst{2,3}$^\star$ \and
Marco Romanelli\inst{2} \and
Francisco Messina\inst{5} \and Pablo Piantanida\inst{4}}
\authorrunning{F. Granese et al.}
\institute{Lix, Inria, Institute Polytechnique de Paris,
Sapienza University of Rome 
\email{federica.granese@inria.fr}
\and
 Laboratoire des signaux et systèmes (L2S), \\ Université Paris-Saclay, CNRS, CentraleSupélec, France
\email{\{marine.picot,marco.romanelli,pablo.piantanida\}@centralesupelec.fr}
\and
McGill University, Canada
\and
International Laboratory on Learning Systems (ILLS),\\ 
McGill - ETS - MILA - CNRS - Université Paris-Saclay - CentraleSupélec, Canada
\and
Universidad de Buenos Aires, Argentina\\
\email{fmessina@fi.uba.ar}}
\maketitle              

\tocauthor{Federica~Granese,Marine~Picot,Marco~Romanelli,Francisco~Messina,Pablo~Piantanida}
\toctitle{\textsc{MEAD}: A Multi-Armed Approach for Evaluation of Adversarial Examples  Detectors}

\begin{abstract}
Detection of adversarial examples has been a hot topic in the last years due to its importance for safely deploying machine learning algorithms in critical applications. However, the detection methods are generally validated by assuming a single implicitly known attack strategy, which does not necessarily account for real-life threats. Indeed, this can lead to an overoptimistic assessment of the detectors' performance and may induce some bias in the comparison between competing detection schemes. We propose a novel multi-armed framework, called \mead, for evaluating detectors based on several attack strategies to overcome this limitation. Among them, we make use of three new objectives to generate attacks. The proposed performance metric is based on the worst-case scenario: detection is successful if and only if all different attacks are correctly recognized. Empirically, we show the effectiveness of our approach. Moreover, the poor performance obtained for state-of-the-art detectors opens a new exciting line of research.
\end{abstract}

\section{Introduction}
\addtocounter{footnote}{-5}%
\label{sec:intro}
Despite recent advances in the application of machine learning, the vulnerability of deep learning models to maliciously crafted examples ~\cite{Szegedy2014ICLR} is still an open problem of great interest for safety-critical applications~\cite{Aldahdooh2021REV,DBLP:conf/icml/AthalyeC018,croce2020minimally,DBLP:conf/aaai/ZhengC019}.
Over time, a large body of literature has been produced on the topic of defense methods against adversarial examples. On the one hand, interest in detecting adversarial examples given a pre-trained model is gaining momentum~\cite{NSS,NIC,MagNet}. On the other hand, several techniques have been proposed to train models with improved robustness to future attacks~\cite{madry2018,FARAD,DBLP:conf/cvpr/ZhengSLG16}.
Interestingly, Croce et al. have recently pointed out that, due to the large number of proposed methods, the problem of crafting an objective approach to evaluate the quality of methods to train robust models is not trivial. To this end, they have presented~\textit{RobustBench}~\cite{RobustBench}, a standardized benchmark to assess adversarial robustness. To the best of our knowledge, we claim that an equivalent benchmark does not exist in the case of methods to detect adversarial examples given a pre-trained model.
Therefore, in this work, we provide a general framework to evaluate the performance of adversarial detection methods. Our idea stems from the following key observation. Generally, the performance of current state-of-the-art (SOTA) adversarial examples detection methods is evaluated assuming a unique and thus implicitly known attack strategy, which does not necessarily correspond to real-life threats. We further argue that this type of evaluation has two main flaws: 
the performance of detection methods may be overestimated, and the comparison between detection schemes may be biased.
We propose a two-fold solution to overcome the aforementioned limitations, leading to a 
less biased evaluation of different approaches. This is accomplished by evaluating the detection methods on simultaneous attacks on 
the target classification model using different adversarial strategies, considering the most popular attack techniques in the literature, and incorporating three new attack objectives to extend the generality of the proposed framework. 
Indeed, we argue that additional attack objectives result in new types of adversarial examples that cannot be constructed otherwise. 
In particular, we translate such an evaluation scheme in~\mead.

\mead~is a novel evaluation framework that uses a simple but still effective ``multi-armed'' attack to remove the implicit assumption that detectors know the attacker's strategy. 
More specifically, for each natural sample, we consider the detection to be successful \emph{if and only if} the detector is able to identify all the different attacks perpetrated by perturbing the testing sample at hand.
We deploy the proposed framework to evaluate the performance of SOTA adversarial examples detection methods over multiple benchmarks of visual datasets. Overall, the collected results are consistent throughout the experiments. The main takeaway is that considering a multi-armed evaluation criterion exposes the weakness of SOTA detection methods, yielding, in some cases, relatively poor performances. The proposed framework, although not exhaustive, sheds light on the fact that evaluations so far presented in the literature are highly biased and unrealistic.  Indeed, the same detector achieves very different performances when it is informed about the current attack as opposed to when it is not. 
Not surprisingly, supervised and unsupervised methods achieve comparable performances with the multi-armed framework, meaning that training the detectors knowing a specific attack used at testing time does not generalize to other attacks enough. 
Indeed the goal of~\mead~is not to show that new attacks can always fool robust classifiers but to show that the detectors that may work well when evaluated with a unique attack strategy end up being defeated by new attacks.

\subsection{Summary of contributions}
We propose \mead, a novel multi-armed evaluation framework for adversarial examples detectors involving several attackers to ensure that the detector is not overfitted to a particular attack strategy. The proposed metric is based on the following criterion. Each adversarial sample is correctly detected if and only if all the possible attacks on it are successfully detected. We show that this approach is less biased and yields a more effective metric than the one obtained by assuming only a single attack at evaluation time (see \cref{sec:framework}).

We make use of three new objective functions which, to the best of our knowledge, have never been used for the purpose of generating adversarial examples at testing time. These are \emph{KL divergence}, \emph{Gini Impurity} and \emph{Fisher-Rao distance}. Moreover, we argue that each of them contributes to jointly creating competitive attacks that cannot be created by a single function (see \cref{sec:new_losses}).

We perform an extensive numerical evaluation of SOTA and uncover their limitations, suggesting new research perspectives in this research line (see \cref{sec:experiments}). \\\\
The remaining paper is organized as follows. First, in~\cref{sec:related_work}, we present a detailed overview of the recent related works. In~\cref{sec:adv_pb}, we describe the adversarial problem along with the new objectives we introduce within the proposed evaluation framework, {\mead}, which is further explained in~\cref{sec:framework}. We extensively experimentally validate {\mead} in~\cref{sec:experiments}. Finally, in~\cref{sec:conclusion},  we provide the summary together with concluding remarks.

\section{Related Works} \label{sec:related_work}
State-of-the-art methods to detect adversarial examples can be separated in two main groups~\cite{Aldahdooh2021REV}: supervised and unsupervised methods. In the supervised setting, detectors can make use of the knowledge of the attacker's procedure. The \textit{network invariant model approach} extracts natural and adversarial features from the activation values of the network's layers~\cite{CarraraBCFA18,LuIF17,MetzenGFB17}, while the \textit{statistical approach} extract features using statistical tools (e.g. maximum mean discrepancy~\cite{MMD}, PCA~\cite{PCA}, kernel density estimation~\cite{FeinmanCSG17},  local intrinsic dimensionality~\cite{LID}, model uncertainty~\cite{FeinmanCSG17} or natural scene statistics~\cite{NSS}) to separate in-training and out-of-training data distribution/manifolds. To overcome the intrinsic limitation of the necessity to have prior knowledge of attacks, unsupervised detection methods consider only clean data at training time. The features extraction can rely on different techniques (e.g., \textit{feature squeezing}~\cite{LiangLSLSW21,FS}, \textit{denoiser approach}~\cite{MagNet}, \textit{network invariant}~\cite{NIC}, \textit{auxiliary model}~\cite{SFAD,DNR,ZhengH18}). 
Moreover, detection methods of adversarial examples can also act on the underlying classifier by considering \textit{a novel training procedure} (e.g., reverse cross-entropy~\cite{PangDDZ18}; the rejection option~\cite{SFAD,DNR}) and a thresholding test strategy towards robust detection of adversarial examples. Finally, detection methods can also be impacted by the learning task of the underlying network (e.g., for human recognition tasks~\cite{TaoMLZ18}).

\subsection{Considered detection methods}
\label{sec:review_detectors}
\textbf{Supervised methods.} Supervised methods can make use of the knowledge of how adversarial examples are crafted. They often use statistical properties of either the input samples or the output of hidden layers. \method{NSS}~\cite{NSS} extract the \textit{Natural Scene Statistics} of the natural and adversarial examples, while \method{LID}~\cite{LID} extract the \textit{local intrinsic dimensionality} features of the output of hidden layers for natural, noisy and adversarial inputs. \method{KD-BU}~\cite{FeinmanCSG17} estimates the \textit{kernel density} of the last hidden layer in the feature space, then estimates the \textit{bayesian uncertainty} of the input sample, following the intuition that the adversarial examples lie off the data manifold. 
Once those features are extracted, all methods train a detector to discriminate between natural and adversarial samples.

\textbf{Unsupervised methods.} Unsupervised method can only rely on features of the natural samples. \method{FS}~\cite{FS} is an unsupervised method that uses \textit{feature squeezing} to compare the model’s predictions. Following the idea of estimating the distance between the test examples and the boundary of the manifold of normal examples, \method{MagNet}~\cite{MagNet} comprises detectors based on \textit{reconstruction error} and detectors based on \textit{probability divergence}.

\subsection{Considered attack mechanisms}\label{sec:review_attacks}
\label{recap_attacks}
The attack mechanisms can be divided into two categories: whitebox attacks, where the adversary has complete knowledge about the targeted classifier (its architecture and weights), and blackbox attacks where the adversary has no access to the internals of the target classifier.

\textbf{Whitebox attacks.} One of the first introduced attack mechanisms is what we call the Fast Gradient Sign Method (\textbf{FGSM})~\cite{goodfellow2014}. It relies on computing the direction gradient of a given objective function with respect to (w.r.t.) the input of the targeted classifier and modifying the original sample following it. 
This method has been improved multiple times. Basic Iterative Method (\textbf{BIM})~\cite{kurakin2016adversarial} and Projected Gradient Descent (\textbf{PGD})~\cite{madry2018} are two iteration extensions of \textbf{FGSM}. They were introduced at the same time, and the main difference between the two is that \textbf{BIM} initializes the algorithm to the original sample while \textbf{PGD} initializes it to the original sample plus a random noise. Despite that \textbf{PGD} was introduced under the L$_{\infty}$-norm constraint, it can be extended to any L$_{p}$-norm constraint.
Deepfool (\textbf{DF})~\cite{moosavi2016deepfool} was later introduced. It is an iterative method based on a local linearization of the targeted classifier and the resolution of this simplified problem. Finally, the Carlini\&Wagner method (\textbf{CW}) aims at finding the smaller noise to solve the adversarial problem. To do so, they present a relaxation based on the minimization of specific objectives that can be chosen depending on the attacker's goal. 

\textbf{Blackbox attacks.} Blackbox attacks can only rely on queries to attack specific models. Square Attack (\textbf{SA})~\cite{andriushchenko2020square} is an iterative method that randomly searches for a perturbation that will increase the attacker's objective at each step, Hop Skip Jump (\textbf{HOP})~\cite{chen2020hopskipjumpattack} tries to estimate the gradient direction to perturb, and Spatial Transformation Attack (\textbf{STA})~\cite{engstrom2019exploring} applies small translations and rotations to the original sample to fool the targeted classifier.  


\section{Adversarial Examples and Novel Objectives}
\label{sec:adv_pb}
Let $\mathcal{X}\subseteq\mathbb{R}^d$ be the input space and let $\mathcal{Y}=\{1,\dots, C\}$ be the label space related to some task of interest. We denote by $P_{XY}$ 
the unknown data distribution over $\mathcal{X}\times\mathcal{Y}$. 
Throughout the paper we refer to the classifier  
$q_{\widehat{Y}|X}(y|\mathbf{x};\theta)$ 
to be the parametric soft-probability model, where $\theta\in\Theta$ are the parameters, $y\in\mathcal{Y}$ the label and $f_{\theta}:\mathcal{X}\rightarrow\mathcal{Y}$ 
s.t.
$f_{\theta}(\mathbf{x})= \arg\max_{y\in\mathcal{Y}} q_{\widehat{Y}|X}(y|\mathbf{x};\theta)$ to be its induced hard decision. Finally, we denote by $\mathbf{x}^{\prime}\in\mathbb{R}^{d}$ an adversarial example, by $\ell(\mathbf{x}, \mathbf{x}^{\prime}; \theta)$ the  objective function used by the attacker to generate that sample, and $a_{\ell}(\cdot; \varepsilon, p)$ the attack mechanism according to a objective function $\ell$, with $\varepsilon$ the maximal perturbation allow and $p$ the L$_{p}$-norm constraint.

\subsection{Generating adversarial examples}
Adversarial examples are slightly modified inputs that can fool a target classifier. Concretely, Szegedy {\it et al.}~\cite{szegedy2013} define the adversarial generation problem as: 
\begin{equation}
   \mathbf{x}^{\prime} =\underset{\mathbf{x}^{\prime}\in \mathbb{R}^{d}\, :\, \lVert\mathbf{x}^{\prime} - \mathbf{x}\rVert_{p}<\varepsilon}{\argmin}\lVert\mathbf{x}^{\prime} - \mathbf{x}\rVert\text{~~s.t.~~}f_{\theta}(\mathbf{x}^{\prime}) \neq y,
   \label{eq:adversarial_problem}
\end{equation}
where $y$ is the true label (supervision) associated to the sample $\mathbf{x}$.
Since this problem is difficult to tackle, it is commonly relaxed as follows~\cite{carlini2017towards}: 
\begin{equation}
   a_{\ell}(\mathbf{x}; \varepsilon, p)\footnote{Throughout the paper, when the values of $\varepsilon$ and $p$ are clear from the context, we denote the attack mechanism as $a_\ell(\cdot)$.} \equiv~\mathbf{x}^\prime_\ell  =\underset{\mathbf{x}^\prime_\ell\in \mathbb{R}^{d} \,:\,\lVert\mathbf{x}^\prime_\ell - \mathbf{x}\rVert_{p}<\varepsilon
   }{\argmax}\ell(\mathbf{x},\mathbf{x}^\prime_\ell; \mathbf{\theta).}
   \label{eq:relaxed_adv_problem}
\end{equation}
It is worth to emphasize that the choice of the objective $\ell(\mathbf{x}, \mathbf{x}^\prime_\ell; \theta)$ plays a crucial role in generating powerful adversarial examples $\mathbf{x}^\prime_\ell$. 
The objective function $\ell$ traditionally used is the Adversarial Cross-Entropy (ACE)~\cite{madry2018}:
\begin{equation}
\label{eq:ACE}
    \ell_{\text{ACE}}(\mathbf{x},\mathbf{x}^\prime_\ell; \theta) =  \E_{Y|\mathbf{x} }\big[ - \log q_{\widehat{Y}|X}(Y|\mathbf{x}^\prime_\ell;\theta) \big],
\end{equation}
It is possible to use any objective function $\ell$ to craft adversarial samples. We present the three losses that we use to generate adversarial examples in the following. While these losses have already been considered in detection/robustness cases, to the best of our knowledge, they have never been used to craft attacks to test the performances of detection methods.
\subsection{Three New Objective Functions }\label{sec:new_losses}
\subsubsection{The Kullback-Leibler divergence.} The Kullback-Leibler (KL) divergence between the natural and the adversarial probability distributions has been widely used in different learning problems, as building training losses for robust models~\cite{zhang2019theoretically}. KL is defined as follows:
\begin{align}\label{KL}
    \ell_{\text{KL}}\left(\mathbf{x}, \mathbf{x}^\prime_\ell; \theta \right)=\mathbb{E}_{\widehat{Y}|
    \mathbf{x};\theta}\left[\log\left(\frac{q_{\widehat{Y}|
    X }(\widehat{Y}|\mathbf{x};\theta)}{q_{\widehat{Y}|
    X }(\widehat{Y}|\mathbf{x}^\prime_\ell;\theta)}\right)\right].
\end{align}
\subsubsection{The Fisher-Rao objective.} The Fisher-Rao  (FR) distance is an information-geometric measure of dissimilarity between soft-predictions~ \cite{atkinson1981rao}. It has been recently used to craft a new regularizer for robust classifiers~\cite{FARAD}. FR
can be computed as follows:
\begin{align}  
\ell_{\text{FR}}(\mathbf{x}, \mathbf{x}^\prime_\ell; \theta) =
&2 \arccos \left( \sum_{y \in \mathcal{Y}} \sqrt{q_{\widehat{Y}|
    X }(y|\mathbf{x};\theta) q_{\widehat{Y}|
    X }(y|\mathbf{x}^\prime_\ell;\theta)} \right).
\label{eq:fisher_rao_multiclass}
\end{align}
\subsubsection{The Gini Impurity score.}
\begin{figure}[!t]
	\centering
	\begin{subfigure}[b]{0.33\textwidth}
	    \centering
	    \includegraphics[width=\textwidth]{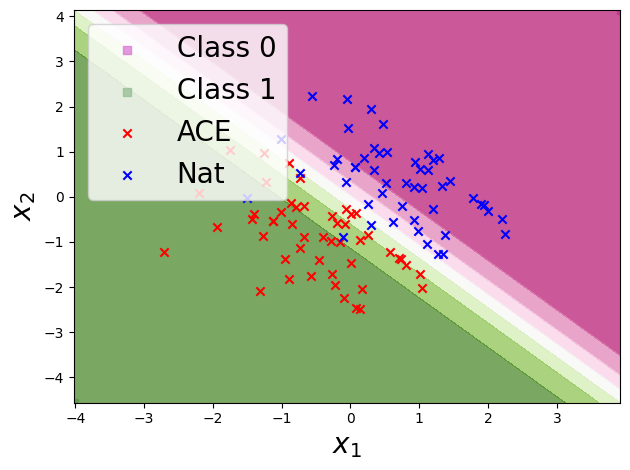}
	    \vspace{-1.5\baselineskip}
	    \caption{Pre-trained classifier}
	    \label{fig:toy_classifier_decisions_ace}
	\end{subfigure}
	\begin{subfigure}[b]{0.33\textwidth}
		\centering
		\includegraphics[width=\textwidth]{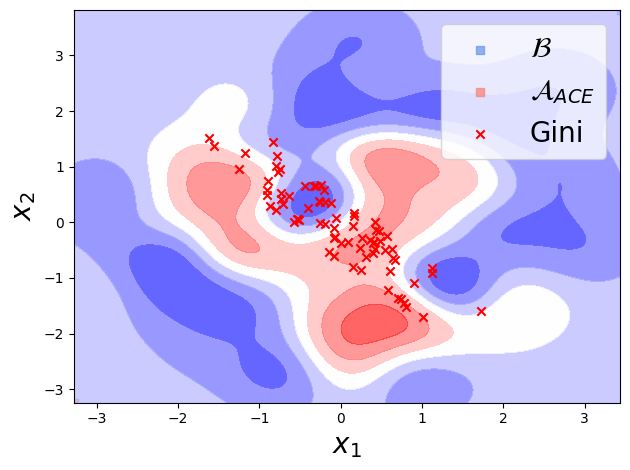}
		\vspace{-1.5\baselineskip}
		\caption{Detector trained on ACE}
		\label{fig:toy_detectors_decisions_ace}
	\end{subfigure}
	\begin{subfigure}[b]{0.33\textwidth}
	    \centering
	    \includegraphics[width=\textwidth]{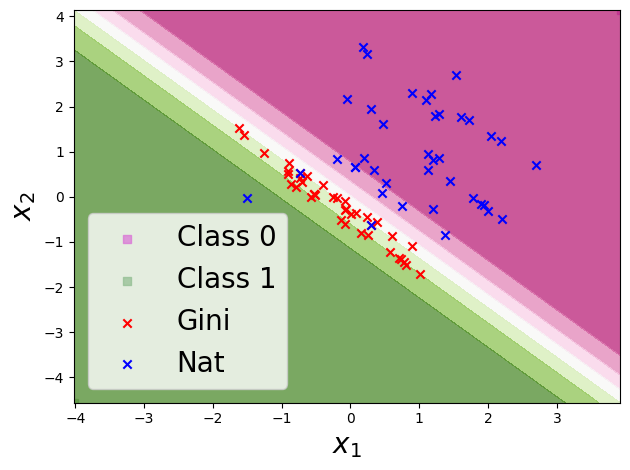}
	    \vspace{-1.5\baselineskip}
	    \caption{Pre-trained classifier}
	    \label{fig:toy_classifier_decisions_g}
	\end{subfigure}
	\begin{subfigure}[b]{0.33\textwidth}
		\centering
		\includegraphics[width=\textwidth]{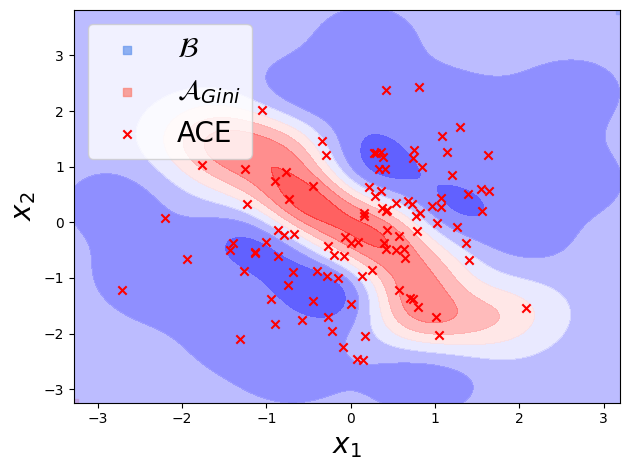}
		\vspace{-1.5\baselineskip}
		\caption{Detector trained on Gini}
		\label{fig:toy_detector_decisions_g}
	\end{subfigure}

	\caption{Decision boundary for the binary classifier \ref{fig:toy_classifier_decisions_ace}-\ref{fig:toy_classifier_decisions_g}: the decision region for class 1 is green, the decision region of class 0 is pink. The natural testing samples belonging to class 0 are reported in blue, the corresponding adversarial examples crafted using ACE (\ref{fig:toy_classifier_decisions_ace}) and Gini Impurity (\ref{fig:toy_classifier_decisions_g}) in red. Decision boundary of the detectors \ref{fig:toy_detectors_decisions_ace}-\ref{fig:toy_detector_decisions_g}: $\mathcal{B}$, the decision region of the natural examples; $\mathcal{A}_{\ell}$, reported in red shades, the decision region of the adversarial examples when the detector is trained on data points crafted via $\ell\in\{\text{ACE},\text{~Gini}\}$ as objective. The darker shades stand for higher confidence. The red points represent the adversarial examples created with the opposite loss (respectively $\ell\in\{\text{Gini},\text{~ACE}\}$).} 
	\label{fig:toy_example_points}
\end{figure}
The Gini Impurity score approximates  the probability of incorrectly classifying the input $\mathbf{x}$ if it was randomly labeled according to the model's output distribution $q_{\widehat{Y}|
    X }(y|\mathbf{x}^\prime_\ell;\theta)$. It was recently used in~\cite{DOCTOR} to determine whether a sample is correctly or incorrectly classified.
\begin{align}
    \label{g_loss}
     \ell_{\text{Gini}}(\cdot, \mathbf{x}^\prime_\ell; \theta) &
    = 1 - \sqrt{\underset{y\in\mathcal{Y}}{\sum}q^2_{\widehat{Y}|
    X }(y|\mathbf{x}^\prime_\ell;\theta)}.
\end{align}

\subsection{A case study: ACE vs. Gini Impurity}
\label{sec:toy_example}
In~\Cref{fig:toy_example_points} we provide insights on why we need to evaluate the detectors on attacks crafted through different objectives. We create a synthetic dataset that consists of 300 data points drawn from $\mathcal{N}_0 = \mathcal{N}(\mu_0, \sigma^2 \mathbf{I})$ 
and 300 data points drawn from $\mathcal{N}_1 = \mathcal{N}(\mu_1, \sigma^2 \mathbf{I})$, where $\mu_0 = [1~1]$, $\mu_1 = [-1~-1]$ and $\sigma = 1$. To each data point $\mathbf{x}$ is assigned true label $0$ or $1$ depending on whether $\mathbf{x}\sim\mathcal{N}_0$ or $\mathbf{x}\sim\mathcal{N}_1$, respectively. The data points have been split between the training set ($70\%$) and the testing set ($30\%$). We finally train a simple binary classifier with one single hidden layer and a learning rate of 0.01 for 20 epochs. We attack the classifier by generating adversarial examples with PGD under the L$_{\infty}$-norm constraints with $\varepsilon=1.2$ for the ACE attacks and $\varepsilon=5$ for the Gini Impurity attacks to have a classification accuracy (classifier performance) of $50\%$ on the corrupted data points. In~\Cref{fig:toy_classifier_decisions_ace}-\ref{fig:toy_classifier_decisions_g} we plot the decision boundary of the binary classifier together with the adversarial and natural examples belonging to class 0. As can be seen, ACE creates points that lie in the opposite decision region with respect to the original points (\cref{fig:toy_classifier_decisions_ace}). Conversely, Gini Impurity tends to create new data points in the region of maximal uncertainty of the classifier (\cref{fig:toy_classifier_decisions_g}). 
Consider the scenario where we train a simple Radial Basis Function (RBF) kernel SVM on a subset of the testing set of the natural points together with the attacked examples, generated with the ACE or the Gini Impurity score depending on the case (\cref{fig:toy_detectors_decisions_ace}-\ref{fig:toy_detector_decisions_g}). We then test the detector on the data points originated with the opposite loss,~\Cref{fig:toy_detectors_decisions_ace} and~\Cref{fig:toy_detector_decisions_g} respectively. The decision region of the detector for natural examples is in blue, and the one for the adversarial examples is in red. The intensity of the color corresponds to the level of certainty of the detector. The accuracy of the detector  on natural and adversarial data points decreases from $71\%$ to $62\%$ when changing to the opposite loss in~\cref{fig:toy_detectors_decisions_ace}, and from $87\%$ to $63\%$ in~~\cref{fig:toy_detector_decisions_g}. 
Hence, testing on samples crafted using a different loss in~\cref{eq:relaxed_adv_problem} means changing the attack and, consequently, evaluating detectors without taking into consideration this possibility leads to a more biased and unrealistic estimation of their performance. When the detector is trained on the adversarial examples created with both the losses, the accuracy is $79.8\%$ when testing on Gini and $66.3\%$ when testing on ACE, which is a better trade-off in adversarial detection performances.\\\\
The aforementioned losses will be included in the following section to design~\mead, our \textit{multi-armed evaluation framework}, a new method to evaluate the performance of adversarial detection with low bias.


\section{Evaluation with a Multi-Armed Attacker }\label{sec:framework}
\input{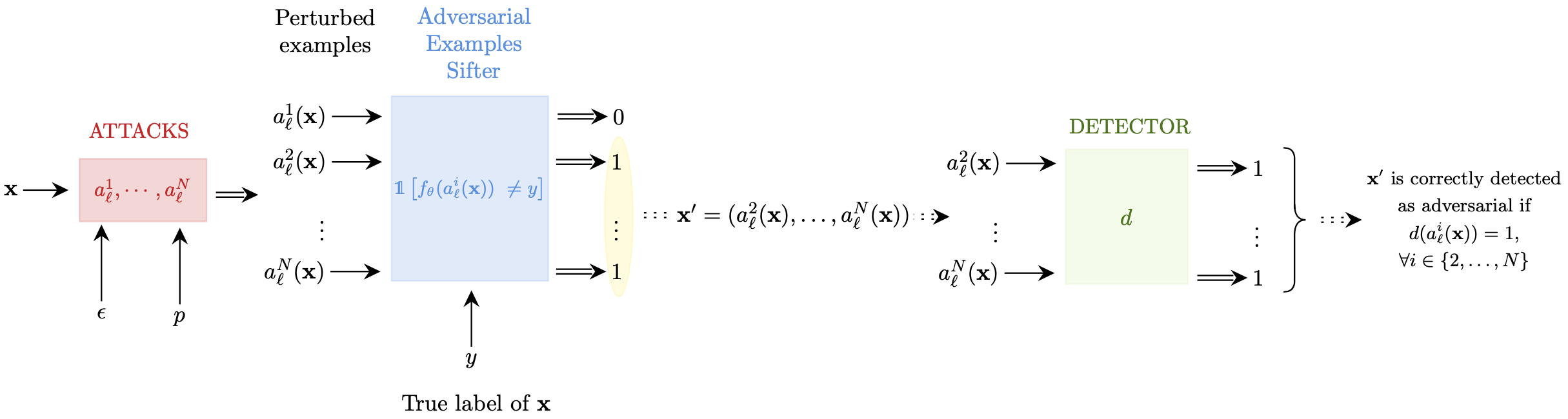}
The proposed evaluation framework,~\mead,
consists in testing an adversarial examples detection method on a large collection of attacks grouped w.r.t. the L$_p$-norm and the maximal perturbation $\varepsilon$ they consider. Each given natural input example is perturbed according to the collection of attacks. 
Note that, for every attack, a perturbed example is considered adversarial \textit{if and only if} it fools the classifier. Otherwise,  it is discarded and will not influence the evaluation. We then feed all the natural and successful adversarial examples to the detector and gather all the predictions. Finally, based on the detection decisions, we evaluate the detector according to a worst-case scenario:\\
\textit{i) Adversarial decision}: for each natural example, we gather all the successful adversarial examples. If the detector detects \textit{all} of them, then the perturbed sample is considered \textit{correctly detected} (i.e., it is a true positive). However, if the detector misses \text{at least one} of them, the noisy sample is considered \textit{undetected} (i.e., it is a false negative). \\
\textit{ii) Natural decision}: for each natural sample, if the detector does not detect it, then the sample is considered \textit{correctly non-detected} (i.e., it is a true negative); otherwise it is \textit{incorrectly detected} (i.e., it is a false positive).

Specifically, let $\mathcal{D}_m = \{ (\mathbf{x}_i,y_i)\}_{i=1}^m \sim P_{XY}$ be the testing set of size $m$, where $\mathbf{x}_i \in \mathcal{X}$ is the natural input sample and $y_i \in \mathcal{Y}$ is its true label. Let
$d$: $\mathcal{X} \times \mathbb{R} \rightarrow \{0,1\}$ be the detection mechanism and $a_\ell: \mathcal{X} \times \mathbb{R} \times\{1, 2, \infty\}\rightarrow\mathcal{X}$ the attack strategy according to the objective function $\ell\in\mathcal{L}$ within a selected collection of objectives $\mathcal{L}$ as described in~\cref{sec:adv_pb}.
For every considered L$_p$-norm, $p\in\{1, 2, \infty \}$, maximal perturbation $\varepsilon \in \mathbb{R}$, and  every threshold $\gamma \in \mathbb{R}$\footnote{With an abuse of notation, $\forall\ell\in\mathcal{L}$ stands for all the considered attack mechanisms for specific values of $\varepsilon$, $p$ within a collection of objectives $\mathcal{L}$.}:
\begin{align}
\label{def:tp}
    &TP_{\varepsilon, p}(\gamma) = \Big\{ (\mathbf{x},y) \in \mathcal{D}_m :  \forall\ell\in\mathcal{L} \;    
    \{f_{\theta}\big(a_\ell(\mathbf{x})\big) \neq y \} ~\wedge~ \{d\big( a_\ell(\mathbf{x}), \gamma\big) = 1\} \Big\}  \\
   & FN_{\varepsilon, p}(\gamma) = \Big\{ (\mathbf{x},y) \in \mathcal{D}_m :  \exists \ell\in\mathcal{L}  \;  \{f_{\theta} \big(a_\ell(\mathbf{x})\big) \neq y \} ~\wedge~ \{d\big( a_\ell(\mathbf{x}), \gamma\big) = 0\} \Big\}\\
  &  TN_{\varepsilon, p}(\gamma) = \{(\mathbf{x},y) \in \mathcal{D}_m : d(\mathbf{x},\gamma) = 0 \}\\
   & FP_{\varepsilon, p}(\gamma) = \{(\mathbf{x},y) \in \mathcal{D}_m : d(\mathbf{x},\gamma) = 1 \}.
\end{align}
In~\cref{fig:framework} we provide a graphical interpretation of {\mead} when the perturbation magnitude and the norm are fixed. 

\section{Experiments}\label{sec:experiments}
In this section, we assess the effectiveness of the proposed evaluation framework, \mead.
The code is available at \url{https://github.com/meadsubmission/MEAD}. 
\begin{table*}[t]
\centering
\resizebox{\columnwidth}{!}{%
\begin{tabular}{c||cc||cc||cc||cc||cc}
\toprule 
\textbf{CIFAR10} &
  \multicolumn{2}{c||}{\mead} & \multicolumn{2}{c||}{ACE}& \multicolumn{2}{c||}{KL}& \multicolumn{2}{c||}{Gini} &
  \multicolumn{2}{c}{FR}
   \\ \hline\midrule
   
\textbf{\method{NSS}} &
  \multicolumn{1}{|c|}{\auc\%} &
   {\fpr\% } &
  \multicolumn{1}{c|}{\auc\%} &
   \fpr\% 
   &
  \multicolumn{1}{c|}{\auc\%} &
  \fpr\% &
  \multicolumn{1}{c|}{\auc\%} &
   \fpr\% &
  \multicolumn{1}{c|}{\auc\%} &
   \fpr\% \\ \midrule
L$_1$ Average & 
    \multicolumn{1}{|c|}{\textbf{62.9}} & \textbf{81.6} &
    \multicolumn{1}{c|}{67.4} & 75.7 &
    \multicolumn{1}{c|}{\underline{67.1}} & 76.0 &
    \multicolumn{1}{c|}{67.8} & \underline{78.2} &
    \multicolumn{1}{c|}{67.6} & 75.6
    \\ 
    L$_2$ Average & 
    \multicolumn{1}{|c|}{\textbf{64.0}} & \textbf{82.0} &
    \multicolumn{1}{c|}{68.7} & 71.0 &
    \multicolumn{1}{c|}{68.5} & 70.9 &
    \multicolumn{1}{c|}{\underline{65.1}} & \underline{82.0} &
    \multicolumn{1}{c|}{68.6} & 71.1
    \\ 
    L$_\infty$ Average & 
    \multicolumn{1}{|c|}{\textbf{71.9}} & \textbf{62.0} &
    \multicolumn{1}{c|}{76.9} & 40.1 &
    \multicolumn{1}{c|}{77.2} & 39.5 &
    \multicolumn{1}{c|}{\underline{73.7}} & \underline{59.6} &
    \multicolumn{1}{c|}{74.1} & 57.2
    \\ 
No norm & 
   \multicolumn{1}{|c|}{\textbf{88.5}} & \textbf{38.8} &
    \multicolumn{1}{c|}{\underline{\textbf{88.5}}} & \underline{\textbf{38.8}} &
    \multicolumn{1}{c|}{\underline{\textbf{88.5}}} & \underline{\textbf{38.8}} &
    \multicolumn{1}{c|}{\underline{\textbf{88.5}}} & \underline{\textbf{38.8}} &
    \multicolumn{1}{c|}{\underline{\textbf{88.5}}} & \underline{\textbf{38.8}}
   \\ \hline\midrule

\textbf{\method{KD-BU}} &
  \multicolumn{1}{|c|}{\auc\%} &
   {\fpr\% } &
  \multicolumn{1}{c|}{\auc\%} &
   \fpr\% 
   &
  \multicolumn{1}{c|}{\auc\%} &
  \fpr\% &
  \multicolumn{1}{c|}{\auc\%} &
   \fpr\% &
  \multicolumn{1}{c|}{\auc\%} &
   \fpr\% \\ \midrule
L$_{1}$ Average & \multicolumn{1}{|c|}{\textbf{50.9}} & \textbf{95.7} &
    \multicolumn{1}{|c|}{70.0} & 88.6 &
    \multicolumn{1}{|c|}{70.0} & 88.4 &
    \multicolumn{1}{|c|}{74.3} & \underline{92.3} &
    \multicolumn{1}{|c|}{\underline{69.8}} & 88.4
    \\ 
 L$_{2}$ Average & 
    \multicolumn{1}{|c|}{\textbf{59.0}} & \textbf{94.1} &
    \multicolumn{1}{|c|}{71.6} & 71.9 &
    \multicolumn{1}{|c|}{71.7} & 71.6 &
    \multicolumn{1}{|c|}{\underline{70.6}} & \underline{92.8} &
    \multicolumn{1}{|c|}{71.7} & 71.8  \\ 
 L$_{\infty}$ Average & 
 \multicolumn{1}{|c|}{\textbf{36.8}} & \textbf{96.9} &
    \multicolumn{1}{|c|}{64.8} & 92.1 &
    \multicolumn{1}{|c|}{68.1} & 91.3 &
    \multicolumn{1}{|c|}{\underline{53.7}} & \underline{95.6} &
    \multicolumn{1}{|c|}{67.8} & 91.7
    \\ 
No norm & 
\multicolumn{1}{|c|}{\textbf{65.4}} & \textbf{94.2} &
\multicolumn{1}{|c|}{\textbf{\underline{65.4}}} & \textbf{\underline{94.2}} &
\multicolumn{1}{|c|}{\textbf{\underline{65.4}}} & \textbf{\underline{94.2}} & \multicolumn{1}{|c|}{\textbf{\underline{65.4}}} & \textbf{\underline{94.2}} &
\multicolumn{1}{|c|}{\textbf{\underline{65.4}}} & \textbf{\underline{94.2}}
   \\ \hline\midrule

\textbf{\method{LID}} &
  \multicolumn{1}{|c|}{\auc\%} &
   {\fpr\% } &
  \multicolumn{1}{c|}{\auc\%} &
   \fpr\% 
   &
  \multicolumn{1}{c|}{\auc\%} &
  \fpr\% &
  \multicolumn{1}{c|}{\auc\%} &
   \fpr\% &
  \multicolumn{1}{c|}{\auc\%} &
   \fpr\% \\ \midrule
L$_{1}$ Average & 
    \multicolumn{1}{|c|}{\textbf{50.8}} & \textbf{95.4} &
    \multicolumn{1}{|c|}{69.6} & 82.1 &
    \multicolumn{1}{|c|}{69.4} & 82.9 &
    \multicolumn{1}{|c|}{88.9} & 49.9 &
    \multicolumn{1}{|c|}{\underline{69.1}} & \underline{83.7}
    \\ 
L$_{2}$ Average & 
    \multicolumn{1}{|c|}{\textbf{63.5}} & \textbf{83.1} &
    \multicolumn{1}{|c|}{73.7} & 70.1 &
    \multicolumn{1}{|c|}{73.4} & 70.7 &
    \multicolumn{1}{|c|}{82.5} & 61.3 &
    \multicolumn{1}{|c|}{\underline{73.2}} & \underline{71.3}\\ 
 L$_{\infty}$ Average & \multicolumn{1}{|c|}{\textbf{53.8}} & \textbf{90.8} &
    \multicolumn{1}{|c|}{75.7} & 56.8 &
    \multicolumn{1}{|c|}{79.9} & 57.6 &
    \multicolumn{1}{|c|}{\underline{71.3}} & \underline{79.7} &
    \multicolumn{1}{|c|}{82.0} & 51.4
    \\ 
No norm & 
\multicolumn{1}{|c|}{\textbf{88.0}} & \textbf{58.1} & 
   \multicolumn{1}{|c|}{\underline{\textbf{88.0}}} & \underline{\textbf{58.1}} & 
   \multicolumn{1}{|c|}{\underline{\textbf{88.0}}} &  \underline{\textbf{58.1}} & 
   \multicolumn{1}{|c|}{\underline{\textbf{88.0}}} & \underline{\textbf{58.1}}  &
    \multicolumn{1}{|c|}{\underline{\textbf{88.0}}} &\underline{\textbf{58.1}}
   \\ \hline\midrule

\textbf{\method{FS}} &
  \multicolumn{1}{|c|}{\auc\%} &
   {\fpr\% } &
  \multicolumn{1}{c|}{\auc\%} &
   \fpr\% 
   &
  \multicolumn{1}{c|}{\auc\%} &
  \fpr\% &
  \multicolumn{1}{c|}{\auc\%} &
   \fpr\% &
  \multicolumn{1}{c|}{\auc\%} &
   \fpr\% \\ \midrule
    L$_{1}$ Average & \multicolumn{1}{|c|}{75.4} & 64.8 &
    \multicolumn{1}{|c|}{92.8} & 25.1 &
    \multicolumn{1}{|c|}{92.9} & 24.9 &
    \multicolumn{1}{|c|}{\textbf{\underline{73.5}}} & \textbf{\underline{67.6}} &
    \multicolumn{1}{|c|}{92.9} & 24.6
    \\ 
    L$_{2}$ Average & 
    \multicolumn{1}{|c|}{74.9} & 65.8&
    \multicolumn{1}{|c|}{87.4} & 31.2 &
    \multicolumn{1}{|c|}{87.6} & 36.9 &
    \multicolumn{1}{|c|}{\textbf{\underline{73.7}}} & \textbf{\underline{67.2}} &
    \multicolumn{1}{|c|}{87.4} & 37.5  
    \\ 
    L$_{\infty}$ Average & \multicolumn{1}{|c|}{\textbf{52.7}} & \textbf{81.1} &
    \multicolumn{1}{|c|}{73.0} & 60.1 &
    \multicolumn{1}{|c|}{77.5} & 55.7 &
    \multicolumn{1}{|c|}{\underline{58.2}} & \underline{78.8} &
    \multicolumn{1}{|c|}{75.7} & 58.5
    \\ 
    No norm &
    \multicolumn{1}{|c|}{\textbf{62.7}} & \textbf{82.5} &
    \multicolumn{1}{|c|}{\underline{\textbf{62.7}}} & \underline{\textbf{82.5}} &
    \multicolumn{1}{|c|}{\underline{\textbf{62.7}}} & \underline{\textbf{82.5}} &
    \multicolumn{1}{|c|}{\underline{\textbf{62.7}}} & \underline{\textbf{82.5}} &
    \multicolumn{1}{|c|}{\underline{\textbf{62.7}}} & \underline{\textbf{82.5}}
       \\ \hline\midrule

\textbf{\method{MagNet}} &
  \multicolumn{1}{|c|}{\auc\%} &
   {\fpr\% } &
  \multicolumn{1}{c|}{\auc\%} &
   \fpr\% 
   &
  \multicolumn{1}{c|}{\auc\%} &
  \fpr\% &
  \multicolumn{1}{c|}{\auc\%} &
   \fpr\% &
  \multicolumn{1}{c|}{\auc\%} &
   \fpr\% \\ \midrule
L$_{1}$ Average & \multicolumn{1}{|c|}{49.6} & 93.7 &
    \multicolumn{1}{|c|}{49.8} & 93.5 &
    \multicolumn{1}{|c|}{49.7} & 93.3 &
    \multicolumn{1}{|c|}{50.1} & 93.2 &
    \multicolumn{1}{|c|}{\textbf{\underline{49.1}}} & \textbf{\underline{93.8}}
    \\ 
L$_{2}$ Average & \multicolumn{1}{|c|}{50.9} & 93.1  &
    \multicolumn{1}{|c|}{52.3} & 89.6 &
    \multicolumn{1}{|c|}{52.3} & 89.4 &
    \multicolumn{1}{|c|}{\textbf{\underline{50.5}}} & \textbf{\underline{93.3}} &
    \multicolumn{1}{|c|}{51.8} & 91.4 \\ 
L$_{\infty}$ Average & 
    \multicolumn{1}{|c|}{\textbf{78.0}} & \textbf{46.1} &
    \multicolumn{1}{|c|}{\underline{79.2}} & \underline{44.6} &
    \multicolumn{1}{|c|}{80.2} & 44.1 &
    \multicolumn{1}{|c|}{\underline{79.2}} & \underline{44.6} &
    \multicolumn{1}{|c|}{80.0} & \underline{44.6}
    \\ 
 
No norm & 
\multicolumn{1}{|c|}{\textbf{79.9}} & \textbf{45.7} &
\multicolumn{1}{|c|}{\textbf{\underline{79.9}}} &\textbf{ \underline{45.7} }&
\multicolumn{1}{|c|}{\textbf{\underline{79.9}}} &\textbf{ \underline{45.7} }&
\multicolumn{1}{|c|}{\textbf{\underline{79.9}}} &\textbf{ \underline{45.7} }&
\multicolumn{1}{|c|}{\textbf{\underline{79.9}}} &\textbf{ \underline{45.7} }
   \\ \hline

   \bottomrule
\end{tabular}
}
\caption{Overall performances on CIFAR10 of all the detectors per objective and in \mead. The worst results among all the settings 
is in \textbf{bold}; the ones in the single-armed setting is \underline{underlined}. No norm denotes the group of attacks that do not depend on the norm constraint.}
\label{tab:cifar_overall}
\end{table*}\begin{table*}[!htb]
\centering
\resizebox{\columnwidth}{!}{%
\begin{tabular}{c||cc||cc||cc||cc||cc}
\toprule 
\textbf{MNIST} &
  \multicolumn{2}{c||}{\mead} & \multicolumn{2}{c||}{ACE}& \multicolumn{2}{c||}{KL}& \multicolumn{2}{c||}{Gini} &
  \multicolumn{2}{c}{FR}
   \\ \hline\midrule
   
\textbf{\method{NSS}} &
  \multicolumn{1}{|c|}{\auc\%} &
   {\fpr\% } &
  \multicolumn{1}{c|}{\auc\%} &
   \fpr\% 
   &
  \multicolumn{1}{c|}{\auc\%} &
  \fpr\% &
  \multicolumn{1}{c|}{\auc\%} &
   \fpr\% &
  \multicolumn{1}{c|}{\auc\%} &
   \fpr\% \\ \midrule
L$_1$ Average & 
    \multicolumn{1}{|c|}{\textbf{96.8}} & \textbf{9.4} & 
    \multicolumn{1}{c|}{\underline{97.0}} & 8.2 &
    \multicolumn{1}{c|}{97.1} & \underline{8.6} &
    \multicolumn{1}{c|}{97.4} & 7.0 &
    \multicolumn{1}{c|}{97.1} & 8.1
    \\ 
    L$_2$ Average & 
    \multicolumn{1}{|c|}{\textbf{90.3}} & \textbf{26.5} &
    \multicolumn{1}{c|}{90.7} & 25.8 &
    \multicolumn{1}{c|}{90.8} & 25.4 &
    \multicolumn{1}{c|}{91.4} & 23.7 &
    \multicolumn{1}{c|}{\underline{90.6}} & \underline{\textbf{26.5}}
    \\ 
    L$_{\infty}$ Average & 
\multicolumn{1}{|c|}{\textbf{88.7}} & 23.5 &
    \multicolumn{1}{c|}{\underline{89.5}} & 23.5 &
    \multicolumn{1}{c|}{\underline{89.5}} & \underline{\textbf{23.6}} &
    \multicolumn{1}{c|}{90.0} & \underline{\textbf{23.6}} &
    \multicolumn{1}{c|}{89.8} & 23.5
    \\ 
No norm & 
    \multicolumn{1}{|c|}{\textbf{87.1}} & \textbf{57.8} &
    \multicolumn{1}{|c|}{\underline{\textbf{87.1}}} & \underline{\textbf{57.8}} &
    \multicolumn{1}{|c|}{\underline{\textbf{87.1}}} & \underline{\textbf{57.8}} &
    \multicolumn{1}{|c|}{\underline{\textbf{87.1}}} & \underline{\textbf{57.8}} &
    \multicolumn{1}{|c|}{\underline{\textbf{87.1}}} & \underline{\textbf{57.8}} 
   \\ \hline\midrule

\textbf{\method{KD-BU}} &
  \multicolumn{1}{|c|}{\auc\%} &
   {\fpr\% } &
  \multicolumn{1}{c|}{\auc\%} &
   \fpr\% 
   &
  \multicolumn{1}{c|}{\auc\%} &
  \fpr\% &
  \multicolumn{1}{c|}{\auc\%} &
   \fpr\% &
  \multicolumn{1}{c|}{\auc\%} &
   \fpr\% \\ \midrule
L$_{1}$ Average & 
    \multicolumn{1}{|c|}{\textbf{45.6}} & \textbf{95.7} &
    \multicolumn{1}{c|}{59.9} & 93.0 &
    \multicolumn{1}{c|}{59.3} & 93.1 &
    \multicolumn{1}{c|}{61.4} & 92.7 &
    \multicolumn{1}{c|}{\underline{58.9}} & \underline{93.3}
    \\ 
 L$_{2}$ Average & 
    \multicolumn{1}{|c|}{\textbf{50.3}} & \textbf{94.8} &
    \multicolumn{1}{c|}{59.9} & \underline{93.0} &
    \multicolumn{1}{c|}{59.7} & 93.1 &
    \multicolumn{1}{c|}{\underline{59.3}} & 93.2 &
    \multicolumn{1}{c|}{59.8} & \underline{93.0}
    \\ 
 L$_{\infty}$ Average & 
\multicolumn{1}{|c|}{\textbf{34.1}} & \textbf{96.7} &
    \multicolumn{1}{c|}{\underline{42.8}} & \underline{96.0}  &
    \multicolumn{1}{c|}{44.7} & 95.8 &
    \multicolumn{1}{c|}{48.6} & 95.3 &
    \multicolumn{1}{c|}{44.9} & 95.8
    \\ 
No norm& 
    \multicolumn{1}{|c|}{\textbf{76.0}} & \textbf{88.2} &
\multicolumn{1}{|c|}{\underline{\textbf{76.0}}} & \underline{\textbf{88.2}} &
\multicolumn{1}{|c|}{\underline{\textbf{76.0}}} & \underline{\textbf{88.2}} &
\multicolumn{1}{|c|}{\underline{\textbf{76.0}}} & \underline{\textbf{88.2}} &
\multicolumn{1}{|c|}{\underline{\textbf{76.0}}} & \underline{\textbf{88.2}} 
   \\ \hline\midrule

\textbf{\method{LID}} &
  \multicolumn{1}{|c|}{\auc\%} &
   {\fpr\% } &
  \multicolumn{1}{c|}{\auc\%} &
   \fpr\% 
   &
  \multicolumn{1}{c|}{\auc\%} &
  \fpr\% &
  \multicolumn{1}{c|}{\auc\%} &
   \fpr\% &
  \multicolumn{1}{c|}{\auc\%} &
   \fpr\% \\ \midrule
L$_{1}$ Average& 
    \multicolumn{1}{|c|}{\textbf{79.9}} & \textbf{54.9} &
\multicolumn{1}{|c|}{\underline{83.7}} & 48.2 &
\multicolumn{1}{|c|}{84.0} & 50.0 &
\multicolumn{1}{|c|}{90.4} & \underline{52.1} &
\multicolumn{1}{|c|}{84.1} & 50.2
    \\ 
L$_{2}$ Average & 
    \multicolumn{1}{|c|}{\textbf{85.6}} & \textbf{46.2} &
\multicolumn{1}{|c|}{87.4} & 44.1 &
\multicolumn{1}{|c|}{87.0} & 45.1 &
\multicolumn{1}{|c|}{87.6} & 44.4 &
\multicolumn{1}{|c|}{\underline{86.1}} & \underline{45.4}
    \\ 
 L$_{\infty}$ Average & 
    \multicolumn{1}{|c|}{\textbf{77.9}} & \textbf{55.1} &
\multicolumn{1}{|c|}{83.3} & 46.3 &
\multicolumn{1}{|c|}{83.6} & 47.8 &
\multicolumn{1}{|c|}{88.7} & 38.8 &
\multicolumn{1}{|c|}{\underline{83.0}} & \underline{49.5}
    \\ 
No norm & 
    \multicolumn{1}{|c|}{\textbf{98.1}} & \textbf{8.2} &
\multicolumn{1}{|c|}{\underline{\textbf{98.1}}} & \underline{\textbf{8.2}} &
\multicolumn{1}{|c|}{\underline{\textbf{98.1}}} & \underline{\textbf{8.2}} &
\multicolumn{1}{|c|}{\underline{\textbf{98.1}}} & \underline{\textbf{8.2}} &
\multicolumn{1}{|c|}{\underline{\textbf{98.1}}} & \underline{\textbf{8.2}}
   \\ \hline\midrule

\textbf{\method{FS}} &
  \multicolumn{1}{|c|}{\auc\%} &
   {\fpr\% } &
  \multicolumn{1}{c|}{\auc\%} &
   \fpr\% 
   &
  \multicolumn{1}{c|}{\auc\%} &
  \fpr\% &
  \multicolumn{1}{c|}{\auc\%} &
   \fpr\% &
  \multicolumn{1}{c|}{\auc\%} &
   \fpr\% \\ \midrule
    L$_{1}$ Average & 
    \multicolumn{1}{|c|}{\textbf{79.8}} & \textbf{66.8} &
    \multicolumn{1}{c|}{83.4} & \underline{57.6} &
    \multicolumn{1}{c|}{83.5} & 57.1 &
    \multicolumn{1}{c|}{\underline{83.2}} & 53.0 &
    \multicolumn{1}{c|}{83.4} & 57.4
    \\ 
    L$_{2}$ Average & 
    \multicolumn{1}{|c|}{\textbf{73.5}} & \textbf{69.0} &
    \multicolumn{1}{c|}{75.6} & 65.0 &
    \multicolumn{1}{c|}{75.5} & 65.4 &
    \multicolumn{1}{c|}{\underline{74.5}} & \underline{67.0} &
    \multicolumn{1}{c|}{74.7} & 65.7
    \\ 
    L$_{\infty}$ Average & 
\multicolumn{1}{|c|}{\textbf{76.4}} & \textbf{63.5} &
    \multicolumn{1}{c|}{80.8} & 54.6 &
    \multicolumn{1}{c|}{80.2} & 54.6 &
    \multicolumn{1}{c|}{\underline{79.0}} & \underline{58.7} &
    \multicolumn{1}{c|}{80.4} & 58.2 
    \\ 
    No norm & 
 \multicolumn{1}{|c|}{\textbf{61.5}} & \textbf{85.9} &
\multicolumn{1}{|c|}{\underline{\textbf{61.5}}} & \underline{\textbf{85.9}} &
\multicolumn{1}{|c|}{\underline{\textbf{61.5}}} & \underline{\textbf{85.9}} &
\multicolumn{1}{|c|}{\underline{\textbf{61.5}}} & \underline{\textbf{85.9}} &
\multicolumn{1}{|c|}{\underline{\textbf{61.5}}} & \underline{\textbf{85.9}} 
       \\ \hline\midrule

\textbf{\method{MagNet}} &
  \multicolumn{1}{|c|}{\auc\%} &
   {\fpr\% } &
  \multicolumn{1}{c|}{\auc\%} &
   \fpr\% 
   &
  \multicolumn{1}{c|}{\auc\%} &
  \fpr\% &
  \multicolumn{1}{c|}{\auc\%} &
   \fpr\% &
  \multicolumn{1}{c|}{\auc\%} &
   \fpr\% \\ \midrule
L$_{1}$ Average & 
    \multicolumn{1}{|c|}{\textbf{98.1}} & 5.7 &
    \multicolumn{1}{c|}{98.2} & 5.4 &
    \multicolumn{1}{c|}{98.3} & \textbf{\underline{5.6}} &
    \multicolumn{1}{c|}{98.3} & 5.2  &
    \multicolumn{1}{c|}{\textbf{\underline{98.1}}} & \textbf{\underline{5.6}}
    \\ 
   L$_{2}$ Average & \multicolumn{1}{|c|}{90.0} & 28.7 &
    \multicolumn{1}{c|}{90.6} & 27.6 &
    \multicolumn{1}{c|}{90.8} & 27.8 &
    \multicolumn{1}{c|}{90.6} & \textbf{\underline{29.1}} &
    \multicolumn{1}{c|}{\textbf{\underline{89.7}}} & 28.1 \\
L$_{\infty}$ Average & 
    \multicolumn{1}{|c|}{98.5} & 10.3 &
    \multicolumn{1}{c|}{98.5} & 10.3 &
    \multicolumn{1}{c|}{\textbf{\underline{98.4}}} & \textbf{\underline{10.6}} &
    \multicolumn{1}{c|}{98.5} & 10.5 &
    \multicolumn{1}{c|}{98.5} & 10.4
    \\ 
 
No norm & 
\multicolumn{1}{|c|}{\textbf{86.9}} & \textbf{74.3} &
\multicolumn{1}{|c|}{\textbf{86.9}} & \textbf{74.3} &
\multicolumn{1}{|c|}{\textbf{86.9}} & \textbf{74.3} &
\multicolumn{1}{|c|}{\textbf{86.9}} &\textbf{ 74.3 }&
\multicolumn{1}{|c|}{\textbf{86.9}} & \textbf{74.3}
   \\ \hline

   \bottomrule
\end{tabular}
}
\caption{Overall performances on MNIST of all the detectors per objective and in {\mead}. The worst results among all the settings 
is in \textbf{bold}; the ones in the single-armed setting is \underline{underlined}. No norm denotes the group of attacks that do not depend on the norm constraint.}
\label{tab:mnist_overall}
\end{table*}
\subsection{Experimental setting}
\noindent \textbf{Evaluation metrics.}
For each L$_p$-norm and each considered $\varepsilon$, we apply our multi-armed detection scheme. We gather the global result considering all the attacks and all the objectives. Moreover, we also report the results per objective. 
The performance is  measured in terms of the \underline{\auc}~\cite{davis2006relationship} and in terms of \underline{\fpr}.
The first metric is the \textit{Area Under the Receiver Operating Characteristic curve} and represents the ability of the detector to discriminate between adversarial and natural examples (higher is better). The second metric represents the percentage of natural examples detected as adversarial when 95 \% of the adversarial examples are detected, i.e., FPR at 95 \% TPR (lower is better).
\subsubsection{Datasets and classifiers.}
We run the experiments on MNIST~\cite{lecun2010mnist} and CIFAR10~\cite{Cifar}. The underlying classifiers are a simple CNN for MNIST, consisting of two blocks of two convolutional layers, a max-pooling layer, one fully-connected layer, one dropout layer, two fully-connected layers, and ResNet-18 for CIFAR10. 
The training procedures involve $100$ epochs with Stochastic Gradient Descent (SGD) optimizer using a learning rate of $0.01$ for the simple CNN and $0.1$ for ResNet-18; a momentum of $0.9$ and a weight decay of $10^{-5}$ for ResNet-18. Once trained, these networks are fixed and never modified again.
\subsubsection{Grouping attacks.} 
We test the methods on the attacks presented in ~\cref{sec:review_attacks}, and we present them based on the norm constraint used to construct the attacks.
Under the L$_{1}$-norm fall PGD  with $\varepsilon$ in $\{5, 10, 15, 20, 25, 30, 40\}$.
Under the L$_{2}$-norm fall PGD  with $\varepsilon$ in $\{0.125, 0.25, 0.3125, 0.5, 1, 1.5, 2\}$, CW with $\varepsilon=0.01$, HOP with $\varepsilon=0.1$, and DF which has no constraint on $\varepsilon$.
Under the L$_{\infty}$-norm fall FGSM, BIM and PGD with $\varepsilon$ in $\{0.0315, 0.0625, 0.125, 0.25, 0.3125, 0.5\}$, CW with $\varepsilon = 0.3125$, and  SA with $\varepsilon = 0.3125$ for MNIST and $\varepsilon = 0.125$ for CIFAR10.
Finally, ST is not constrained by a norm or a maximum perturbation, as it is limited in maximum rotation (30 for CIFAR10 and 60 for MNIST) and translation (8 for CIFAR10 and 10 for MNIST).
\begin{table*}[t]
\centering
\resizebox{\columnwidth}{!}{%
\begin{tabular}{c||cc||cc||cc||cc||cc}
\toprule 
 &
  \multicolumn{6}{c||}{Supervised Methods} & \multicolumn{4}{c}{Unsupervised Methods}
   \\ \cline{2-11}
 &
  \multicolumn{2}{c||}{\method{NSS}} & \multicolumn{2}{c||}{\method{KD-BU}}& \multicolumn{2}{c||}{\method{LID}}& \multicolumn{2}{c||}{\method{FS}} &
  \multicolumn{2}{c}{\method{MagNet}}
   \\ \cline{2-11}
   
 &
  \multicolumn{1}{|c|}{\auc\%} &
   {\fpr\% } &
  \multicolumn{1}{c|}{\auc\%} &
   \fpr\% 
   &
  \multicolumn{1}{c|}{\auc\%} &
  \fpr\% &
  \multicolumn{1}{c|}{\auc\%} &
   \fpr\% &
  \multicolumn{1}{c|}{\auc\%} &
   \fpr\% \\ \hline\midrule
MNIST & 
    \multicolumn{1}{|c|}{\underline{90.7}} & \underline{\textbf{29.3}} &
    \multicolumn{1}{|c|}{51.5} & 93.9 &
    \multicolumn{1}{|c|}{85.4} & 41.1 &
    \multicolumn{1}{|c|}{72.8} & 71.3 &
    \multicolumn{1}{|c|}{\underline{\textbf{93.4}}} & \underline{29.8} 
    \\ 
    
CIFAR10 & 
    \multicolumn{1}{|c|}{\underline{\textbf{71.8}}} & \underline{\textbf{66.1}} &
    \multicolumn{1}{|c|}{53.0} & 95.2 &
    \multicolumn{1}{|c|}{64.0} & 81.8 &
    \multicolumn{1}{|c|}{\underline{66.4}} & 73.6 &
    \multicolumn{1}{|c|}{64.6} &  \underline{69.7}
   \\ \hline
   \bottomrule
\end{tabular}
}
\caption{Performances of each detection method under the \mead~framework on CIFAR10 and MNIST averaged over the norm-based constraint. The best results among all the methods is in \textbf{bold}; the ones per type of detection method (i.e. Supervised and Unsupervised) are \underline{underlined}.}
\label{tab:method_comparison}
\end{table*}
\subsubsection{Detection Methods.}\label{sec:detection_meth} 
We tested dection methods introduced in~\cref{sec:review_detectors}. 
In the supervised case, 
we train the detectors using adversarial examples created by perturbing the samples in the original training sets with PGD under L$_{\infty}$-norm and $\varepsilon=0.03125$.
~In the unsupervised case,
~the detectors only need natural samples in the training sets. They are tested on all the previously mentioned attacks, generated on the testing sets.

\subsection{Experimental results}
\label{sec:results}
In this section, we refer to \textit{single-armed setting} when we consider the setup where the adversarial examples are generated w.r.t. one of the objectives in~\cref{sec:adv_pb}. 
We provide the average of the performances of all the detection methods on CIFAR10 in~\cref{tab:cifar_overall} and on MNIST in~\cref{tab:mnist_overall}. Due to space constraints, the detailed tables for each detection method (i.e., \method{NSS}, \method{LID}, \method{KD-BU}, \method{MagNet}, and \method{FS}) and for each dataset (i.e., CIFAR10 and MNIST) are reported in~\cref{appendix:add_res}.

\subsubsection{\mead~and the single-armed setting.}
~\Cref{tab:cifar_overall} shows a decrease in the performance of all the detectors when going from the single-armed setting to~\mead. 
\method{NSS} is the more robust among the supervised methods when passing from the single-armed setting to the proposed setting. Indeed, (cf~\cref{tab:cifar_overall}), in terms of \auc, it registers a decrease of up 4.9 percentage points under the L$_1$-norm constraint, 4.7 under the L$_2$-norm constraint, and 5.3 under the L$_\infty$-norm constraint. This can be explained by the fact that the network in \method{NSS} is trained on the natural scene statistics extracted from the trained samples differently from the other detectors. In particular, these statistical
properties are altered by the presence of adversarial perturbations and hence are found to be a good candidate to determine if a sample is adversarial or not. By looking closely at the results for \method{NSS} in~\cref{tab:cifar_nss_per_loss}, it comes out that it performs better when evaluated on L$_\infty$ norm constraint. Indeed, in this case, the adversarial examples at testing time are similar to those used at training time. Not surprisingly, the performance decreases when evaluated on other kinds of attacks.
Notice that, in the single-armed setting, all the supervised methods turn out to be much more inefficient than when presented in the original papers. Indeed, as already explained in~\cref{sec:detection_meth}, we train the detectors using adversarial examples created by perturbing the samples in the original training sets with PGD under L$_{\infty}$-norm and $\varepsilon=0.03125$, and then we test them on a variety of attacks. Hence, we do not train a different detector for each kind of attack seen at testing time. On the other side, the unsupervised detector \method{MagNet} appears to be more robust than \method{FS} when changing from the single-armed setting to~\mead. Indeed, in terms of \auc, it loses at most 2.2 percentage points (L$_\infty$ norm case). On average, \method{FS} is the unsupervised detector that achieves the best performance on CIFAR10, while \method{MagNet} is the one to achieve the best performance on MNIST.

\paragraph{Remark:} Some single-armed setting results turn out to be worse than the corresponding results in \mead~(cf~\cref{tab:cifar_nss_per_loss}-\ref{tab:cifar_magnet_per_loss} and~\cref{tab:mnist_nss_per_loss}-\ref{tab:mnist_magnet_per_loss} in~\cref{appendix:add_res}). We provide here an  explanation of this phenomenon. Given a natural input sample $\mathbf{x}$, let $\mathbf{x}_{\ell}$ denotes the perturbed version of $\mathbf{x}$ according to some fixed norm $p$, fixed perturbation magnitude $\varepsilon$ and objective function $\ell$ between ACE, KL, Gini and FR. Suppose $f_{\theta}(\mathbf{x}_{\text{ACE}})=y$, where $y$ is the ground true label of $\mathbf{x}$, this means that  $\mathbf{x}_{\text{ACE}}$ is a perturbed version of the  natural example but not adversarial. Assume instead $f_{\theta}(\mathbf{x}_{\text{KL}})\not=y$,  $f_{\theta}(\mathbf{x}_{\text{Gini}})\not=y$ and  $f_{\theta}(\mathbf{x}_{\text{FR}})\not=y$. If at testing time the detector is able to recognize all of them as being positive (i.e., adversarial), then under {\mead}, $\mathbf{x}_{\text{KL}}, \mathbf{x}_{\text{Gini}}, \mathbf{x}_{\text{FR}}$ would be considered a \textit{true positive}. This example, counting as a true positive under~\mead, would instead be discarded under the single-armed setting of ACE, as $\mathbf{x}_{\text{ACE}}$ is neither a clean example nor an adversarial one. Then, the larger amount of true positives in~\mead~can potentially lead to an increase in the global \auc. 

\subsubsection{Effectiveness of the proposed objective functions.}\label{sec:attack_effectiveness} In~\cref{tab:cifar_attack_rate} and \cref{tab:mnist_attack_rate}, relegated to the Appendix due to space constraints,  we report the averaged number of successful adversarial examples under the multi-armed setting as well as the details per single-armed settings on CIFAR10 and MNIST, respectively.
The attacks are most successful when the value of the constraint $\varepsilon$ for every L$_{p}$-norm increases. 
Generating adversarial examples using the ACE for each attack scheme creates more harmful (adversarial) examples for the classifier than using any other objective. However, using either the Gini Impurity score, the Fisher-Rao objective, or the Kullback-Leibler divergence seems to create examples that are either equally or more difficult to be detected by the detection methods. For this purpose, we provide two examples. First, by looking at the results in~\cref{tab:cifar_lid_per_loss}, we can deduce that \method{LID} finds it difficult to recognize the attacks based on KL and FR objective functions but not the ones created through Gini. For example, with PGD1 and $\varepsilon=40$, we register a decrease in \auc~of 9.5 percentage points when going from the single-armed setting of Gini to the one of FR. Similarly, the decrease is 8.3 percentage points in the case of KL. This behavior is even more remarkable when we look at the results in terms of \fpr: the gap between the best \fpr~values (obtained via Gini) and the worst (via FR) is 30.7 percentage points. On the other side, the situation is reversed if we look at the results in~\cref{tab:cifar_fs_per_loss} as \method{FS} turns out to be highly inefficient at recognizing adversarial examples generated via the Gini Impurity score. By considering the results associated to the highest value of $\varepsilon$ for each norm, namely $\varepsilon = 40$ for L$_1$-norm; $\varepsilon = 2$ for L$_2$-norm; $\varepsilon = 0.5$ for L$_\infty$-norm, the gap between best \fpr~values (obtained via KL divergence) and the worst (via Gini Impurity score), varies from a minimum of $41.7$ (L$_\infty$-norm) to a maximum of $64.4$ (L$_2$-norm) percentage points. This example, in agreement with~\cref{sec:toy_example}, testify on real data that testing the detectors without taking into consideration the possibility of creating attacks through different objective functions leads to a biased and unrealistic estimation of their performances.

\subsubsection{Comparison between supervised and unsupervised detectors.}
The unsupervised methods find it challenging to recognize attacks crafted using the Gini Impurity score. Indeed, according to~\cref{sec:toy_example}, that objective function creates attacks on the decision boundary of the pre-trained classifier. Consequently, the unsupervised detectors can easily associate such input samples with the cluster of naturals. 
Supervised methods detect Adversarial Cross-Entropy loss-based attacks more and, therefore, more volatile when it comes to other types of loss-based attacks. Overall, by looking at the results in~\cref{tab:method_comparison} on both the datasets, most of the supervised and unsupervised methods achieve comparable performances with the multi-armed framework, meaning that the current use of the knowledge about the specific attack is not general enough. The exception to this is \method{NSS}, which, as already explained, seems to be the most general detector.

\subsubsection{On the effects of the norm and $\varepsilon$.}
The detection methods recognize attacks with a large perturbation more easily than other attacks  (cf~\cref{tab:cifar_nss_per_loss}-\ref{tab:cifar_magnet_per_loss} and~\cref{tab:mnist_nss_per_loss}-\ref{tab:mnist_magnet_per_loss}). L$_{\infty}$-norm attacks are less easily detectable than any other L$_{p}$-norm attack. 
Indeed, multiple attacks are tested simultaneously for a single $\varepsilon$ under the L$_\infty$ norm constraint. For example, in CIFAR10 with $\varepsilon=0.3125$ and L$_\infty$, PGD, FGSM, BIM, and CW are tested together, whereas, with any other norm constraint, only one typology of attack is examined. Indeed the more attack we consider for a given $\varepsilon$, the more likely at least one attack will remain undetected.
Globally, under the L$_{\infty}$-norm constraint, Gini Impurity score-based attacks are the least detected attacks. However, each method has different behaviors under L$_{1}$ and L$_{2}$. \method{NSS} is more sensitive to Kullback-Leibler divergence-based attacks while \method{MagNet} is more volatile to the Fisher-Rao distance-based attacks. As already pointed out, \method{FS} achieves inferior performance when evaluated against attacks crafted through the Gini Impurity objective, 
while the sensitivity of \method{LID} and \method{KD-BU} to a specific objective depends on the L$_p$-norm constraint.


\section{Summary and Concluding Remarks}\label{sec:conclusion}
We introduced \mead~a new framework to evaluate detection methods of adversarial examples. Contrary to what is generally assumed, the proposed setup ensures that the detector does not know the attacks at the testing time and is evaluated based on simultaneous attack strategies. Our experiments showed that the SOTA detectors for adversarial examples (both supervised and unsupervised) mostly fail when evaluated in \mead~with a remarkable deterioration in performance compared to single-armed settings. We enrich the proposed evaluation framework by involving three new objective functions to generate adversarial examples that create adversarial examples which can simultaneously fool the classifier while not being successfully identified by the investigated detectors. The poor performance of the current SOTA adversarial examples detectors should be seen as a challenge when developing novel methods. However, our evaluation framework assumes that the attackers do not know the detection method. As future work we plan to enrich the framework to a complete whitebox scenario.

\subsubsection*{Acknowledgements} The work of Federica Granese was supported by the European Research Council (ERC) project HYPATIA under the European Union’s Horizon 2020 research and innovation program. Grant agreement \textnumero 835294. This work has been supported by the project PSPC AIDA: 2019-PSPC-09 funded by BPI-France. This work was performed using HPC resources from GENCI-IDRIS (Grant 2022-[AD011012352R1]) and thanks to the Saclay-IA computing platform.

\bibliographystyle{splncs04}
\bibliography{biblio.bib}

\newpage
\appendix
\section{Additional results}
\label{appendix:add_res}
In the following sections, we include results that, due to space limitations, were not included in~\cref{sec:results}. 
\subsection{Additional Results on CIFAR10}\label{appendix_cifar_results}
\subsubsection{Attack Rate}
In \cref{tab:cifar_attack_rate}, we report the average number of successful attacks per natural sample considered in \mead~and in the single-armed settings. 

\begin{table}[!htb]
\centering
\resizebox{0.7\columnwidth}{!}{%
\setlength{\tabcolsep}{10pt}
\begin{tabular}{c||c||c||c||c||c}
\toprule

   \multicolumn{6}{c}{\textbf{Avg. Num. of Successful Attack / Tot. Num. of Attack }}
      \\ \cline{1-6} 
    \textbf{Norm L$_1$} &
  \multicolumn{1}{c||}{\mead} & \multicolumn{1}{c||}{ACE}& \multicolumn{1}{c||}{KL}& \multicolumn{1}{c||}{Gini} &
  \multicolumn{1}{c}{FR}
       \\ \midrule
     \underline{PGD1}  &\multicolumn{1}{c||}{ } & \multicolumn{1}{c||}{ } &  \multicolumn{1}{c||}{ } &  \multicolumn{1}{c||}{ } &  \multicolumn{1}{c}{ } \\
 $\varepsilon$ = 5 & 
   1.28 / 4 & 
   0.45 / 1 &          
   0.31 / 1 &     
   0.27 / 1 & 
   0.25 / 1   
     \\ 
 $\varepsilon$ = 10 & 
   2.24 / 4  
   & 0.81 / 1
   & 0.53 / 1
   & 0.47 / 1
   & 0.43 / 1
     \\ 
 $\varepsilon$ = 15 & 
   2.60 / 4 & 
   0.92 / 1 & 
   0.60 / 1 &
   0.59 / 1 & 
   0.50 / 1 
     \\ 
 $\varepsilon$ = 20 & 
   2.72 / 4 &    
   0.95 / 1 & 
   0.61 / 1 & 
   0.64 / 1& 
   0.52 /1 
     \\ 
 $\varepsilon$ = 25 & 
   2.76 / 4 &   
   0.95 / 1&
   0.61 / 1& 
   0.67 / 1& 
   0.53 / 1
     \\
 $\varepsilon$ = 30 & 
   2.80 / 4&   
   0.95 / 1 &
   0.62 / 1& 
   0.68 / 1& 
    0.54 / 1  
     \\
 $\varepsilon$ = 40 & 
   2.80 / 4&    
   0.95 / 1& 
   0.62 / 1& 
   0.70 / 1& 
   0.54 / 1
     \\ \hline\midrule
  
  
 \textbf{Norm L$_2$}  \\ \midrule
 
 \underline{PGD2}  &\multicolumn{1}{c||}{ } & \multicolumn{1}{c||}{ } &  \multicolumn{1}{c||}{ } &  \multicolumn{1}{c||}{ } &  \multicolumn{1}{c}{ } \\
$\varepsilon$ = 0.125 & 
   1.12 / 4& 
   0.39 / 1&    
   0.27 / 1&     
   0.24 / 1& 
   0.22 / 1 
     \\ 
$\varepsilon$ = 0.25 & 
   2.16 / 4& 
   0.78 / 1& 
   0.51 / 1&   
   0.45 / 1& 
   0.40 / 1 
     \\ 
$\varepsilon$ = 0.3125 & 
   2.40 / 4& 
   0.86 / 1& 
   0.56 / 1&
   0.54 / 1& 
   0.45 / 1 
     \\ 
$\varepsilon$ = 0.5 & 
   2.68 / 4& 
   0.93 / 1& 
   0.60 / 1&
   0.64 / 1& 
   0.50 / 1     
     \\ 
$\varepsilon$ = 1 & 
   2.76 / 4& 
   0.94 / 1& 
   0.61 / 1& 
   0.70 / 1& 
   0.52 / 1 
     \\
$\varepsilon$ = 1.5 & 
   2.76 / 4&
   0.94 / 1&
   0.61 / 1& 
   0.70 / 1&
   0.52 / 1  
     \\ 
$\varepsilon$ = 2 & 
   2.76 / 4& 
   0.94 / 1& 
   0.61 / 1& 
   0.70 / 1& 
   0.52 / 1  
     \\ \

   \underline{DeepFool}  &\multicolumn{1}{c||}{ } & \multicolumn{1}{c||}{ } &  \multicolumn{1}{c||}{ } &  \multicolumn{1}{c||}{ } &  \multicolumn{1}{c}{ }  \\
No $\varepsilon$ & 
    0.48 / 1& 
    0.48 / 1& 
    0.48 / 1& 
    0.48 / 1& 
    0.48 / 1 
     \\

   \underline{CW2}   &\multicolumn{1}{c||}{ } & \multicolumn{1}{c||}{ } &  \multicolumn{1}{c||}{ } &  \multicolumn{1}{c||}{ } &  \multicolumn{1}{c}{ } \\ 
$\varepsilon$ = 0.01 & 
   0.95 / 1& 
   0.95 / 1& 
   0.95 / 1& 
   0.95 / 1& 
   0.95 / 1
     \\

   \underline{HOP}   &\multicolumn{1}{c||}{ } & \multicolumn{1}{c||}{ } &  \multicolumn{1}{c||}{ } &  \multicolumn{1}{c||}{ } &  \multicolumn{1}{c}{ }   \\ 
$\varepsilon$ = 0.1 & 
   0.41 / 1& 
   0.41 / 1&     
   0.41 / 1&     
   0.41 / 1&   
   0.41  / 1 
     \\  \hline\midrule
  

\textbf{Norm L$_\infty$} \\ \midrule
 \underline{PGDi, FGSM, BIM}  &\multicolumn{1}{c||}{ } & \multicolumn{1}{c||}{ } &  \multicolumn{1}{c||}{ } &  \multicolumn{1}{c||}{ } &  \multicolumn{1}{c}{ } \\
$\varepsilon$ = 0.03125 & 
   8.28 / 12& 
   2.76 / 3& 
   1.86 / 3& 
   2.13 / 3& 
   1.53 / 3
     \\ 
$\varepsilon$ = 0.0625 & 
   8.52 / 12& 
   2.85 / 3&
   1.89 / 3&
   2.22 / 3&
   1.56 / 3
     \\ 
$\varepsilon$ = 0.25& 
   8.88 / 12&
   2.85 / 3& 
   1.98 / 3& 
   2.31 / 3& 
   1.71 / 3   
     \\ 
$\varepsilon$ = 0.5 & 
   9.24 / 12& 
   2.85 / 3&
   2.13 / 3& 
   2.31 / 3&
   1.92 / 3
     \\

\underline{PGDi, FGSM, BIM, SA}  &\multicolumn{1}{c||}{ } & \multicolumn{1}{c||}{ } &  \multicolumn{1}{c||}{ } &  \multicolumn{1}{c||}{ } &  \multicolumn{1}{c}{ } \\
$\varepsilon$ = 0.125 & 
    9.00 / 13 & 
   3.21 / 4& 
   2.26 / 4&
   2.61 / 4& 
   1.95 / 4   \\

 \underline{PGDi, FGSM, BIM, CWi}  &\multicolumn{1}{c||}{ } & \multicolumn{1}{c||}{ } &  \multicolumn{1}{c||}{ } &  \multicolumn{1}{c||}{ } &  \multicolumn{1}{c}{ } \\
$\varepsilon$ = 0.3125 & 
 9.88 / 13 & 
   3.79 / 4& 
   2.99 / 4&
   3.25 / 4& 
   2.73 / 4
     \\ \hline
  \midrule
  
\textbf{No norm}  \\ \midrule
   
   \underline{STA}  &\multicolumn{1}{c||}{ } & \multicolumn{1}{c||}{ } &  \multicolumn{1}{c||}{ } &  \multicolumn{1}{c||}{ } &  \multicolumn{1}{c}{ } \\
No $\varepsilon$ & 
    0.24 / 1& 
    0.24 / 1& 
    0.24 / 1& 
    0.24 / 1& 
    0.24 / 1
     \\ \hline
   \bottomrule
\end{tabular}
}
\caption{Average number of successful attacks per natural sample considered in the single-armed setting and \mead~(CIFAR10). The results are reported in the table together with the total number of attacks performed per natural sample(\textit{Avg. / Tot.}). No norm denotes the group of attacks that do not depend on the norm constraint.
}
\label{tab:cifar_attack_rate}
\end{table}

As explained in \cref{sec:results}, the attacks generated thanks to the Adversarial Cross-Entropy seem to be the most harmful ones for the underlying classifier. It is not surprising given that the Cross-Entropy was used as the loss to train the classifier. Attacks generated through the maximization of the Kullback-Leibler divergence, the Gini Impurity score, and the Fisher-Rao distance all seem to be equally damaging. 

\subsubsection{\method{NSS}}\label{appendix:cifar-nss}
In \Cref{tab:cifar_nss_per_loss}, we report the performances of the NSS detector in CIFAR10. 

\begin{table*}[!htb]
\centering
\resizebox{\columnwidth}{!}{%
\begin{tabular}{c||cc||cc||cc||cc||cc}
\toprule 
    \textbf{\method{NSS}} &
  \multicolumn{2}{c||}{\mead} & \multicolumn{2}{c||}{ACE}& \multicolumn{2}{c||}{KL}& \multicolumn{2}{c||}{Gini} &
  \multicolumn{2}{c}{FR}
   \\ \hline\midrule
  \textbf{Norm L$_1$}&
  \multicolumn{1}{|c|}{\auc\% } &
   {\fpr\% } &
  \multicolumn{1}{c|}{\auc\% } &
   \fpr\% 
   &
  \multicolumn{1}{c|}{\auc\% } &
  \fpr\% &
  \multicolumn{1}{c|}{\auc\% } &
   \fpr\% &
  \multicolumn{1}{c|}{\auc\% } &
   \fpr\% \\ \midrule
   \underline{PGD1}  & \multicolumn{1}{|c|}{ } & \multicolumn{1}{c||}{ } & \multicolumn{1}{c|}{ } & \multicolumn{1}{c||}{ } & \multicolumn{1}{c|}{ } & \multicolumn{1}{c||}{ } & \multicolumn{1}{c|}{ } & \multicolumn{1}{c||}{ } & \multicolumn{1}{c|}{ } & \multicolumn{1}{c}{ } \\
 $\varepsilon$ = 5 & 
    \multicolumn{1}{|c|}{\textbf{48.5}} & \textbf{94.2} &
    \multicolumn{1}{c|}{49.9} & \underline{93.5} &
    \multicolumn{1}{c|}{\underline{49.6}} & 93.0 &
    \multicolumn{1}{c|}{50.3} & 93.2 &
    \multicolumn{1}{c|}{49.9} & 93.3 \\ 
$\varepsilon$ = 10 & 
    \multicolumn{1}{|c|}{\textbf{54.0}} & \textbf{90.3} &
    \multicolumn{1}{c|}{56.9} & 88.4 &
    \multicolumn{1}{c|}{\underline{56.6}} & 88.3 &
    \multicolumn{1}{c|}{57.1} & \underline{88.8} &
    \multicolumn{1}{c|}{57.0} & 88.1 \\ 
$\varepsilon$ = 15 & 
    \multicolumn{1}{|c|}{\textbf{58.8}} & \textbf{86.8} &
    \multicolumn{1}{c|}{63.1} & 83.0 &
    \multicolumn{1}{c|}{\underline{62.8}} & 83.1 &
    \multicolumn{1}{c|}{63.5} & \underline{84.0} &
    \multicolumn{1}{c|}{63.2} & 82.5 \\ 
$\varepsilon$ = 20 & 
    \multicolumn{1}{|c|}{\textbf{63.5}} & \textbf{82.3} &
    \multicolumn{1}{c|}{68.5} & 77.1 &
    \multicolumn{1}{c|}{\underline{68.1}} & 77.3 &
    \multicolumn{1}{c|}{69.9} & \underline{77.6} &
    \multicolumn{1}{c|}{68.7} & 76.4 \\ 
$\varepsilon$ = 25 & 
    \multicolumn{1}{|c|}{\textbf{67.7}} & \textbf{77.2} &
    \multicolumn{1}{c|}{73.1} & 71.1 &
    \multicolumn{1}{c|}{\underline{72.7}} & \underline{71.8} &
    \multicolumn{1}{c|}{75.0} & 71.4 &
    \multicolumn{1}{c|}{73.4} & 70.9 \\ 
$\varepsilon$ = 30 & 
    \multicolumn{1}{|c|}{\textbf{71.4}} & \textbf{73.4} &
    \multicolumn{1}{c|}{77.1} & 64.5 &
    \multicolumn{1}{c|}{\underline{76.8}} & 65.1 &
    \multicolumn{1}{c|}{78.6} & \underline{67.3} &
    \multicolumn{1}{c|}{77.4} & 65.2
     \\ 
$\varepsilon$ = 40 & 
    \multicolumn{1}{|c|}{\textbf{76.1}} & \textbf{67.3} &
    \multicolumn{1}{c|}{83.5} & 52.7 &
    \multicolumn{1}{c|}{83.3} & 53.5 &
    \multicolumn{1}{c|}{\underline{80.1}} & \underline{64.9} &
    \multicolumn{1}{c|}{83.6} & 52.7
    \\
\hline
L$_{1}$ Average & \multicolumn{1}{|c|}{\textbf{62.9}} & \textbf{81.6} &
    \multicolumn{1}{c|}{67.4} & 75.7 &
    \multicolumn{1}{c|}{\underline{67.1}} & 76.0 &
    \multicolumn{1}{c|}{67.8} & \underline{78.2} &
    \multicolumn{1}{c|}{67.6} & 75.6
    \\ \hline
\midrule
  
  
 \textbf{Norm L$_2$}  &
  \multicolumn{1}{|c|}{\auc\% } &
   \fpr\% &
  \multicolumn{1}{c|}{\auc\% } &
   \fpr\% &
  \multicolumn{1}{c|}{\auc\% } &
   \fpr\% &
  \multicolumn{1}{c|}{\auc\% } &
   \fpr\% &
  \multicolumn{1}{c|}{\auc\% } &
   \fpr\%\\ \midrule
   \underline{PGD2}  & \multicolumn{1}{|c|}{ } & \multicolumn{1}{c||}{ } & \multicolumn{1}{c|}{ } & \multicolumn{1}{c||}{ } & \multicolumn{1}{c|}{ } & \multicolumn{1}{c||}{ } & \multicolumn{1}{c|}{ } & \multicolumn{1}{c||}{ } & \multicolumn{1}{c|}{ } & \multicolumn{1}{c}{ }\\
    $\varepsilon$ = 0.125 &
    \multicolumn{1}{|c|}{\textbf{48.3}} & \textbf{94.3} &
    \multicolumn{1}{c|}{49.5} & 93.8 &
    \multicolumn{1}{c|}{\underline{49.1}} & 93.5 &
    \multicolumn{1}{c|}{49.5} & \underline{\textbf{94.3}} &
    \multicolumn{1}{c|}{49.6} & 93.5
     \\ 
    $\varepsilon$ = 0.25 &
    \multicolumn{1}{|c|}{\textbf{53.2}} & \textbf{91.2} &
    \multicolumn{1}{c|}{55.9} & 89.1 &
    \multicolumn{1}{c|}{\underline{55.6}} & 89.2 &
    \multicolumn{1}{c|}{55.9} & \underline{89.8} &
    \multicolumn{1}{c|}{55.8} & 89.4
     \\ 
    $\varepsilon$ = 0.3125 &
    \multicolumn{1}{|c|}{\textbf{55.8}} & \textbf{89.2} &
    \multicolumn{1}{c|}{59.4} & 86.5 &
    \multicolumn{1}{c|}{\underline{59.0}} & 86.6 &
    \multicolumn{1}{c|}{59.3} & \underline{87.7} &
    \multicolumn{1}{c|}{59.3} & 86.6
     \\ 
    $\varepsilon$ = 0.5 &
    \multicolumn{1}{|c|}{\textbf{63.3}} & \textbf{82.6} &
    \multicolumn{1}{c|}{68.3} & 77.4 &
    \multicolumn{1}{c|}{\underline{68.0}} & 77.4 &
    \multicolumn{1}{c|}{69.0} & \underline{78.7} &
    \multicolumn{1}{c|}{68.4} & 77.2
     \\ 
    $\varepsilon$ = 1 &
    \multicolumn{1}{|c|}{\textbf{76.4}} & \textbf{67.5} &
    \multicolumn{1}{c|}{84.4} & 50.6 &
    \multicolumn{1}{c|}{84.3} & 50.5 &
    \multicolumn{1}{c|}{\underline{79.3}} & \underline{66.8} &
    \multicolumn{1}{c|}{84.7} & 50.7
     \\ 
    $\varepsilon$ = 1.5 &
    \multicolumn{1}{|c|}{81.0} & 63.0 &
    \multicolumn{1}{c|}{92.8} & 28.7 &
    \multicolumn{1}{c|}{92.7} & 28.9 &
    \multicolumn{1}{c|}{\underline{\textbf{79.5}}} & \underline{\textbf{66.5}} &
    \multicolumn{1}{c|}{93.0} & 27.3
     \\ 
    $\varepsilon$ = 2 &
    \multicolumn{1}{|c|}{82.6} & 62.3 &
    \multicolumn{1}{c|}{96.8} & 13.9 &
    \multicolumn{1}{c|}{96.9} & 13.1 &
    \multicolumn{1}{c|}{\underline{\textbf{79.5}}} & \underline{\textbf{66.5}} &
    \multicolumn{1}{c|}{95.9} & 17.2\\

   \multicolumn{1}{c||}{\underline{DeepFool}} & \multicolumn{1}{|c|}{ } & \multicolumn{1}{c||}{ } & \multicolumn{1}{c|}{ } & \multicolumn{1}{c||}{ } & \multicolumn{1}{c|}{ } & \multicolumn{1}{c||}{ } & \multicolumn{1}{c|}{ } & \multicolumn{1}{c||}{ } & \multicolumn{1}{c|}{ } & \multicolumn{1}{c}{ } \\ 
   No $\varepsilon$ & 
    \multicolumn{1}{|c|}{\textbf{57.0}} & \textbf{91.7} &
    \multicolumn{1}{c|}{\underline{\textbf{57.0}}} & \underline{\textbf{91.7}} &
    \multicolumn{1}{c|}{\underline{\textbf{57.0}}} & \underline{\textbf{91.7}} &
    \multicolumn{1}{c|}{\underline{\textbf{57.0}}} & \underline{\textbf{91.7}} &
    \multicolumn{1}{c|}{\underline{\textbf{57.0}}} & \underline{\textbf{91.7}}
     \\

   \multicolumn{1}{c||}{\underline{CW2}} & \multicolumn{1}{|c|}{ } & \multicolumn{1}{c||}{ } & \multicolumn{1}{c|}{ } & \multicolumn{1}{c||}{ } & \multicolumn{1}{c|}{ } & \multicolumn{1}{c||}{ } & \multicolumn{1}{c|}{ } & \multicolumn{1}{c||}{ } & \multicolumn{1}{c|}{ } & \multicolumn{1}{c}{ }  \\
    $\varepsilon$ = 0.01 & 
    \multicolumn{1}{|c|}{\textbf{56.4}} & \textbf{90.8} &
    \multicolumn{1}{c|}{\underline{\textbf{56.4}}} & \underline{\textbf{90.8}} &
    \multicolumn{1}{c|}{\underline{\textbf{56.4}}} & \underline{\textbf{90.8}} &
    \multicolumn{1}{c|}{\underline{\textbf{56.4}}} & \underline{\textbf{90.8}} &
    \multicolumn{1}{c|}{\underline{\textbf{56.4}}} & \underline{\textbf{90.8}}
     \\   

   \multicolumn{1}{c||}{\underline{HOP}}  & \multicolumn{1}{|c|}{ } & \multicolumn{1}{c||}{ } & \multicolumn{1}{c|}{ } & \multicolumn{1}{c||}{ } & \multicolumn{1}{c|}{ } & \multicolumn{1}{c||}{ } & \multicolumn{1}{c|}{ } & \multicolumn{1}{c||}{ } & \multicolumn{1}{c|}{ } & \multicolumn{1}{c}{ }   \\
    $\varepsilon$ = 0.1 &
    \multicolumn{1}{|c|}{\textbf{66.1}} & \textbf{87.0} &
    \multicolumn{1}{c|}{\underline{\textbf{66.1}}} & \underline{\textbf{87.0}} &
    \multicolumn{1}{c|}{\underline{\textbf{66.1}}} & \underline{\textbf{87.0}} &
    \multicolumn{1}{c|}{\underline{\textbf{66.1}}} & \underline{\textbf{87.0}} &
    \multicolumn{1}{c|}{\underline{\textbf{66.1}}} & \underline{\textbf{87.0}}

    \\
\hline

L$_2$ Average & 
    \multicolumn{1}{|c|}{\textbf{64.0}} & \textbf{82.0} &
    \multicolumn{1}{c|}{68.7} & 71.0 &
    \multicolumn{1}{c|}{68.5} & 70.9 &
    \multicolumn{1}{c|}{\underline{65.1}} & \underline{82.0} &
    \multicolumn{1}{c|}{68.6} & 71.1\\  \hline
\midrule

 \textbf{Norm L$_\infty$} &
  \multicolumn{1}{|c|}{\auc\% } &
   \fpr\% &
  \multicolumn{1}{c|}{\auc\% } &
   \fpr\% &
  \multicolumn{1}{c|}{\auc\% } &
   \fpr\% &
  \multicolumn{1}{c|}{\auc\% } &
   \fpr\% &
  \multicolumn{1}{c|}{\auc\% } &
   \fpr\%\\ \midrule
   
   \underline{PGDi, FGSM, BIM} & \multicolumn{1}{|c|}{ } & \multicolumn{1}{c||}{ } & \multicolumn{1}{c|}{ } & \multicolumn{1}{c||}{ } & \multicolumn{1}{c|}{ } & \multicolumn{1}{c||}{ } & \multicolumn{1}{c|}{ } & \multicolumn{1}{c||}{ } & \multicolumn{1}{c|}{ } & \multicolumn{1}{c}{ }\\
    $\varepsilon$ = 0.03125 &
    \multicolumn{1}{|c|}{\textbf{83.0}} & \textbf{55.3} &
    \multicolumn{1}{c|}{88.5} & 42.3 &
    \multicolumn{1}{c|}{89.6} & 39.9 &
    \multicolumn{1}{c|}{\underline{87.5}} & \underline{47.8} &
    \multicolumn{1}{c|}{89.3} & 39.8
     \\ 
    $\varepsilon$ = 0.0625 &
    \multicolumn{1}{|c|}{\textbf{96.0}} & \textbf{17.2} &
    \multicolumn{1}{c|}{98.1} & 7.9 &
    \multicolumn{1}{c|}{98.4} & 6.8 &
    \multicolumn{1}{c|}{\underline{97.1}} & \underline{13.2} &
    \multicolumn{1}{c|}{98.4} & 6.8
     \\ 
    $\varepsilon$ = 0.25 &
    \multicolumn{1}{|c|}{\textbf{97.3}} & \textbf{0.6} &
    \multicolumn{1}{c|}{99.7} & \underline{\textbf{0.6}} &
    \multicolumn{1}{c|}{99.7} & \underline{\textbf{0.6}} &
    \multicolumn{1}{c|}{\underline{97.8}} & \underline{\textbf{0.6}} &
    \multicolumn{1}{c|}{98.0} & \underline{\textbf{0.6}}
     \\ 
    $\varepsilon$ = 0.5 &
    \multicolumn{1}{|c|}{\textbf{82.5}} & \textbf{100.0} &
    \multicolumn{1}{c|}{99.7} & 0.6 &
    \multicolumn{1}{c|}{99.7} & 0.6 &
    \multicolumn{1}{c|}{86.2} & \underline{\textbf{100.0}} &
    \multicolumn{1}{c|}{\underline{85.7}} & \underline{\textbf{100.0}}
     \\
  
    \multicolumn{1}{c||}{\underline{PGDi, FGSM, BIM, SA}} & \multicolumn{1}{|c|}{ } & \multicolumn{1}{c||}{ } & \multicolumn{1}{c|}{ } & \multicolumn{1}{c||}{ } & \multicolumn{1}{c|}{ } & \multicolumn{1}{c||}{ } & \multicolumn{1}{c|}{ } & \multicolumn{1}{c||}{ } & \multicolumn{1}{c|}{ } & \multicolumn{1}{c}{ }  \\ 
    $\varepsilon$ = 0.125 &
    \multicolumn{1}{|c|}{9.4} & 99.9 &
    \multicolumn{1}{c|}{\underline{\textbf{9.4}}} & \underline{\textbf{99.9}} &
    \multicolumn{1}{c|}{\underline{\textbf{9.4}}} & \underline{\textbf{99.9}} &
    \multicolumn{1}{c|}{\underline{\textbf{9.4}}} & \underline{\textbf{99.9}} &
    \multicolumn{1}{c|}{\underline{\textbf{9.4}}} & \underline{\textbf{99.9}}
     \\

    \multicolumn{1}{c||}{\underline{PGDi, FGSM, BIM, CWi}} & \multicolumn{1}{|c|}{ } & \multicolumn{1}{c||}{ } & \multicolumn{1}{c|}{ } & \multicolumn{1}{c||}{ } & \multicolumn{1}{c|}{ } & \multicolumn{1}{c||}{ } & \multicolumn{1}{c|}{ } & \multicolumn{1}{c||}{ } & \multicolumn{1}{c|}{ } & \multicolumn{1}{c}{ }  \\ 
    $\varepsilon$ = 0.3125 &
    \multicolumn{1}{|c|}{\textbf{63.2}} & \textbf{99.1} &
    \multicolumn{1}{c|}{66.1} & 89.4 &
    \multicolumn{1}{c|}{66.1} & 89.4 &
    \multicolumn{1}{c|}{\underline{63.9}} & \underline{96.2} &
    \multicolumn{1}{c|}{\underline{63.9}} & 95.8

    \\
\hline
L$_\infty$ Average & 
    \multicolumn{1}{|c|}{\textbf{71.9}} & \textbf{62.0} &
    \multicolumn{1}{c|}{76.9} & 40.1 &
    \multicolumn{1}{c|}{77.2} & 39.5 &
    \multicolumn{1}{c|}{\underline{73.7}} & \underline{59.6} &
    \multicolumn{1}{c|}{74.1} & 57.2
    \\ \hline
\midrule

  \textbf{No norm} &
  \multicolumn{1}{|c|}{\auc\% } &
   \fpr\% &
  \multicolumn{1}{c|}{\auc\% } &
   \fpr\% &
  \multicolumn{1}{c|}{\auc\% } &
   \fpr\% &
  \multicolumn{1}{c|}{\auc\% } &
   \fpr\% &
  \multicolumn{1}{c|}{\auc\% } &
   \fpr\%\\ \midrule
   \underline{STA}  & \multicolumn{1}{|c|}{ } & \multicolumn{1}{c||}{ } & \multicolumn{1}{c|}{ } & \multicolumn{1}{c||}{ } & \multicolumn{1}{c|}{ } & \multicolumn{1}{c||}{ } & \multicolumn{1}{c|}{ } & \multicolumn{1}{c||}{ } & \multicolumn{1}{c|}{ } & \multicolumn{1}{c}{ } \\
No $\varepsilon$ & 
   \multicolumn{1}{|c|}{\textbf{88.5}} & \textbf{38.8} &
    \multicolumn{1}{c|}{\underline{\textbf{88.5}}} & \underline{\textbf{38.8}} &
    \multicolumn{1}{c|}{\underline{\textbf{88.5}}} & \underline{\textbf{38.8}} &
    \multicolumn{1}{c|}{\underline{\textbf{88.5}}} & \underline{\textbf{38.8}} &
    \multicolumn{1}{c|}{\underline{\textbf{88.5}}} & \underline{\textbf{38.8}}
     \\ \hline
No norm Average & \multicolumn{1}{|c|}{\textbf{88.5}} & \textbf{38.8} &
    \multicolumn{1}{c|}{\underline{\textbf{88.5}}} & \underline{\textbf{38.8}} &
    \multicolumn{1}{c|}{\underline{\textbf{88.5}}} & \underline{\textbf{38.8}} &
    \multicolumn{1}{c|}{\underline{\textbf{88.5}}} & \underline{\textbf{38.8}} &
    \multicolumn{1}{c|}{\underline{\textbf{88.5}}} & \underline{\textbf{38.8}}
    \\ \hline
   \bottomrule
\end{tabular}
}
\caption{Performances on \method{NSS} per objective and in \mead~on CIFAR10. The worst results among all the settings 
is in \textbf{bold}; the ones in the single-armed setting is \underline{underlined}. No norm denotes the group of attacks that do not depend on the norm constraint.}
\label{tab:cifar_nss_per_loss}
\end{table*}
\method{NSS} is, by far, the best performing  method to detect adversarial examples under the \mead~framework that we consider in this paper. The decrease in performances due to the worst-case scenario that we consider is up to 5.0 percentage points in terms of \auc. Under the single-armed setting, \method{NSS} is the most sensitive to the Kullback-Leibler divergence, even though the results are quite similar amongst the different attack objectives. 
\newpage
\subsubsection{\method{KD-BU}}\label{appendix:cifar-kd-bu}
In~\cref{tab:cifar_kd_bu_per_loss}, we show the result of our \mead~framework as well as the single-armed settings on CIFAR10, evaluated on \method{KD-BU}.

\begin{table*}[!htb]
\centering
\resizebox{\columnwidth}{!}{%
\begin{tabular}{c||cc||cc||cc||cc||cc}
\toprule 
    \textbf{\method{KD-BU}} &
  \multicolumn{2}{c||}{\mead} & \multicolumn{2}{c||}{ACE}& \multicolumn{2}{c||}{KL}& \multicolumn{2}{c||}{Gini} &
  \multicolumn{2}{c}{FR}
   \\ \hline\midrule
  \textbf{Norm L$_1$}&
  \multicolumn{1}{|c|}{\auc\%} &
   \fpr\%  &
  \multicolumn{1}{c|}{\auc\%} &
   \fpr\% 
   &
  \multicolumn{1}{c|}{\auc\%} &
  \fpr\% &
  \multicolumn{1}{c|}{\auc\%} &
   \fpr\% &
  \multicolumn{1}{c|}{\auc\%} &
   \fpr\% \\ \midrule
   \underline{PGD1}  & \multicolumn{1}{|c|}{ } & \multicolumn{1}{c||}{ } & \multicolumn{1}{c|}{ } & \multicolumn{1}{c||}{ } & \multicolumn{1}{c|}{ } & \multicolumn{1}{c||}{ } & \multicolumn{1}{c|}{ } & \multicolumn{1}{c||}{ } & \multicolumn{1}{c|}{ } & \multicolumn{1}{c}{ } \\
   $\varepsilon$ = 5 &
\multicolumn{1}{|c|}{\textbf{41.3}} & \textbf{96.6} &
\multicolumn{1}{|c|}{58.0} & 94.7 &
\multicolumn{1}{|c|}{57.1} & \underline{94.8} &
\multicolumn{1}{|c|}{67.9} & 93.0 &
\multicolumn{1}{|c|}{\underline{56.5}} & \underline{94.8}
   \\ 
$\varepsilon$ = 10 &
\multicolumn{1}{|c|}{\textbf{36.9}} & \textbf{97.2} &
\multicolumn{1}{|c|}{55.1} & 94.7 &
\multicolumn{1}{|c|}{\underline{54.0}} & 94.7 &
\multicolumn{1}{|c|}{72.4} & 91.9 &
\multicolumn{1}{|c|}{54.6} & \underline{94.8}
   \\ 
$\varepsilon$ = 15 &
\multicolumn{1}{|c|}{\textbf{39.9}} & \textbf{96.7} &
\multicolumn{1}{|c|}{\underline{57.9}} & \underline{94.5} &
\multicolumn{1}{|c|}{\underline{57.9}} & \underline{94.5} &
\multicolumn{1}{|c|}{73.2} & 92.9 &
\multicolumn{1}{|c|}{58.2} & \underline{94.5}
   \\ 
$\varepsilon$ = 20 &
\multicolumn{1}{|c|}{\textbf{47.3}} & \textbf{96.3} &
\multicolumn{1}{|c|}{66.7} & 93.2 &
\multicolumn{1}{|c|}{66.9} & 93.2 &
\multicolumn{1}{|c|}{75.9} & 92.3 &
\multicolumn{1}{|c|}{\underline{65.9}} & \underline{93.4}
   \\ 
$\varepsilon$ = 25 &
\multicolumn{1}{|c|}{\textbf{55.5}} & \textbf{95.6} &
\multicolumn{1}{|c|}{\underline{75.5}} & 91.0 &
\multicolumn{1}{|c|}{76.2} & \underline{90.8} &
\multicolumn{1}{|c|}{76.5} & 92.0 &
\multicolumn{1}{|c|}{75.7} & 91.0
   \\ 
$\varepsilon$ = 30 &
\multicolumn{1}{|c|}{\textbf{62.6}} & \textbf{94.7} &
\multicolumn{1}{|c|}{83.5} & 86.9 &
\multicolumn{1}{|c|}{84.3} & 86.3 &
\multicolumn{1}{|c|}{\underline{77.2}} & \underline{91.8} &
\multicolumn{1}{|c|}{84.1} & 86.4
   \\ 
$\varepsilon$ = 40 &
\multicolumn{1}{|c|}{\textbf{72.6}} & \textbf{92.6} &
\multicolumn{1}{|c|}{93.5} & 65.4 &
\multicolumn{1}{|c|}{93.5} & 64.5 &
\multicolumn{1}{|c|}{\underline{77.0}} & \underline{91.9} &
\multicolumn{1}{|c|}{93.7} & 63.9
   \\ \hline

 L$_{1}$ Average & \multicolumn{1}{|c|}{\textbf{50.9}} & \textbf{95.7} &
    \multicolumn{1}{|c|}{70.0} & 88.6 &
    \multicolumn{1}{|c|}{70.0} & 88.4 &
    \multicolumn{1}{|c|}{74.3} & \underline{92.3} &
    \multicolumn{1}{|c|}{\underline{69.8}} & 88.4
        \\ \hline
   \midrule
  
  
\textbf{Norm L$_2$} &
  \multicolumn{1}{|c|}{\auc\%} &
   \fpr\% &
  \multicolumn{1}{|c|}{\auc\%} &
   \fpr\% &
  \multicolumn{1}{|c|}{\auc\%} &
   \fpr\% &
  \multicolumn{1}{|c|}{\auc\%} &
   \fpr\% &
  \multicolumn{1}{|c|}{\auc\%} &
   \fpr\%\\ \midrule
   \underline{PGD2}  & \multicolumn{1}{|c|}{ } & \multicolumn{1}{c||}{ } & \multicolumn{1}{c|}{ } & \multicolumn{1}{c||}{ } & \multicolumn{1}{c|}{ } & \multicolumn{1}{c||}{ } & \multicolumn{1}{c|}{ } & \multicolumn{1}{c||}{ } & \multicolumn{1}{c|}{ } & \multicolumn{1}{c}{ }\\
$\varepsilon$ = 0.125 &
\multicolumn{1}{|c|}{\textbf{42.0}} & \textbf{96.6} &
\multicolumn{1}{|c|}{59.0} & \underline{94.6} &
\multicolumn{1}{|c|}{58.2} & \underline{94.6} &
\multicolumn{1}{|c|}{68.2} & 92.6 &
\multicolumn{1}{|c|}{\underline{57.8}} & \underline{94.6}
   \\ 
$\varepsilon$ = 0.25 &
\multicolumn{1}{|c|}{\textbf{38.4}} & \textbf{96.8} &
\multicolumn{1}{|c|}{54.2} & \underline{95.0} &
\multicolumn{1}{|c|}{\underline{53.9}} & \underline{95.0} &
\multicolumn{1}{|c|}{70.5} & 92.5 &
\multicolumn{1}{|c|}{55.0} & 94.8
   \\ 
$\varepsilon$ = 0.3125 &
\multicolumn{1}{|c|}{\textbf{38.6}} & \textbf{96.8} &
\multicolumn{1}{|c|}{\underline{55.1}} & \underline{94.7} &
\multicolumn{1}{|c|}{55.8} & 94.6 &
\multicolumn{1}{|c|}{72.9} & 92.0 &
\multicolumn{1}{|c|}{55.6} & 94.6
   \\ 
$\varepsilon$ = 0.5 &
\multicolumn{1}{|c|}{\textbf{47.9}} & \textbf{96.2} &
\multicolumn{1}{|c|}{\underline{66.8}} & \underline{93.2} &
\multicolumn{1}{|c|}{67.8} & 92.9 &
\multicolumn{1}{|c|}{75.4} & 92.4 &
\multicolumn{1}{|c|}{67.0} & 93.1
   \\ 
$\varepsilon$ = 1 &
\multicolumn{1}{|c|}{\textbf{74.2}} & \textbf{92.1} &
\multicolumn{1}{|c|}{94.4} & 58.1 &
\multicolumn{1}{|c|}{94.6} & 55.7 &
\multicolumn{1}{|c|}{\underline{77.1}} & \underline{91.8} &
\multicolumn{1}{|c|}{94.7} & 57.3
   \\ 
$\varepsilon$ = 1.5 &
\multicolumn{1}{|c|}{80.1} & 90.1 &
\multicolumn{1}{|c|}{99.0} & 0.0 &
\multicolumn{1}{|c|}{99.2} & 0.0 &
\multicolumn{1}{|c|}{\textbf{\underline{77.1}}} & \textbf{\underline{91.9}} &
\multicolumn{1}{|c|}{99.3} & 0.0
   \\ 
$\varepsilon$ = 2 &
\multicolumn{1}{|c|}{81.3} & 89.7 &
\multicolumn{1}{|c|}{99.8} & 0.0 &
\multicolumn{1}{|c|}{99.9} & 0.0 &
\multicolumn{1}{|c|}{\textbf{\underline{77.1}}} & \textbf{\underline{91.9}} &
\multicolumn{1}{|c|}{99.8} & 0.0
   \\

   \underline{DeepFool} & \multicolumn{1}{|c|}{ } & \multicolumn{1}{c||}{ } & \multicolumn{1}{c|}{ } & \multicolumn{1}{c||}{ } & \multicolumn{1}{c|}{ } & \multicolumn{1}{c||}{ } & \multicolumn{1}{c|}{ } & \multicolumn{1}{c||}{ } & \multicolumn{1}{c|}{ } & \multicolumn{1}{c}{ }\\ 
No $\varepsilon$ & 
\multicolumn{1}{|c|}{\textbf{67.1}} & \textbf{94.0} &
\multicolumn{1}{|c|}{\textbf{\underline{67.1}}} & \textbf{\underline{94.0}} &
\multicolumn{1}{|c|}{\textbf{\underline{67.1}}} & \textbf{\underline{94.0}} &
\multicolumn{1}{|c|}{\textbf{\underline{67.1}}} & \textbf{\underline{94.0}} &
\multicolumn{1}{|c|}{\textbf{\underline{67.1}}} & \textbf{\underline{94.0}} 
   \\

   \underline{CW2} & \multicolumn{1}{|c|}{ } & \multicolumn{1}{c||}{ } & \multicolumn{1}{c|}{ } & \multicolumn{1}{c||}{ } & \multicolumn{1}{c|}{ } & \multicolumn{1}{c||}{ } & \multicolumn{1}{c|}{ } & \multicolumn{1}{c||}{ } & \multicolumn{1}{c|}{ } & \multicolumn{1}{c}{ }\\
$\varepsilon$ = 0.01 &
\multicolumn{1}{|c|}{\textbf{53.0}} & \textbf{95.1} &
\multicolumn{1}{|c|}{\textbf{\underline{53.0}}} & \textbf{\underline{95.1}} &
\multicolumn{1}{|c|}{\textbf{\underline{53.0}}} & \textbf{\underline{95.1}} &
\multicolumn{1}{|c|}{\textbf{\underline{53.0}}} & \textbf{\underline{95.1}} &
\multicolumn{1}{|c|}{\textbf{\underline{53.0}}} & \textbf{\underline{95.1}} 
   \\

   \underline{HOP} & \multicolumn{1}{|c|}{ } & \multicolumn{1}{c||}{ } & \multicolumn{1}{c|}{ } & \multicolumn{1}{c||}{ } & \multicolumn{1}{c|}{ } & \multicolumn{1}{c||}{ } & \multicolumn{1}{c|}{ } & \multicolumn{1}{c||}{ } & \multicolumn{1}{c|}{ } & \multicolumn{1}{c}{ }\\
$\varepsilon$ = 0.1 &
\multicolumn{1}{|c|}{\textbf{67.3}} & \textbf{94.0} &
\multicolumn{1}{|c|}{\textbf{\underline{67.3}}} & \textbf{\underline{94.0}} &
\multicolumn{1}{|c|}{\textbf{\underline{67.3}}} & \textbf{\underline{94.0}} &
\multicolumn{1}{|c|}{\textbf{\underline{67.3}}} & \textbf{\underline{94.0}} &
\multicolumn{1}{|c|}{\textbf{\underline{67.3}}} & \textbf{\underline{94.0}} 
   \\ \hline
 L$_{2}$ Average & 
    \multicolumn{1}{|c|}{\textbf{59.0}} & \textbf{94.1} &
    \multicolumn{1}{|c|}{71.6} & 71.9 &
    \multicolumn{1}{|c|}{71.7} & 71.6 &
    \multicolumn{1}{|c|}{\underline{70.6}} & \underline{92.8} &
    \multicolumn{1}{|c|}{71.7} & 71.8 
        \\ \hline
\midrule
  

\textbf{Norm L$_\infty$}  &
  \multicolumn{1}{|c|}{\auc\%} &
   \fpr\% &
  \multicolumn{1}{|c|}{\auc\%} &
   \fpr\% &
  \multicolumn{1}{|c|}{\auc\%} &
   \fpr\% &
  \multicolumn{1}{|c|}{\auc\%} &
   \fpr\% &
  \multicolumn{1}{|c|}{\auc\%} &
   \fpr\%\\ \midrule
   
   \underline{PGDi, FGSM, BIM}  & \multicolumn{1}{|c|}{ } & \multicolumn{1}{c||}{ } & \multicolumn{1}{c|}{ } & \multicolumn{1}{c||}{ } & \multicolumn{1}{c|}{ } & \multicolumn{1}{c||}{ } & \multicolumn{1}{c|}{ } & \multicolumn{1}{c||}{ } & \multicolumn{1}{c|}{ } & \multicolumn{1}{c}{ }\\
$\varepsilon$ = 0.03125 &
\multicolumn{1}{|c|}{\textbf{29.2}} & \textbf{97.3} &
\multicolumn{1}{|c|}{52.8} & 94.7 &
\multicolumn{1}{|c|}{62.4} & 92.7 &
\multicolumn{1}{|c|}{\underline{46.5}} & \underline{96.2} &
\multicolumn{1}{|c|}{58.2} & 93.8
   \\ 
$\varepsilon$ = 0.0625 &
\multicolumn{1}{|c|}{\textbf{35.2}} & \textbf{96.9} &
\multicolumn{1}{|c|}{67.1} & 91.7 &
\multicolumn{1}{|c|}{74.8} & 88.7 &
\multicolumn{1}{|c|}{\underline{52.3}} & \underline{95.7} &
\multicolumn{1}{|c|}{75.0} & 89.0
   \\ 
$\varepsilon$ = 0.25 &
\multicolumn{1}{|c|}{\textbf{45.1}} & \textbf{96.4} &
\multicolumn{1}{|c|}{84.4} & 84.8 &
\multicolumn{1}{|c|}{83.5} & 86.6 &
\multicolumn{1}{|c|}{\underline{65.0}} & \underline{94.5} &
\multicolumn{1}{|c|}{81.8} & 88.5
   \\
$\varepsilon$ = 0.5 &
\multicolumn{1}{|c|}{\textbf{43.1}} & \textbf{96.6} &
\multicolumn{1}{|c|}{77.4} & 90.7 &
\multicolumn{1}{|c|}{79.5} & 89.2 &
\multicolumn{1}{|c|}{\underline{68.0}} & \underline{94.1} &
\multicolumn{1}{|c|}{83.4} & 88.3
   \\


   \underline{PGDi, FGSM, BIM, SA} & \multicolumn{1}{|c|}{ } & \multicolumn{1}{c||}{ } & \multicolumn{1}{c|}{ } & \multicolumn{1}{c||}{ } & \multicolumn{1}{c|}{ } & \multicolumn{1}{c||}{ } & \multicolumn{1}{c|}{ } & \multicolumn{1}{c||}{ } & \multicolumn{1}{c|}{ } & \multicolumn{1}{c}{ }\\
$\varepsilon$ = 0.125 &
\multicolumn{1}{|c|}{\textbf{34.9}} & \textbf{97.0} &
\multicolumn{1}{|c|}{59.2} & 94.9 &
\multicolumn{1}{|c|}{60.5} & 94.9 &
\multicolumn{1}{|c|}{\underline{46.5}} & \underline{96.4} &
\multicolumn{1}{|c|}{60.6} & 94.9
   \\
  
 \underline{PGDi, FGSM, BIM, CWi~} & \multicolumn{1}{|c|}{ } & \multicolumn{1}{c||}{ } & \multicolumn{1}{c|}{ } & \multicolumn{1}{c||}{ } & \multicolumn{1}{c|}{ } & \multicolumn{1}{c||}{ } & \multicolumn{1}{c|}{ } & \multicolumn{1}{c||}{ } & \multicolumn{1}{c|}{ } & \multicolumn{1}{c}{ }\\
$\varepsilon$ = 0.3125 &
\multicolumn{1}{|c|}{\textbf{33.2}} & \textbf{97.1} &
\multicolumn{1}{|c|}{47.8} & 95.8 &
\multicolumn{1}{|c|}{47.7} & 95.8 &
\multicolumn{1}{|c|}{\underline{43.8}} & \underline{96.4} &
\multicolumn{1}{|c|}{47.7} & 95.9
   \\ \hline
 L$_{\infty}$ Average & 
 \multicolumn{1}{|c|}{\textbf{36.8}} & \textbf{96.9} &
    \multicolumn{1}{|c|}{64.8} & 92.1 &
    \multicolumn{1}{|c|}{68.1} & 91.3 &
    \multicolumn{1}{|c|}{\underline{53.7}} & \underline{95.6} &
    \multicolumn{1}{|c|}{67.8} & 91.7 
        \\ \hline

  \midrule
  
  \textbf{No norm}  &
  \multicolumn{1}{|c|}{\auc\%} &
   \fpr\% &
  \multicolumn{1}{|c|}{\auc\%} &
   \fpr\% &
  \multicolumn{1}{|c|}{\auc\%} &
   \fpr\% &
  \multicolumn{1}{|c|}{\auc\%} &
   \fpr\% &
  \multicolumn{1}{|c|}{\auc\%} &
   \fpr\%\\ \midrule
   \underline{STA}  & \multicolumn{1}{|c|}{ } & \multicolumn{1}{c||}{ } & \multicolumn{1}{c|}{ } & \multicolumn{1}{c||}{ } & \multicolumn{1}{c|}{ } & \multicolumn{1}{c||}{ } & \multicolumn{1}{c|}{ } & \multicolumn{1}{c||}{ } & \multicolumn{1}{c|}{ } & \multicolumn{1}{c}{ }\\
No $\varepsilon$ & 
\multicolumn{1}{|c|}{\textbf{65.4}} & \textbf{94.2} &
\multicolumn{1}{|c|}{\textbf{\underline{65.4}}} & \textbf{\underline{94.2}} &
\multicolumn{1}{|c|}{\textbf{\underline{65.4}}} & \textbf{\underline{94.2}} & \multicolumn{1}{|c|}{\textbf{\underline{65.4}}} & \textbf{\underline{94.2}} &
\multicolumn{1}{|c|}{\textbf{\underline{65.4}}} & \textbf{\underline{94.2}} 
   \\ \hline
No norm Average & \multicolumn{1}{|c|}{\textbf{65.4}} & \textbf{94.2} &
\multicolumn{1}{|c|}{\textbf{\underline{65.4}}} & \textbf{\underline{94.2}} &
\multicolumn{1}{|c|}{\textbf{\underline{65.4}}} & \textbf{\underline{94.2}} & \multicolumn{1}{|c|}{\textbf{\underline{65.4}}} & \textbf{\underline{94.2}} &
\multicolumn{1}{|c|}{\textbf{\underline{65.4}}} & \textbf{\underline{94.2}}
    \\ \hline
   \bottomrule
\end{tabular}
}
\caption{Performances on \method{KD-BU} per objective and in \mead~on CIFAR10. The worst results among all the settings 
are shown in \textbf{bold}; the ones in the single-armed setting is \underline{underlined}. No norm denotes the group of attacks that do not depend on the norm constraint.}
\label{tab:cifar_kd_bu_per_loss}
\end{table*}

\method{KD-BU} seems to work poorly on \mead~as well as on the single-armed setting. For most settings, \method{KD-BU} is worst than a random detector. The decrease in \auc~between the worst single-armed setting and \mead~is up to 24.9 percentage points.

\newpage
\subsubsection{\method{LID}}\label{appendix:cifar-lid}
In \Cref{tab:cifar_lid_per_loss}, we report the detection performances of the \method{LID} method under the \mead~framework and the different single-armed settings. 

\begin{table*}[!htb]
\centering
\resizebox{\columnwidth}{!}{%
\begin{tabular}{c||cc||cc||cc||cc||cc}
\toprule 
    \textbf{\method{LID}} &
  \multicolumn{2}{c||}{\mead} & \multicolumn{2}{c||}{ACE}& \multicolumn{2}{c||}{KL}& \multicolumn{2}{c||}{Gini} &
  \multicolumn{2}{c}{FR}
   \\ \hline\midrule
  \textbf{Norm L$_1$}&
  \multicolumn{1}{|c|}{\auc\%} &
   {\fpr\% } &
  \multicolumn{1}{c|}{\auc\%} &
   \fpr\% 
   &
  \multicolumn{1}{c|}{\auc\%} &
  \fpr\% &
  \multicolumn{1}{c|}{\auc\%} &
   \fpr\% &
  \multicolumn{1}{c|}{\auc\%} &
   \fpr\% \\ \midrule
   \underline{PGD1}  & \multicolumn{1}{|c|}{ } & \multicolumn{1}{c||}{ } & \multicolumn{1}{c|}{ } & \multicolumn{1}{c||}{ } & \multicolumn{1}{c|}{ } & \multicolumn{1}{c||}{ } & \multicolumn{1}{c|}{ } & \multicolumn{1}{c||}{ } & \multicolumn{1}{c|}{ } & \multicolumn{1}{c}{ } \\
 $\varepsilon$ = 5 & 
   \multicolumn{1}{|c|}{\textbf{57.5}} & \textbf{93.5} & 
   \multicolumn{1}{|c|}{66.2} & 86.9 & 
   \multicolumn{1}{|c|}{65.5} & 88.0 &
    \multicolumn{1}{|c|}{77.9} & 77.0 &
    \multicolumn{1}{|c|}{\underline{64.4}} & \underline{88.7} 
     \\
 $\varepsilon$ = 10 & 
   \multicolumn{1}{|c|}{\textbf{48.3}} &   \textbf{96.4}&
   \multicolumn{1}{|c|}{62.3} & 89.6 & 
   \multicolumn{1}{|c|}{\underline{60.6}} & \underline{92.0} & \multicolumn{1}{|c|}{84.3} & 70.7  &
  \multicolumn{1}{|c|}{61.9} & 90.2 
     \\ 
 $\varepsilon$ = 15 & 
   \multicolumn{1}{|c|}{\textbf{53.1}} & \textbf{95.4} & 
   \multicolumn{1}{|c|}{\underline{61.6}} & 90.7 &
   \multicolumn{1}{|c|}{61.8} & 90.9 &
   \multicolumn{1}{|c|}{89.1} & 55.6 &
    \multicolumn{1}{|c|}{61.8} & \underline{91.1}
     \\ 
 $\varepsilon$ = 20 & 
    \multicolumn{1}{|c|}{\textbf{37.5}} &  \textbf{99.0} & 
  \multicolumn{1}{|c|}{65.7} & 89.2 &
 \multicolumn{1}{|c|}{\underline{65.4}} & \underline{90.2} &
 \multicolumn{1}{|c|}{91.4} & 40.8 &
   \multicolumn{1}{|c|}{66.3} & 87.1  
     \\ 
 $\varepsilon$ = 25 & 
   \multicolumn{1}{|c|}{\textbf{47.7}} & \textbf{96.3} &  
   \multicolumn{1}{|c|}{70.7} & 84.0 &
  \multicolumn{1}{|c|}{70.9} & 84.7 &   \multicolumn{1}{|c|}{92.8} & 37.1 &
   \multicolumn{1}{|c|}{\underline{70.3}} & \underline{83.3}
     \\ 
 $\varepsilon$ = 30 & 
   \multicolumn{1}{|c|}{\textbf{56.4}} & \textbf{95.0} &  
   \multicolumn{1}{|c|}{76.1} & 75.7  &
   \multicolumn{1}{|c|}{76.5} & 79.8 &
   \multicolumn{1}{|c|}{93.2} & 35.2  &
  \multicolumn{1}{|c|}{\underline{75.5}} & \underline{81.9}
     \\ 
 $\varepsilon$ = 40 & 
   \multicolumn{1}{|c|}{\textbf{54.9}} & \textbf{92.5} &  
   \multicolumn{1}{|c|}{84.6} & 58.5 &
   \multicolumn{1}{|c|}{85.0} & 54.7 &
   \multicolumn{1}{|c|}{93.3} & 32.6 &
    \multicolumn{1}{|c|}{\underline{83.8}} & \underline{63.3} 
     \\ \hline
    L$_{1}$ Average & 
    \multicolumn{1}{|c|}{\textbf{50.8}} & \textbf{95.4} &
    \multicolumn{1}{|c|}{69.6} & 82.1 &
    \multicolumn{1}{|c|}{69.4} & 82.9 &
    \multicolumn{1}{|c|}{88.9} & 49.9 &
    \multicolumn{1}{|c|}{\underline{69.1}} & \underline{83.7}
        \\ \hline
   \midrule
  
  
\textbf{Norm L$_2$} &
  \multicolumn{1}{|c|}{\auc\%} &
   \fpr\% &
  \multicolumn{1}{|c|}{\auc\%} &
   \fpr\% &
  \multicolumn{1}{|c|}{\auc\%} &
   \fpr\% &
  \multicolumn{1}{|c|}{\auc\%} &
   \fpr\% &
  \multicolumn{1}{|c|}{\auc\%} &
   \fpr\%\\ \midrule
   
   \underline{PGD2}  & \multicolumn{1}{|c|}{ } & \multicolumn{1}{c||}{ } & \multicolumn{1}{c|}{ } & \multicolumn{1}{c||}{ } & \multicolumn{1}{c|}{ } & \multicolumn{1}{c||}{ } & \multicolumn{1}{c|}{ } & \multicolumn{1}{c||}{ } & \multicolumn{1}{c|}{ } & \multicolumn{1}{c}{ }\\
$\varepsilon$ = 0.125 & 
   \multicolumn{1}{|c|}{\textbf{61.3}} & \textbf{90.6} & 
   \multicolumn{1}{|c|}{67.5} & 86.3 & 
   \multicolumn{1}{|c|}{66.2} & 88.7  & \multicolumn{1}{|c|}{77.8} & 76.3  &
    \multicolumn{1}{|c|}{\underline{64.6}} & \underline{88.8} 
     \\ 
$\varepsilon$ = 0.25 & 
   \multicolumn{1}{|c|}{\textbf{47.9}} & \textbf{96.5} & 
   \multicolumn{1}{|c|}{62.0} & 90.3 & 
   \multicolumn{1}{|c|}{\underline{60.8}} & \underline{91.5}  & \multicolumn{1}{|c|}{84.1} & 68.3  &
    \multicolumn{1}{|c|}{61.4} & 90.6
     \\ 
$\varepsilon$ = 0.3125 & 
   \multicolumn{1}{|c|}{\textbf{49.7}} & \textbf{95.8} & 
   \multicolumn{1}{|c|}{60.8} & \underline{91.5} & 
   \multicolumn{1}{|c|}{\underline{60.5}} & \underline{91.5}  & \multicolumn{1}{|c|}{87.4} & 60.3  &
    \multicolumn{1}{|c|}{61.9} & 88.8
     \\ 
$\varepsilon$ = 0.5 & 
   \multicolumn{1}{|c|}{\textbf{47.6}} & \textbf{97.9} & 
   \multicolumn{1}{|c|}{66.1} & 87.7 & 
   \multicolumn{1}{|c|}{\underline{65.6}} & \underline{91.2}  & \multicolumn{1}{|c|}{91.0} & 49.2  &
    \multicolumn{1}{|c|}{65.9} & 88.9
     \\ 
$\varepsilon$ = 1 & 
   \multicolumn{1}{|c|}{\textbf{65.2}} & \textbf{84.1} & 
   \multicolumn{1}{|c|}{86.0} & 50.6 & 
   \multicolumn{1}{|c|}{86.3} & 48.9  & \multicolumn{1}{|c|}{93.4} & 32.5  &
    \multicolumn{1}{|c|}{\underline{85.2}} & \underline{55.9} 
     \\ 
$\varepsilon$ = 1.5 & 
  \multicolumn{1}{|c|}{\textbf{77.0}} & \textbf{56.2} & 
   \multicolumn{1}{|c|}{94.0} & 20.0 & 
   \multicolumn{1}{|c|}{93.9} & 19.8  & \multicolumn{1}{|c|}{93.5} &  \underline{31.5} &
    \multicolumn{1}{|c|}{\underline{93.1}} & 21.7
     \\ 
$\varepsilon$ = 2 & 
   \multicolumn{1}{|c|}{\textbf{81.5}} & \textbf{46.4} & 
   \multicolumn{1}{|c|}{96.3} & 11.9 & 
   \multicolumn{1}{|c|}{96.3} & 12.2  & \multicolumn{1}{|c|}{\underline{93.4}} & \underline{31.8}  &
    \multicolumn{1}{|c|}{95.0} & 15.2
     \\

   \underline{DeepFool}  & \multicolumn{1}{|c|}{ } & \multicolumn{1}{c||}{ } & \multicolumn{1}{c|}{ } & \multicolumn{1}{c||}{ } & \multicolumn{1}{c|}{ } & \multicolumn{1}{c||}{ } & \multicolumn{1}{c|}{ } & \multicolumn{1}{c||}{ } & \multicolumn{1}{c|}{ } & \multicolumn{1}{c}{ }\\
    No $\varepsilon$  & 
   \multicolumn{1}{|c|}{\textbf{70.9}} & \textbf{86.9} & 
   \multicolumn{1}{|c|}{\underline{\textbf{70.9}}} & \underline{\textbf{86.9}} & 
   \multicolumn{1}{|c|}{\underline{\textbf{70.9}}} &  \underline{\textbf{86.9}} & \multicolumn{1}{|c|}{\underline{\textbf{70.9}}} & \underline{\textbf{86.9}}  &
    \multicolumn{1}{|c|}{\underline{\textbf{70.9}}} &\underline{\textbf{86.9}}
     \\

   \underline{CW2}  & \multicolumn{1}{|c|}{ } & \multicolumn{1}{c||}{ } & \multicolumn{1}{c|}{ } & \multicolumn{1}{c||}{ } & \multicolumn{1}{c|}{ } & \multicolumn{1}{c||}{ } & \multicolumn{1}{c|}{ } & \multicolumn{1}{c||}{ } & \multicolumn{1}{c|}{ } & \multicolumn{1}{c}{ }\\
$\varepsilon$ = 0.01 & 
   \multicolumn{1}{|c|}{\textbf{61.6}} & \textbf{92.5} & 
   \multicolumn{1}{|c|}{\underline{\textbf{61.6}}} & \underline{\textbf{92.5}} & 
   \multicolumn{1}{|c|}{\underline{\textbf{61.6}}} & \underline{\textbf{92.5}}  & \multicolumn{1}{|c|}{\underline{\textbf{61.6}}} &  \underline{\textbf{92.5}} &
    \multicolumn{1}{|c|}{\underline{\textbf{61.6}}} &\underline{\textbf{92.5}}
     \\

   \underline{HOP}  & \multicolumn{1}{|c|}{ } & \multicolumn{1}{c||}{ } & \multicolumn{1}{c|}{ } & \multicolumn{1}{c||}{ } & \multicolumn{1}{c|}{ } & \multicolumn{1}{c||}{ } & \multicolumn{1}{c|}{ } & \multicolumn{1}{c||}{ } & \multicolumn{1}{c|}{ } & \multicolumn{1}{c}{ }\\
$\varepsilon$ = 0.1 & 
  \multicolumn{1}{|c|}{\textbf{72.2}} & \textbf{83.6} & 
   \multicolumn{1}{|c|}{\underline{\textbf{72.2}}} & \underline{\textbf{83.6}} & 
   \multicolumn{1}{|c|}{\underline{\textbf{72.2}}} & \underline{\textbf{83.6}}  & \multicolumn{1}{|c|}{\underline{\textbf{72.2}}} & \underline{\textbf{83.6}}  &
    \multicolumn{1}{|c|}{\underline{\textbf{72.2}}} &\underline{\textbf{83.6}}
     \\  \hline
    L$_{2}$ Average & 
    \multicolumn{1}{|c|}{\textbf{63.5}} & \textbf{83.1} &
    \multicolumn{1}{|c|}{73.7} & 70.1 &
    \multicolumn{1}{|c|}{73.4} & 70.7 &
    \multicolumn{1}{|c|}{82.5} & 61.3 &
    \multicolumn{1}{|c|}{\underline{73.2}} & \underline{71.3}
        \\ \hline
   \midrule
  

\textbf{Norm L$_\infty$}  &
  \multicolumn{1}{|c|}{\auc\%} &
   \fpr\% &
  \multicolumn{1}{|c|}{\auc\%} &
   \fpr\% &
  \multicolumn{1}{|c|}{\auc\%} &
   \fpr\% &
  \multicolumn{1}{|c|}{\auc\%} &
   \fpr\% &
  \multicolumn{1}{|c|}{\auc\%} &
   \fpr\%\\ \midrule
   \underline{PGDi, FGSM, BIM}  & \multicolumn{1}{|c|}{ } & \multicolumn{1}{c||}{ } & \multicolumn{1}{c|}{ } & \multicolumn{1}{c||}{ } & \multicolumn{1}{c|}{ } & \multicolumn{1}{c||}{ } & \multicolumn{1}{c|}{ } & \multicolumn{1}{c||}{ } & \multicolumn{1}{c|}{ } & \multicolumn{1}{c}{ }\\
$\varepsilon$ = 0.03125 & 
   \multicolumn{1}{|c|}{\textbf{41.4}} & \textbf{96.5} &  
   \multicolumn{1}{|c|}{\underline{55.5}} & \underline{89.9}  &
   \multicolumn{1}{|c|}{60.5} & 88.9  &
  \multicolumn{1}{|c|}{69.7} & 86.7  &
    \multicolumn{1}{|c|}{65.5} & 82.5 
     \\
$\varepsilon$ = 0.0625 & 
  \multicolumn{1}{|c|}{\textbf{58.4}} & \textbf{83.9} & 
   \multicolumn{1}{|c|}{\underline{76.7}} & 59.0  & 
   \multicolumn{1}{|c|}{78.6} & 57.0 & 
  \multicolumn{1}{|c|}{83.6} & \underline{59.2} &
    \multicolumn{1}{|c|}{87.1} & 44.8
     \\ 
$\varepsilon$ = 0.25& 
 \multicolumn{1}{|c|}{\textbf{58.0}} & \textbf{89.5} &
   \multicolumn{1}{|c|}{88.4} & 27.0 &
  \multicolumn{1}{|c|}{92.1} & 41.7 &  \multicolumn{1}{|c|}{\underline{77.7}} & \underline{70.9} &
   \multicolumn{1}{|c|}{93.0} & 20.6
     \\ 
$\varepsilon$ = 0.5 & 
   \multicolumn{1}{|c|}{\textbf{58.6}} & \textbf{86.7} &
   \multicolumn{1}{|c|}{90.3} & 25.7  &  
   \multicolumn{1}{|c|}{93.6} & 19.2  &  \multicolumn{1}{|c|}{\underline{76.3}} & \underline{74.3}  &
   \multicolumn{1}{|c|}{91.8} & 23.9  
     \\ 
   
   \underline{PGDi, FGSM, BIM, SA} & \multicolumn{1}{|c|}{ } & \multicolumn{1}{c||}{ } & \multicolumn{1}{c|}{ } & \multicolumn{1}{c||}{ } & \multicolumn{1}{c|}{ } & \multicolumn{1}{c||}{ } & \multicolumn{1}{c|}{ } & \multicolumn{1}{c||}{ } & \multicolumn{1}{c|}{ } & \multicolumn{1}{c}{ }\\
$\varepsilon$ = 0.125 & 
   \multicolumn{1}{|c|}{\textbf{56.7}} & \textbf{91.8} &
   \multicolumn{1}{|c|}{82.5} & 48.6   & 
   \multicolumn{1}{|c|}{85.6} & 49.8  &  \multicolumn{1}{|c|}{\underline{64.6}} & \underline{91.0}  &
   \multicolumn{1}{|c|}{85.5} &  48.8 
     \\ 
  
  
 \underline{PGDi, FGSM, BIM, CWi}   & \multicolumn{1}{|c|}{ } & \multicolumn{1}{c||}{ } & \multicolumn{1}{c|}{ } & \multicolumn{1}{c||}{ } & \multicolumn{1}{c|}{ } & \multicolumn{1}{c||}{ } & \multicolumn{1}{c|}{ } & \multicolumn{1}{c||}{ } & \multicolumn{1}{c|}{ } & \multicolumn{1}{c}{ }\\
$\varepsilon$ = 0.3125 & 
   \multicolumn{1}{|c|}{\textbf{49.5}} & \textbf{96.1} &
   \multicolumn{1}{|c|}{60.5} &  90.7  & 
   \multicolumn{1}{|c|}{68.7} & 88.8  &  \multicolumn{1}{|c|}{\underline{56.0}}& \underline{96.0}  &
   \multicolumn{1}{|c|}{68.8} & 87.9  
     \\ \hline
   
    L$_{\infty}$ Average & \multicolumn{1}{|c|}{\textbf{53.8}} & \textbf{90.8} &
    \multicolumn{1}{|c|}{75.7} & 56.8 &
    \multicolumn{1}{|c|}{79.9} & 57.6 &
    \multicolumn{1}{|c|}{\underline{71.3}} & \underline{79.7} &
    \multicolumn{1}{|c|}{82.0} & 51.4
        \\ \hline
  \midrule
  
\textbf{No norm} &
  \multicolumn{1}{|c|}{\auc\%} &
   \fpr\% &
  \multicolumn{1}{|c|}{\auc\%} &
   \fpr\% &
  \multicolumn{1}{|c|}{\auc\%} &
   \fpr\% &
  \multicolumn{1}{|c|}{\auc\%} &
   \fpr\% &
  \multicolumn{1}{|c|}{\auc\%} &
   \fpr\%\\ \midrule
   
   \underline{STA}  & \multicolumn{1}{|c|}{ } & \multicolumn{1}{c||}{ } & \multicolumn{1}{c|}{ } & \multicolumn{1}{c||}{ } & \multicolumn{1}{c|}{ } & \multicolumn{1}{c||}{ } & \multicolumn{1}{c|}{ } & \multicolumn{1}{c||}{ } & \multicolumn{1}{c|}{ } & \multicolumn{1}{c}{ }\\
No $\varepsilon$ & 
   \multicolumn{1}{|c|}{\textbf{88.0}} & \textbf{58.1} & 
   \multicolumn{1}{|c|}{\underline{\textbf{88.0}}} & \underline{\textbf{58.1}} & 
   \multicolumn{1}{|c|}{\underline{\textbf{88.0}}} &  \underline{\textbf{58.1}} & 
   \multicolumn{1}{|c|}{\underline{\textbf{88.0}}} & \underline{\textbf{58.1}}  &
    \multicolumn{1}{|c|}{\underline{\textbf{88.0}}} &\underline{\textbf{58.1}}
     \\ \hline
   No norm Average &\multicolumn{1}{|c|}{\textbf{88.0}} & \textbf{58.1} & 
   \multicolumn{1}{|c|}{\underline{\textbf{88.0}}} & \underline{\textbf{58.1}} & 
   \multicolumn{1}{|c|}{\underline{\textbf{88.0}}} &  \underline{\textbf{58.1}} & 
   \multicolumn{1}{|c|}{\underline{\textbf{88.0}}} & \underline{\textbf{58.1}}  &
    \multicolumn{1}{|c|}{\underline{\textbf{88.0}}} &\underline{\textbf{58.1}}
    \\ \hline
   \bottomrule
\end{tabular}
}
\caption{Performances on \method{LID} per objective and in \mead~on CIFAR10. The worst results among all the settings 
is in \textbf{bold}; the ones in the single-armed setting is \underline{underlined}. No norm denotes the group of attacks that do not depend on the norm constraint.}
\label{tab:cifar_lid_per_loss}
\end{table*}
\method{LID} is quite sensitive to all the attacker's objectives. Depending on the norm-constraint and on the $\varepsilon$ value, each one of the four objectives can be the most harmful one. Moreover, this detection method is quite affected by the \mead~setting. Indeed, the decrease of performances in terms of \auc due to the use of the worst-case scenario is up to 28.9 percentage points compared to the worst single-armed setting (value obtained under the L$_{1}$-norm constraint for $\varepsilon$ = 40).

\newpage

\subsubsection{\method{FS}}\label{appendix:cifar-fs}
In~\cref{tab:cifar_fs_per_loss}, we present the summary of the \method{FS} detection method. 

\begin{table*}[!htb]
\centering
\resizebox{\columnwidth}{!}{%
\begin{tabular}{c||cc||cc||cc||cc||cc}
\toprule 
    \textbf{\method{FS}} &
  \multicolumn{2}{c||}{\mead} & \multicolumn{2}{c||}{ACE}& \multicolumn{2}{c||}{KL}& \multicolumn{2}{c||}{Gini} &
  \multicolumn{2}{c}{FR}
   \\ \hline\midrule
  \textbf{Norm L$_1$}&
  \multicolumn{1}{|c|}{\auc\%} &
   \fpr\%  &
  \multicolumn{1}{c|}{\auc\%} &
   \fpr\% 
   &
  \multicolumn{1}{c|}{\auc\%} &
  \fpr\% &
  \multicolumn{1}{c|}{\auc\%} &
   \fpr\% &
  \multicolumn{1}{c|}{\auc\%} &
   \fpr\% \\ \midrule
   \underline{PGD1}  & \multicolumn{1}{|c|}{ } & \multicolumn{1}{c||}{ } & \multicolumn{1}{c|}{ } & \multicolumn{1}{c||}{ } & \multicolumn{1}{c|}{ } & \multicolumn{1}{c||}{ } & \multicolumn{1}{c|}{ } & \multicolumn{1}{c||}{ } & \multicolumn{1}{c|}{ } & \multicolumn{1}{c}{ } \\
 $\varepsilon$ = 5 &
    \multicolumn{1}{|c|}{69.1} & 76.1 &
    \multicolumn{1}{|c|}{76.5} & 66.5 &
    \multicolumn{1}{|c|}{76.8} & 65.8 &
    \multicolumn{1}{|c|}{\underline{\textbf{68.2}}} & \underline{\textbf{77.9}} &
    \multicolumn{1}{|c|}{76.5} & 66.4
       \\ 
    $\varepsilon$ = 10 &
    \multicolumn{1}{|c|}{76.6} & 65.9 &
    \multicolumn{1}{|c|}{88.3} & 42.2 &
    \multicolumn{1}{|c|}{88.4} & 42.5 &
    \multicolumn{1}{|c|}{\underline{\textbf{74.8}}} & \underline{\textbf{68.4}} &
    \multicolumn{1}{|c|}{88.6} & 42.1
       \\ 
    $\varepsilon$ = 15 &
    \multicolumn{1}{|c|}{77.2} & 61.9 &
    \multicolumn{1}{|c|}{93.6} & 26.3 &
    \multicolumn{1}{|c|}{93.8} & 25.6 &
    \multicolumn{1}{|c|}{\underline{\textbf{76.7}}} & \underline{\textbf{63.8}} &
    \multicolumn{1}{|c|}{94.0} & 25.0
       \\ 
    $\varepsilon$ = 20 &
    \multicolumn{1}{|c|}{78.1} & 60.0 &
    \multicolumn{1}{|c|}{96.3} & 16.5 &
    \multicolumn{1}{|c|}{96.4} & 16.1 &
    \multicolumn{1}{|c|}{\underline{\textbf{76.3}}} & \underline{\textbf{63.2}} &
    \multicolumn{1}{|c|}{96.5} & 15.3
       \\
    $\varepsilon$ = 25 &
    \multicolumn{1}{|c|}{76.9} & 61.7 &
    \multicolumn{1}{|c|}{97.6} & 11.5 &
    \multicolumn{1}{|c|}{97.7} & 11.3 &
    \multicolumn{1}{|c|}{\underline{\textbf{74.7}}} & \underline{\textbf{64.9}} &
    \multicolumn{1}{|c|}{97.7} & 10.9
       \\
    $\varepsilon$ = 30 &
    \multicolumn{1}{|c|}{75.5} & 63.4 &
    \multicolumn{1}{|c|}{98.3} & 8.1 &
    \multicolumn{1}{|c|}{98.4} & 8.0 &
    \multicolumn{1}{|c|}{\underline{\textbf{72.6}}} & \underline{\textbf{66.8}} &
    \multicolumn{1}{|c|}{98.4} & 7.9
       \\ 
    $\varepsilon$ = 40 &
    \multicolumn{1}{|c|}{74.6} & 64.9 &
    \multicolumn{1}{|c|}{99.0} & 5.0 &
    \multicolumn{1}{|c|}{99.0} & 4.8 &
    \multicolumn{1}{|c|}{\underline{\textbf{71.4}}} & \underline{\textbf{68.1}} &
    \multicolumn{1}{|c|}{99.0} & 5.0
         \\ \hline
    L$_{1}$ Average & \multicolumn{1}{|c|}{75.4} & 64.8 &
    \multicolumn{1}{|c|}{92.8} & 25.1 &
    \multicolumn{1}{|c|}{92.9} & 24.9 &
    \multicolumn{1}{|c|}{\textbf{\underline{73.5}}} & \textbf{\underline{67.6}} &
    \multicolumn{1}{|c|}{92.9} & 24.6 
    \\ \hline
 \midrule
  
  
 \textbf{Norm L$_2$} &
  \multicolumn{1}{|c|}{\auc\%} &
  \fpr\% &
  \multicolumn{1}{|c|}{\auc\%} &
  \fpr\% &
  \multicolumn{1}{|c|}{\auc\%} &
  \fpr\% &
  \multicolumn{1}{|c|}{\auc\%} &
  \fpr\% &
  \multicolumn{1}{|c|}{\auc\%} &
  \fpr\%\\ \midrule
  \underline{PGD2}  & \multicolumn{1}{|c|}{ } & \multicolumn{1}{c||}{ } & \multicolumn{1}{c|}{ } & \multicolumn{1}{c||}{ } & \multicolumn{1}{c|}{ } & \multicolumn{1}{c||}{ } & \multicolumn{1}{c|}{ } & \multicolumn{1}{c||}{ } & \multicolumn{1}{c|}{ } & \multicolumn{1}{c}{ } \\
    $\varepsilon$ = 0.125 &
    \multicolumn{1}{|c|}{\textbf{67.5}} & \textbf{77.4} &
    \multicolumn{1}{|c|}{74.6} & 68.9 &
    \multicolumn{1}{|c|}{75.1} & 67.6 &
    \multicolumn{1}{|c|}{\underline{68.9}} & \underline{76.9} &
    \multicolumn{1}{|c|}{74.8} & 69.2
       \\ 
    $\varepsilon$ = 0.25 &
    \multicolumn{1}{|c|}{75.9} & 66.1 &
    \multicolumn{1}{|c|}{87.3} & 45.0 &
    \multicolumn{1}{|c|}{87.4} & 45.0 &
    \multicolumn{1}{|c|}{\underline{\textbf{74.7}}} & \underline{\textbf{67.7}} &
    \multicolumn{1}{|c|}{87.4} & 45.4
       \\
    $\varepsilon$ = 0.3125 &
    \multicolumn{1}{|c|}{77.0} & 64.1 &
    \multicolumn{1}{|c|}{90.7} & 35.5 &
    \multicolumn{1}{|c|}{90.9} & 34.8 &
    \multicolumn{1}{|c|}{\underline{\textbf{76.1}}} & \underline{\textbf{65.7}} &
    \multicolumn{1}{|c|}{90.9} & 35.3
       \\ 
    $\varepsilon$ = 0.5 &
    \multicolumn{1}{|c|}{77.6} & 61.4 &
    \multicolumn{1}{|c|}{96.2} & 17.0 &
    \multicolumn{1}{|c|}{96.3} & 16.6 &
    \multicolumn{1}{|c|}{\underline{\textbf{75.7}}} & \underline{\textbf{64.6}} &
    \multicolumn{1}{|c|}{96.3} & 16.5
       \\ 
    $\varepsilon$ = 1 &
    \multicolumn{1}{|c|}{74.7} & 65.0 &
    \multicolumn{1}{|c|}{99.0} & 4.8 &
    \multicolumn{1}{|c|}{99.0} & 4.6 &
    \multicolumn{1}{|c|}{\underline{\textbf{71.2}}} & \underline{\textbf{67.9}} &
    \multicolumn{1}{|c|}{98.9} & 5.0
       \\ 
    $\varepsilon$ = 1.5 &
    \multicolumn{1}{|c|}{74.5} & 65.1 &
    \multicolumn{1}{|c|}{99.3} & 3.4 &
    \multicolumn{1}{|c|}{99.3} & 3.5 &
    \multicolumn{1}{|c|}{\underline{\textbf{71.1}}} & \underline{\textbf{68.0}} &
    \multicolumn{1}{|c|}{98.9} & 4.9
       \\ 
    $\varepsilon$ = 2 &
    \multicolumn{1}{|c|}{73.7} & 65.2 &
    \multicolumn{1}{|c|}{99.3} & 3.6 &
    \multicolumn{1}{|c|}{99.3} & 3.6 &
    \multicolumn{1}{|c|}{\underline{\textbf{71.1}}} & \underline{\textbf{68.0}} &
    \multicolumn{1}{|c|}{98.7} & 5.7
       \\

  \underline{DeepFool} & \multicolumn{1}{|c|}{ } & \multicolumn{1}{c||}{ } & \multicolumn{1}{c|}{ } & \multicolumn{1}{c||}{ } & \multicolumn{1}{c|}{ } & \multicolumn{1}{c||}{ } & \multicolumn{1}{c|}{ } & \multicolumn{1}{c||}{ } & \multicolumn{1}{c|}{ } & \multicolumn{1}{c}{ } \\ 
    No $\varepsilon$ &
    \multicolumn{1}{|c|}{\textbf{66.0}} & \textbf{78.2} &
    \multicolumn{1}{|c|}{\underline{\textbf{66.0}}} & \underline{\textbf{78.2}} &
    \multicolumn{1}{|c|}{\underline{\textbf{66.0}}} & \underline{\textbf{78.2}} &
    \multicolumn{1}{|c|}{\underline{\textbf{66.0}}} & \underline{\textbf{78.2}} &
    \multicolumn{1}{|c|}{\underline{\textbf{66.0}}} & \underline{\textbf{78.2}}
       \\ 
  
  \underline{CW2}  & \multicolumn{1}{|c|}{ } & \multicolumn{1}{c||}{ } & \multicolumn{1}{c|}{ } & \multicolumn{1}{c||}{ } & \multicolumn{1}{c|}{ } & \multicolumn{1}{c||}{ } & \multicolumn{1}{c|}{ } & \multicolumn{1}{c||}{ } & \multicolumn{1}{c|}{ } & \multicolumn{1}{c}{ } \\ 
    $\varepsilon$ = 0.01 &
    \multicolumn{1}{|c|}{\textbf{86.7}} & \textbf{46.3} &
    \multicolumn{1}{|c|}{\underline{\textbf{86.7}}} & \underline{\textbf{46.3}} &
    \multicolumn{1}{|c|}{\underline{\textbf{86.7}}} & \underline{\textbf{46.3}} &
    \multicolumn{1}{|c|}{\underline{\textbf{86.7}}} & \underline{\textbf{46.3}} &
    \multicolumn{1}{|c|}{\underline{\textbf{86.7}}} & \underline{\textbf{46.3}}
       \\
  
  \underline{HOP} & \multicolumn{1}{|c|}{ } & \multicolumn{1}{c||}{ } & \multicolumn{1}{c|}{ } & \multicolumn{1}{c||}{ } & \multicolumn{1}{c|}{ } & \multicolumn{1}{c||}{ } & \multicolumn{1}{c|}{ } & \multicolumn{1}{c||}{ } & \multicolumn{1}{c|}{ } & \multicolumn{1}{c}{ } \\ 
  $\varepsilon$ = 0.1 &
    \multicolumn{1}{|c|}{\textbf{75.7}} & \textbf{69.0} &
    \multicolumn{1}{|c|}{\underline{\textbf{75.7}}} & \underline{\textbf{69.0}} &
    \multicolumn{1}{|c|}{\underline{\textbf{75.7}}} & \underline{\textbf{69.0}} &
    \multicolumn{1}{|c|}{\underline{\textbf{75.7}}} & \underline{\textbf{69.0}} &
    \multicolumn{1}{|c|}{\underline{\textbf{75.7}}} & \underline{\textbf{69.0}}
         \\ \hline
    L$_{2}$ Average & \multicolumn{1}{|c|}{74.9} & 65.8&
    \multicolumn{1}{|c|}{87.4} & 31.2 &
    \multicolumn{1}{|c|}{87.6} & 36.9 &
    \multicolumn{1}{|c|}{\textbf{\underline{73.7}}} & \textbf{\underline{67.2}} &
    \multicolumn{1}{|c|}{87.4} & 37.5 
    \\ \hline
  \midrule
  

\textbf{Norm L$_\infty$} &
  \multicolumn{1}{|c|}{\auc\%} &
  \fpr\% &
  \multicolumn{1}{|c|}{\auc\%} &
  \fpr\% &
  \multicolumn{1}{|c|}{\auc\%} &
  \fpr\% &
  \multicolumn{1}{|c|}{\auc\%} &
  \fpr\% &
  \multicolumn{1}{|c|}{\auc\%} &
  \fpr\%\\ \midrule
   
    \underline{PGDi, FGSM, BIM} & \multicolumn{1}{|c|}{ } & \multicolumn{1}{c||}{ } & \multicolumn{1}{c|}{ } & \multicolumn{1}{c||}{ } & \multicolumn{1}{c|}{ } & \multicolumn{1}{c||}{ } & \multicolumn{1}{c|}{ } & \multicolumn{1}{c||}{ } & \multicolumn{1}{c|}{ } & \multicolumn{1}{c}{ } \\
    $\varepsilon$ = 0.03125 &
    \multicolumn{1}{|c|}{\textbf{59.6}} & \textbf{79.0} &
    \multicolumn{1}{|c|}{75.2} & 63.1 &
    \multicolumn{1}{|c|}{78.5} & 57.9 &
    \multicolumn{1}{|c|}{\underline{66.1}} & \underline{73.6} &
    \multicolumn{1}{|c|}{77.4} & 61.3
       \\ 
    $\varepsilon$ = 0.0625 &
    \multicolumn{1}{|c|}{\textbf{53.7}} & \textbf{80.3} &
    \multicolumn{1}{|c|}{71.9} & 63.8 &
    \multicolumn{1}{|c|}{76.3} & 59.5 &
    \multicolumn{1}{|c|}{\underline{60.2}} & \underline{77.1} &
    \multicolumn{1}{|c|}{77.2} & 57.6
       \\ 
    $\varepsilon$ = 0.25 &
    \multicolumn{1}{|c|}{\textbf{49.4}} & \textbf{83.2} &
    \multicolumn{1}{|c|}{72.8} & 60.0 &
    \multicolumn{1}{|c|}{78.2} & 53.9 &
    \multicolumn{1}{|c|}{\underline{55.2}} & \underline{81.6} &
    \multicolumn{1}{|c|}{76.8} & 53.8
       \\ 
    $\varepsilon$ = 0.5 &
    \multicolumn{1}{|c|}{\textbf{54.1}} & \textbf{78.8} &
    \multicolumn{1}{|c|}{82.7} & 39.9 &
    \multicolumn{1}{|c|}{86.4} & 36.5 &
    \multicolumn{1}{|c|}{\underline{57.2}} & \underline{78.2} &
    \multicolumn{1}{|c|}{78.6} & 52.3
       \\ 
  

 \underline{PGDi, FGSM, BIM, SA} & \multicolumn{1}{|c|}{ } & \multicolumn{1}{c||}{ } & \multicolumn{1}{c|}{ } & \multicolumn{1}{c||}{ } & \multicolumn{1}{c|}{ } & \multicolumn{1}{c||}{ } & \multicolumn{1}{c|}{ } & \multicolumn{1}{c||}{ } & \multicolumn{1}{c|}{ } & \multicolumn{1}{c}{ }\\ 
    $\varepsilon$ = 0.125 &
\multicolumn{1}{|c|}{\textbf{46.5}} & \textbf{86.3} &
\multicolumn{1}{|c|}{64.3} & 71.6 &
\multicolumn{1}{|c|}{69.7} & 68.1 &
\multicolumn{1}{|c|}{\underline{51.5}} & \underline{85.2} &
\multicolumn{1}{|c|}{70.8} & 66.3
   \\

 \underline{PGDi, FGSM, BIM, CWi}  & \multicolumn{1}{|c|}{ } & \multicolumn{1}{c||}{ } & \multicolumn{1}{c|}{ } & \multicolumn{1}{c||}{ } & \multicolumn{1}{c|}{ } & \multicolumn{1}{c||}{ } & \multicolumn{1}{c|}{ } & \multicolumn{1}{c||}{ } & \multicolumn{1}{c|}{ } & \multicolumn{1}{c}{ }\\ 
$\varepsilon$ = 0.3125 &
\multicolumn{1}{|c|}{\textbf{52.7}} & \textbf{79.3} &
\multicolumn{1}{|c|}{71.1} & 62.2 &
\multicolumn{1}{|c|}{75.7} & 58.2 &
\multicolumn{1}{|c|}{\underline{58.9}} & \underline{77.2} &
\multicolumn{1}{|c|}{73.4} & 59.9
    \\ \hline
L$_{\infty}$ Average & \multicolumn{1}{|c|}{\textbf{52.7}} & \textbf{81.1} &
    \multicolumn{1}{|c|}{73.0} & 60.1 &
    \multicolumn{1}{|c|}{77.5} & 55.7 &
    \multicolumn{1}{|c|}{\underline{58.2}} & \underline{78.8} &
    \multicolumn{1}{|c|}{75.7} & 58.5
    \\ \hline
  \midrule
  
  \textbf{No norm} &
  \multicolumn{1}{|c|}{\auc\%} &
  \fpr\% &
  \multicolumn{1}{|c|}{\auc\%} &
  \fpr\% &
  \multicolumn{1}{|c|}{\auc\%} &
  \fpr\% &
  \multicolumn{1}{|c|}{\auc\%} &
  \fpr\% &
  \multicolumn{1}{|c|}{\auc\%} &
  \fpr\%\\ \midrule
  \underline{STA}  & \multicolumn{1}{|c|}{ } & \multicolumn{1}{c||}{ } & \multicolumn{1}{c|}{ } & \multicolumn{1}{c||}{ } & \multicolumn{1}{c|}{ } & \multicolumn{1}{c||}{ } & \multicolumn{1}{c|}{ } & \multicolumn{1}{c||}{ } & \multicolumn{1}{c|}{ } & \multicolumn{1}{c}{ }\\
No $\varepsilon$ & 
\multicolumn{1}{|c|}{\textbf{62.7}} & \textbf{82.5} &
\multicolumn{1}{|c|}{\underline{\textbf{62.7}}} & \underline{\textbf{82.5}} &
\multicolumn{1}{|c|}{\underline{\textbf{62.7}}} & \underline{\textbf{82.5}} &
\multicolumn{1}{|c|}{\underline{\textbf{62.7}}} & \underline{\textbf{82.5}} &
\multicolumn{1}{|c|}{\underline{\textbf{62.7}}} & \underline{\textbf{82.5}}
   \\ \hline
No norm Average & \multicolumn{1}{|c|}{\textbf{62.7}} & \textbf{82.5} &
    \multicolumn{1}{|c|}{\underline{\textbf{62.7}}} & \underline{\textbf{82.5}} &
    \multicolumn{1}{|c|}{\underline{\textbf{62.7}}} & \underline{\textbf{82.5}} &
    \multicolumn{1}{|c|}{\underline{\textbf{62.7}}} & \underline{\textbf{82.5}} &
    \multicolumn{1}{|c|}{\underline{\textbf{62.7}}} & \underline{\textbf{82.5}}
    \\ \hline
  \bottomrule
\end{tabular}
}
\caption{Performances on \method{FS} per objective and in \mead~on CIFAR10. The worst results among all the settings 
is in \textbf{bold}; the ones in the single-armed setting is \underline{underlined}. No norm denotes the group of attacks that do not depend on the norm constraint.}
\label{tab:cifar_fs_per_loss}
\end{table*}
\method{FS} is not quite affected by the \mead~framework under the L$_{1}$ and L$_{2}$-norm constraints. The reason why is explained in \cref{sec:results} in the remark of the paragraph called \textbf{\mead~and the single-armed setting}. However, under the L$_{\infty}$-norm constraint, our \mead~framework is quite damaging, creating a decrease in terms of \auc up to 6.5 percentage points and a maximal increase in \fpr of 5.3 percentage points. Under the single-armed setting, \method{FS} is extremely sensitive to the attacks generated by maximizing the Gini Impurity score. 

\newpage
\subsubsection{\method{MagNet}}\label{appendix:cifar-magnet}
In~\cref{tab:cifar_magnet_per_loss}, we show the result of our \mead~framework on CIFAR10, evaluated on \method{MagNet}.
\begin{table*}[!htb]
\centering
\resizebox{\columnwidth}{!}{%
\begin{tabular}{c||cc||cc||cc||cc||cc}
\toprule 
    \textbf{\method{MagNet}} &
  \multicolumn{2}{c||}{\mead} & \multicolumn{2}{c||}{ACE}& \multicolumn{2}{c||}{KL}& \multicolumn{2}{c||}{Gini} &
  \multicolumn{2}{c}{FR}
   \\ \hline\midrule
  \textbf{Norm L$_1$}&
  \multicolumn{1}{|c|}{\auc\%} &
   {\fpr\% } &
  \multicolumn{1}{c|}{\auc\%} &
   \fpr\% 
   &
  \multicolumn{1}{c|}{\auc\%} &
  \fpr\% &
  \multicolumn{1}{c|}{\auc\%} &
   \fpr\% &
  \multicolumn{1}{c|}{\auc\%} &
   \fpr\% \\ \midrule
   \underline{PGD1}  & \multicolumn{1}{|c|}{ } & \multicolumn{1}{c||}{ } & \multicolumn{1}{c|}{ } & \multicolumn{1}{c||}{ } & \multicolumn{1}{c|}{ } & \multicolumn{1}{c||}{ } & \multicolumn{1}{c|}{ } & \multicolumn{1}{c||}{ } & \multicolumn{1}{c|}{ } & \multicolumn{1}{c}{ } \\
 $\varepsilon$ = 5 &
\multicolumn{1}{|c|}{43.9} & 96.6 &
\multicolumn{1}{|c|}{43.6} & 96.6 &
\multicolumn{1}{|c|}{43.7} & 96.5 &
\multicolumn{1}{|c|}{\textbf{\underline{43.3}}} & \textbf{\underline{97.1}} &
\multicolumn{1}{|c|}{43.6} & 96.7
   \\
$\varepsilon$ = 10 &
\multicolumn{1}{|c|}{47.8} & 95.2 &
\multicolumn{1}{|c|}{47.9} & 95.1 &
\multicolumn{1}{|c|}{47.6} & 94.9 &
\multicolumn{1}{|c|}{\textbf{46.7}} & \textbf{\underline{95.6}} &
\multicolumn{1}{|c|}{\textbf{46.7}} & \textbf{\underline{95.6}}
   \\
$\varepsilon$ = 15 &
\multicolumn{1}{|c|}{49.8} & 94.3 &
\multicolumn{1}{|c|}{50.0} & 94.3 &
\multicolumn{1}{|c|}{49.8} & 94.1 &
\multicolumn{1}{|c|}{\textbf{\underline{49.1}}} & 94.5 &
\multicolumn{1}{|c|}{\textbf{\underline{49.1}}} & \textbf{\underline{94.7}}
   \\ 
$\varepsilon$ = 20 &
\multicolumn{1}{|c|}{50.6} & 93.8 &
\multicolumn{1}{|c|}{50.8} & 93.7 &
\multicolumn{1}{|c|}{50.6} & 93.5 &
\multicolumn{1}{|c|}{50.7} & 93.3 &
\multicolumn{1}{|c|}{\textbf{\underline{49.8}}} & \textbf{\underline{94.1}}
   \\ 
$\varepsilon$ = 25 &
\multicolumn{1}{|c|}{51.1} & 93.1 &
\multicolumn{1}{|c|}{51.4} & 93.0 &
\multicolumn{1}{|c|}{51.1} & 92.8 &
\multicolumn{1}{|c|}{52.5} & 91.7 &
\multicolumn{1}{|c|}{\textbf{\underline{50.6}}} & \textbf{\underline{93.3}}
   \\ 
$\varepsilon$ = 30 &
\multicolumn{1}{|c|}{51.6} & 92.4 &
\multicolumn{1}{|c|}{51.9} & 92.1 &
\multicolumn{1}{|c|}{51.7} & 91.8 &
\multicolumn{1}{|c|}{53.7} & 90.2 &
\multicolumn{1}{|c|}{\textbf{\underline{51.2}}} & \textbf{\underline{92.5}}
   \\ 
$\varepsilon$ = 40 &
\multicolumn{1}{|c|}{\textbf{52.7}} & \textbf{90.6} &
\multicolumn{1}{|c|}{53.3} & 89.6 &
\multicolumn{1}{|c|}{53.1} & 89.2 &
\multicolumn{1}{|c|}{54.5} & 89.7 &
\multicolumn{1}{|c|}{\textbf{\underline{52.7}}} & \underline{89.8}
   \\ \hline

L$_{1}$ Average & \multicolumn{1}{|c|}{49.6} & 93.7 &
    \multicolumn{1}{|c|}{49.8} & 93.5 &
    \multicolumn{1}{|c|}{49.7} & 93.3 &
    \multicolumn{1}{|c|}{50.1} & 93.2 &
    \multicolumn{1}{|c|}{\textbf{\underline{49.1}}} & \textbf{\underline{93.8}}
        \\ \hline
 \midrule
  
  
\textbf{Norm L$_2$} &
  \multicolumn{1}{|c|}{\auc\%} &
   \fpr\% &
  \multicolumn{1}{|c|}{\auc\%} &
   \fpr\% &
  \multicolumn{1}{|c|}{\auc\%} &
   \fpr\% &
  \multicolumn{1}{|c|}{\auc\%} &
   \fpr\% &
  \multicolumn{1}{|c|}{\auc\%} &
   \fpr\%\\ \midrule
   \underline{PGD2}  & \multicolumn{1}{|c|}{ } & \multicolumn{1}{c||}{ } & \multicolumn{1}{c|}{ } & \multicolumn{1}{c||}{ } & \multicolumn{1}{c|}{ } & \multicolumn{1}{c||}{ } & \multicolumn{1}{c|}{ } & \multicolumn{1}{c||}{ } & \multicolumn{1}{c|}{ } & \multicolumn{1}{c}{ } \\
   $\varepsilon$ = 0.125 &
\multicolumn{1}{|c|}{43.1} & 96.9 &
\multicolumn{1}{|c|}{\textbf{\underline{42.6}}} & 96.9 &
\multicolumn{1}{|c|}{43.2} & 96.8 &
\multicolumn{1}{|c|}{43.8} & 96.7 &
\multicolumn{1}{|c|}{43.2} & \textbf{\underline{97.0}}
   \\ 
$\varepsilon$ = 0.25 &
\multicolumn{1}{|c|}{47.3} & 95.4 &
\multicolumn{1}{|c|}{47.3} & 95.4 &
\multicolumn{1}{|c|}{47.1} & 95.1 &
\multicolumn{1}{|c|}{\textbf{\underline{45.6}}} & \textbf{\underline{95.7}} &
\multicolumn{1}{|c|}{46.4} & \textbf{\underline{95.7}}
   \\ 
$\varepsilon$ = 0.3125 &
\multicolumn{1}{|c|}{48.7} & 94.8 &
\multicolumn{1}{|c|}{48.8} & 94.7 &
\multicolumn{1}{|c|}{48.7} & 94.5 &
\multicolumn{1}{|c|}{\textbf{\underline{47.4}}} & 95.0 &
\multicolumn{1}{|c|}{48.0} & \textbf{\underline{95.1}}
   \\ 
$\varepsilon$ = 0.5 &
\multicolumn{1}{|c|}{50.6} & 93.8 &
\multicolumn{1}{|c|}{50.7} & 93.6 &
\multicolumn{1}{|c|}{50.5} & 93.4 &
\multicolumn{1}{|c|}{50.5} & 93.3 &
\multicolumn{1}{|c|}{\textbf{\underline{49.9}}} & \textbf{\underline{94.2}}
   \\ 
$\varepsilon$ = 1 &
\multicolumn{1}{|c|}{\textbf{53.0}} & \textbf{90.1} &
\multicolumn{1}{|c|}{53.7} & 88.7 &
\multicolumn{1}{|c|}{53.6} & 88.4 &
\multicolumn{1}{|c|}{54.4} & \underline{89.6} &
\multicolumn{1}{|c|}{\underline{53.3}} & 88.8
   \\ 
$\varepsilon$ = 1.5 &
\multicolumn{1}{|c|}{55.3} & 88.3 &
\multicolumn{1}{|c|}{59.3} & 79.2 &
\multicolumn{1}{|c|}{59.1} & 78.8 &
\multicolumn{1}{|c|}{\textbf{\underline{54.5}}} & \textbf{\underline{89.5}} &
\multicolumn{1}{|c|}{59.2} & 80.9
   \\
$\varepsilon$ = 2 &
\multicolumn{1}{|c|}{56.9} & 88.2 &
\multicolumn{1}{|c|}{67.2} & 64.4 &
\multicolumn{1}{|c|}{67.2} & 64.0 &
\multicolumn{1}{|c|}{\textbf{\underline{54.5}}} & \textbf{\underline{89.5}} &
\multicolumn{1}{|c|}{63.7} & 79.3
   \\

   \underline{DeepFool} & \multicolumn{1}{|c|}{ } & \multicolumn{1}{c||}{ } & \multicolumn{1}{c|}{ } & \multicolumn{1}{c||}{ } & \multicolumn{1}{c|}{ } & \multicolumn{1}{c||}{ } & \multicolumn{1}{c|}{ } & \multicolumn{1}{c||}{ } & \multicolumn{1}{c|}{ } & \multicolumn{1}{c}{ } \\
No $\varepsilon$  & 
    \multicolumn{1}{|c|}{\textbf{51.1}} & \textbf{94.7} &
\multicolumn{1}{|c|}{\textbf{\underline{51.1}}} &\textbf{\underline{94.7}} &
\multicolumn{1}{|c|}{\textbf{\underline{51.1}}} &\textbf{\underline{94.7}} &\multicolumn{1}{|c|}{\textbf{\underline{51.1}}} &\textbf{\underline{94.7}} &
\multicolumn{1}{|c|}{\textbf{\underline{51.1}}} &\textbf{\underline{94.7}} 
   \\

   \underline{CW2}  & \multicolumn{1}{|c|}{ } & \multicolumn{1}{c||}{ } & \multicolumn{1}{c|}{ } & \multicolumn{1}{c||}{ } & \multicolumn{1}{c|}{ } & \multicolumn{1}{c||}{ } & \multicolumn{1}{c|}{ } & \multicolumn{1}{c||}{ } & \multicolumn{1}{c|}{ } & \multicolumn{1}{c}{ } \\
$\varepsilon$ = 0.01 &
\multicolumn{1}{|c|}{\textbf{50.5}} & \textbf{94.7} &
\multicolumn{1}{|c|}{\textbf{\underline{50.5}}} & \textbf{\underline{94.7}} &
\multicolumn{1}{|c|}{\textbf{\underline{50.5}}} & \textbf{\underline{94.7}} &
\multicolumn{1}{|c|}{\textbf{\underline{50.5}}} & \textbf{\underline{94.7}} &
\multicolumn{1}{|c|}{\textbf{\underline{50.5}}} & \textbf{\underline{94.7}}  
   \\ 
  
   \underline{HOP} & \multicolumn{1}{|c|}{ } & \multicolumn{1}{c||}{ } & \multicolumn{1}{c|}{ } & \multicolumn{1}{c||}{ } & \multicolumn{1}{c|}{ } & \multicolumn{1}{c||}{ } & \multicolumn{1}{c|}{ } & \multicolumn{1}{c||}{ } & \multicolumn{1}{c|}{ } & \multicolumn{1}{c}{ } \\
$\varepsilon$ = 0.1 &
\multicolumn{1}{|c|}{\textbf{52.2}} & \textbf{93.8} &
\multicolumn{1}{|c|}{\textbf{\underline{52.2}}} & \textbf{\underline{93.8}} &
\multicolumn{1}{|c|}{\textbf{\underline{52.2}}} & \textbf{\underline{93.8}} &
\multicolumn{1}{|c|}{\textbf{\underline{52.2}}} & \textbf{\underline{93.8}} &
\multicolumn{1}{|c|}{\textbf{\underline{52.2}}} & \textbf{\underline{93.8}} 
   \\ \hline

L$_{2}$ Average & \multicolumn{1}{|c|}{50.9} & 93.1  &
    \multicolumn{1}{|c|}{52.3} & 89.6 &
    \multicolumn{1}{|c|}{52.3} & 89.4 &
    \multicolumn{1}{|c|}{\textbf{\underline{50.5}}} & \textbf{\underline{93.3}} &
    \multicolumn{1}{|c|}{51.8} & 91.4 
        \\ \hline\midrule
  

\textbf{Norm L$_\infty$}  &
  \multicolumn{1}{|c|}{\auc\%} &
   \fpr\% &
  \multicolumn{1}{|c|}{\auc\%} &
   \fpr\% &
  \multicolumn{1}{|c|}{\auc\%} &
   \fpr\% &
  \multicolumn{1}{|c|}{\auc\%} &
   \fpr\% &
  \multicolumn{1}{|c|}{\auc\%} &
   \fpr\%\\ \midrule
   
   \underline{PGDi, FGSM, BIM}  & \multicolumn{1}{|c|}{ } & \multicolumn{1}{c||}{ } & \multicolumn{1}{c|}{ } & \multicolumn{1}{c||}{ } & \multicolumn{1}{c|}{ } & \multicolumn{1}{c||}{ } & \multicolumn{1}{c|}{ } & \multicolumn{1}{c||}{ } & \multicolumn{1}{c|}{ } & \multicolumn{1}{c}{ } \\
    $\varepsilon$ = 0.03125 &
\multicolumn{1}{|c|}{\textbf{58.6}} & \textbf{82.0} &
\multicolumn{1}{|c|}{60.0} & 80.0 &
\multicolumn{1}{|c|}{60.4} & 79.0 &
\multicolumn{1}{|c|}{60.4} & 80.8 &
\multicolumn{1}{|c|}{\underline{59.0}} & \underline{81.0}
   \\ 
$\varepsilon$ = 0.0625 &
\multicolumn{1}{|c|}{\textbf{74.6}} & \textbf{51.2} &
\multicolumn{1}{|c|}{\underline{76.8}} & 48.0 &
\multicolumn{1}{|c|}{79.4} & 46.2 &
\multicolumn{1}{|c|}{77.1} & \underline{48.2} &
\multicolumn{1}{|c|}{78.8} & 47.0
   \\
$\varepsilon$ = 0.25 &
\multicolumn{1}{|c|}{\textbf{97.0}} & \textbf{5.2} &
\multicolumn{1}{|c|}{98.2} & 3.6 &
\multicolumn{1}{|c|}{98.7} & 3.4 &
\multicolumn{1}{|c|}{\underline{97.7}} & \underline{4.1} &
\multicolumn{1}{|c|}{98.7} & 3.4
   \\ 
$\varepsilon$ = 0.5 &
\multicolumn{1}{|c|}{\textbf{98.0}} & \textbf{3.5} &
\multicolumn{1}{|c|}{99.0} & 2.2 &
\multicolumn{1}{|c|}{99.2} & 2.0 &
\multicolumn{1}{|c|}{\underline{98.6}} & \underline{2.6} &
\multicolumn{1}{|c|}{99.2} & 2.1
   \\ 
  
 \underline{PGDi, FGSM, BIM, SA}  & \multicolumn{1}{|c|}{ } & \multicolumn{1}{c||}{ } & \multicolumn{1}{c|}{ } & \multicolumn{1}{c||}{ } & \multicolumn{1}{c|}{ } & \multicolumn{1}{c||}{ } & \multicolumn{1}{c|}{ } & \multicolumn{1}{c||}{ } & \multicolumn{1}{c|}{ } & \multicolumn{1}{c}{ } \\
$\varepsilon$ = 0.125 &
\multicolumn{1}{|c|}{\textbf{87.0}} & \textbf{40.0} &
\multicolumn{1}{|c|}{\underline{88.8}} & \underline{39.3} &
\multicolumn{1}{|c|}{90.9} & \underline{39.3} &
\multicolumn{1}{|c|}{88.9} & 37.3 &
\multicolumn{1}{|c|}{91.4} & \underline{39.3}
   \\

  
 \underline{PGDi, FGSM, BIM, CWi}   & \multicolumn{1}{|c|}{ } & \multicolumn{1}{c||}{ } & \multicolumn{1}{c|}{ } & \multicolumn{1}{c||}{ } & \multicolumn{1}{c|}{ } & \multicolumn{1}{c||}{ } & \multicolumn{1}{c|}{ } & \multicolumn{1}{c||}{ } & \multicolumn{1}{c|}{ } & \multicolumn{1}{c}{ } \\
$\varepsilon$ = 0.3125 &
\multicolumn{1}{|c|}{52.6} & \textbf{94.5} &
\multicolumn{1}{|c|}{\textbf{\underline{52.5}}} & \textbf{\underline{94.5}} &
\multicolumn{1}{|c|}{\textbf{\underline{52.5}}} & \textbf{\underline{94.5}} &
\multicolumn{1}{|c|}{52.6} & \textbf{\underline{94.5}} &
\multicolumn{1}{|c|}{52.6} & \textbf{\underline{94.5}}
   \\ \hline

L$_{\infty}$ Average & 
    \multicolumn{1}{|c|}{\textbf{78.0}} & \textbf{46.1} &
    \multicolumn{1}{|c|}{\underline{79.2}} & \underline{44.6} &
    \multicolumn{1}{|c|}{80.2} & 44.1 &
    \multicolumn{1}{|c|}{\underline{79.2}} & \underline{44.6} &
    \multicolumn{1}{|c|}{80.0} & \underline{44.6}
        \\ \hline
  \midrule
  
\textbf{No norm} &
  \multicolumn{1}{|c|}{\auc\%} &
   \fpr\% &
  \multicolumn{1}{|c|}{\auc\%} &
   \fpr\% &
  \multicolumn{1}{|c|}{\auc\%} &
   \fpr\% &
  \multicolumn{1}{|c|}{\auc\%} &
   \fpr\% &
  \multicolumn{1}{|c|}{\auc\%} &
   \fpr\%\\ \midrule
   
   \underline{STA}  & \multicolumn{1}{|c|}{ } & \multicolumn{1}{c||}{ } & \multicolumn{1}{c|}{ } & \multicolumn{1}{c||}{ } & \multicolumn{1}{c|}{ } & \multicolumn{1}{c||}{ } & \multicolumn{1}{c|}{ } & \multicolumn{1}{c||}{ } & \multicolumn{1}{c|}{ } & \multicolumn{1}{c}{ } \\
No $\varepsilon$ & 
\multicolumn{1}{|c|}{\textbf{79.9}} & \textbf{45.7} &
\multicolumn{1}{|c|}{\textbf{\underline{79.9}}} &\textbf{ \underline{45.7} }&
\multicolumn{1}{|c|}{\textbf{\underline{79.9}}} &\textbf{ \underline{45.7} }&
\multicolumn{1}{|c|}{\textbf{\underline{79.9}}} &\textbf{ \underline{45.7} }&
\multicolumn{1}{|c|}{\textbf{\underline{79.9}}} &\textbf{ \underline{45.7} }
   \\ \hline
No norm Average & \multicolumn{1}{|c|}{\textbf{79.9}} & \textbf{45.7} &
\multicolumn{1}{|c|}{\textbf{\underline{79.9}}} &\textbf{ \underline{45.7} }&
\multicolumn{1}{|c|}{\textbf{\underline{79.9}}} &\textbf{ \underline{45.7} }&
\multicolumn{1}{|c|}{\textbf{\underline{79.9}}} &\textbf{ \underline{45.7} }&
\multicolumn{1}{|c|}{\textbf{\underline{79.9}}} &\textbf{ \underline{45.7} }
    \\ \hline
   \bottomrule
\end{tabular}
}
\caption{Performances on \method{MagNet} per objective and in \mead~on CIFAR10. The worst results among all the settings 
are shown in \textbf{bold}; the ones in the single-armed setting is \underline{underlined}. No norm denotes the group of attacks that do not depend on the norm constraint.}
\label{tab:cifar_magnet_per_loss}
\end{table*}

\method{MagNet} is an unsupervised detection method. In most cases, on CIFAR10, the results using \mead~are close to the worst results for the single-armed settings. In other words, it seems that if an example generated using the worst loss is detected (usually the Fisher-Rao objective), then the samples generated using all the others are detected. The decrease in \auc~between the worst single-armed setting and \mead~is up to 1.8 percentage points.
\newpage
\subsection{Additional Results on MNIST}\label{appendix:mnist_additional_results}
\subsubsection{Success of attacks}\label{appendix:attack_mnist}
In~\cref{tab:mnist_attack_rate}, we show the average and total numbers of successful attack per settings (\mead, Adversarial Cross-Entropy, KL divergence, Gini Impurity score and Fisher-Rao loss) on the MNIST dataset. 
\begin{table}[h]
\centering
\resizebox{0.8\columnwidth}{!}{%
\setlength{\tabcolsep}{10pt}
\begin{tabular}{c||c||c||c||c||c}
\toprule

   \multicolumn{6}{c}{\textbf{Avg. Num. of Successful Attack / Tot. Num. of Attack }}
      \\ \cline{1-6} 
    \textbf{Norm L$_1$} &
  \multicolumn{1}{c||}{\mead} & \multicolumn{1}{c||}{ACE}& \multicolumn{1}{c||}{KL}& \multicolumn{1}{c||}{Gini} &
  \multicolumn{1}{c}{FR}
       \\ \midrule
     \underline{PGD1}  &\multicolumn{1}{c||}{ } & \multicolumn{1}{c||}{ } &  \multicolumn{1}{c||}{ } &  \multicolumn{1}{c||}{ } &  \multicolumn{1}{c}{ } \\
 $\varepsilon$ = 5 & 
   0.06 / 4 & 
   0.02 / 1 &          
   0.02 / 1 &     
   0.01 / 1 & 
   0.01 / 1   
     \\
 $\varepsilon$ = 10 & 
   0.23 / 4  
   & 0.09 / 1
   & 0.06 / 1
   & 0.03 / 1
   & 0.05 / 1
     \\ 
 $\varepsilon$ = 15 & 
   0.77 / 4 & 
   0.31 / 1 & 
   0.20 / 1 &
   0.10 / 1 & 
   0.16 / 1 
     \\
 $\varepsilon$ = 20 & 
   1.50 / 4 &    
   0.58 / 1 & 
   0.38 / 1 & 
   0.22 / 1& 
   0.33 /1 
     \\
 $\varepsilon$ = 25 & 
   2.03 / 4 &   
   0.73 / 1&
   0.48 / 1& 
   0.33 / 1& 
   0.48 / 1
     \\
 $\varepsilon$ = 30 & 
    2.35 / 4&   
    0.80 / 1 &
    0.54 / 1& 
    0.42 / 1& 
    0.59 / 1  
     \\ 
 $\varepsilon$ = 40 & 
   2.67 / 4&    
   0.85 / 1& 
   0.58 / 1& 
   0.54 / 1& 
   0.70 / 1
     \\ \hline\midrule
  
  
\textbf{Norm L$_2$} \\ \midrule
     \underline{PGD2}  &\multicolumn{1}{c||}{ } & \multicolumn{1}{c||}{ } &  \multicolumn{1}{c||}{ } &  \multicolumn{1}{c||}{ } &  \multicolumn{1}{c}{ } \\
$\varepsilon$ = 0.125 & 
    0.04 / 4&    
    0.01 / 1& 
    0.01 / 1& 
    0.01 / 1& 
    0.01 / 1
     \\ 
$\varepsilon$ = 0.25 & 
    0.04 / 4&    
    0.01 / 1& 
    0.01 / 1& 
    0.01 / 1& 
    0.01 / 1
     \\
$\varepsilon$ = 0.3125 & 
    0.05 / 4&    
    0.01 / 1& 
    0.01 / 1& 
    0.01 / 1& 
    0.01 / 1
     \\ 
$\varepsilon$ = 0.5 & 
    0.07 / 4&    
    0.02 / 1& 
    0.02 / 1& 
    0.01 / 1& 
    0.02 / 1    
     \\
$\varepsilon$ = 1 & 
    0.29 / 4&    
    0.12 / 1& 
    0.08 / 1& 
    0.04 / 1& 
    0.06 / 1
     \\ 
$\varepsilon$ = 1.5 & 
    0.99 / 4&    
    0.40 / 1& 
    0.26 / 1& 
    0.13 / 1& 
    0.21 / 1 
     \\ 
$\varepsilon$ = 2 & 
    1.75 / 4&    
    0.63 / 1& 
    0.44 / 1& 
    0.28 / 1& 
    0.40 / 1
     \\

   \underline{DeepFool}   &\multicolumn{1}{c||}{ } & \multicolumn{1}{c||}{ } &  \multicolumn{1}{c||}{ } &  \multicolumn{1}{c||}{ } &  \multicolumn{1}{c}{ } \\
No $\varepsilon$ & 
    0.97 / 1&    
    0.97 / 1& 
    0.97 / 1& 
    0.97 / 1& 
    0.97 / 1
     \\

   \underline{CW2} &\multicolumn{1}{c||}{ } & \multicolumn{1}{c||}{ } &  \multicolumn{1}{c||}{ } &  \multicolumn{1}{c||}{ } &  \multicolumn{1}{c}{ } \\
$\varepsilon$ = 0.01 & 
    0.74 / 1&    
    0.74 / 1& 
    0.74 / 1& 
    0.74 / 1& 
    0.74 / 1
     \\

   \underline{HOP} &\multicolumn{1}{c||}{ } & \multicolumn{1}{c||}{ } &  \multicolumn{1}{c||}{ } &  \multicolumn{1}{c||}{ } &  \multicolumn{1}{c}{ } \\
$\varepsilon$ = 0.1 & 
    0.99 / 1&    
    0.99 / 1& 
    0.99 / 1& 
    0.99 / 1& 
    0.99 / 1
     \\  \hline\midrule
  

\textbf{Norm L$_\infty$} \\ \midrule
 
      \underline{PGDi, FGSM, BIM}  &\multicolumn{1}{c||}{ } & \multicolumn{1}{c||}{ } &  \multicolumn{1}{c||}{ } &  \multicolumn{1}{c||}{ } &  \multicolumn{1}{c}{ } \\
$\varepsilon$ = 0.03125 & 
    0.14 / 12&    
    0.04 / 3& 
    0.03 / 3& 
    0.04 / 3& 
    0.03 / 3
     \\ 
$\varepsilon$ = 0.0625 & 
    0.38 / 12&    
    0.13 / 3& 
    0.08 / 3& 
    0.09 / 3& 
    0.07 / 3
     \\ 
$\varepsilon$ = 0.125 & 
    2.07 / 12 & 
    0.79 / 3& 
    0.46 / 3&
    0.45 / 3& 
    0.37 / 3   \\ 
$\varepsilon$ = 0.25& 
    7.41 / 12&    
    2.62 / 3& 
    1.50 / 3& 
    1.70/ 3& 
    1.60 / 3  
     \\ 
$\varepsilon$ = 0.5 & 
    8.85 / 12&    
    2.90 / 3& 
    1.76 / 3& 
    2.20 / 3& 
    1.99 / 3
     \\


 \underline{PGDi, FGSM, BIM, CWi, SA} &\multicolumn{1}{c||}{ } & \multicolumn{1}{c||}{ } &  \multicolumn{1}{c||}{ } &  \multicolumn{1}{c||}{ } &  \multicolumn{1}{c}{ } \\
$\varepsilon$ = 0.3125 & 
    9.73 / 14 & 
    4.34 / 5& 
    3.17 / 5&
    3.51 / 5& 
    3.34 / 5
     \\ \hline
  \midrule
  
\textbf{No norm} \\ \midrule
        \underline{STA}  &\multicolumn{1}{c||}{ } & \multicolumn{1}{c||}{ } &  \multicolumn{1}{c||}{ } &  \multicolumn{1}{c||}{ } &  \multicolumn{1}{c}{ } \\
No $\varepsilon$ & 
     0.85 / 1& 
     0.85 / 1& 
     0.85 / 1& 
     0.85 / 1& 
     0.85 / 1
     \\ \hline
   \bottomrule
\end{tabular}
}
\caption{Average number of successful attacks per natural sample considered in the single-armed setting and \mead~(MNIST). The results are reported in the table together with the total number of attacks performed per natural sample (\textit{Avg. / Tot.}). No norm denotes the group of attacks that do not depend on the norm constraint.
}
\label{tab:mnist_attack_rate}
\end{table}

We can observe the same behavior in MNIST as in CIFAR10. The most harmful attacks for the classifier are the ones generated according to the Adversarial Cross-Entropy. The attacks generated thanks to the three other objectives have a similar strength. 
\newpage
\subsubsection{\method{NSS}}\label{appendix:mnist-nss}
In~\cref{tab:mnist_nss_per_loss}, we show the result of our \mead~framework on MNIST, evaluated on \method{NSS}.
\begin{table*}[!htb]
\centering
\resizebox{\columnwidth}{!}{%
\begin{tabular}{c||cc||cc||cc||cc||cc}
\toprule 
  
    \textbf{\method{NSS}} &
  \multicolumn{2}{c||}{\mead} & \multicolumn{2}{c||}{ACE}& \multicolumn{2}{c||}{KL}& \multicolumn{2}{c||}{Gini} &
  \multicolumn{2}{c}{FR}
  \\ \hline\midrule
  \textbf{Norm L$_1$}&
  \multicolumn{1}{|c|}{\auc\%} &
  {\fpr\% } &
  \multicolumn{1}{c|}{\auc\%} &
  \fpr\% 
  &
  \multicolumn{1}{c|}{\auc\%} &
  \fpr\% &
  \multicolumn{1}{c|}{\auc\%} &
  \fpr\% &
  \multicolumn{1}{c|}{\auc\%} &
  \fpr\% \\ \midrule
  \underline{PGD1}  & \multicolumn{1}{|c|}{ } & \multicolumn{1}{c||}{ } & \multicolumn{1}{c|}{ } & \multicolumn{1}{c||}{ } & \multicolumn{1}{c|}{ } & \multicolumn{1}{c||}{ } & \multicolumn{1}{c|}{ } & \multicolumn{1}{c||}{ } & \multicolumn{1}{c|}{ } & \multicolumn{1}{c}{ } \\
  $\varepsilon$ = 5 &
\multicolumn{1}{|c|}{\textbf{91.4}} & \textbf{30.5} &
\multicolumn{1}{|c|}{92.1} & 24.6 &
\multicolumn{1}{|c|}{92.3} & \underline{27.2} &
\multicolumn{1}{|c|}{93.0} & 24.2 &
\multicolumn{1}{|c|}{\underline{92.0}} & 26.9
 \\
$\varepsilon$ = 10 &
\multicolumn{1}{|c|}{\textbf{96.4}} & \textbf{11.7} &
\multicolumn{1}{|c|}{\underline{96.6}} & \underline{10.8} &
\multicolumn{1}{|c|}{96.7} & 10.7 &
\multicolumn{1}{|c|}{97.3} & 8.0 &
\multicolumn{1}{|c|}{97.1} & 8.5
 \\
$\varepsilon$ = 15 &
\multicolumn{1}{|c|}{\textbf{97.3}} & \textbf{8.6} &
\multicolumn{1}{|c|}{\underline{97.5}} & \underline{8.0} &
\multicolumn{1}{|c|}{\underline{97.5}} & \underline{8.0} &
\multicolumn{1}{|c|}{98.0} & 4.2 &
\multicolumn{1}{|c|}{\underline{97.5}} & 7.7
 \\
$\varepsilon$ = 20 &
\multicolumn{1}{|c|}{\textbf{97.9}} & \textbf{5.3} &
\multicolumn{1}{|c|}{\underline{98.0}} & \underline{4.7} &
\multicolumn{1}{|c|}{\underline{98.0}} & 4.6 &
\multicolumn{1}{|c|}{98.2} & 3.3 &
\multicolumn{1}{|c|}{98.1} & 3.9
 \\
$\varepsilon$ = 25 &
\multicolumn{1}{|c|}{\textbf{98.2}} & \textbf{3.2} &
\multicolumn{1}{|c|}{\underline{98.3}} & \underline{\textbf{3.2}} &
\multicolumn{1}{|c|}{\underline{98.3}} & \underline{\textbf{3.2}} &
\multicolumn{1}{|c|}{\underline{98.3}} & 3.1 &
\multicolumn{1}{|c|}{\underline{98.3}} & \underline{\textbf{3.2}}
 \\
$\varepsilon$ = 30 &
\multicolumn{1}{|c|}{\textbf{98.3}} & \textbf{3.1} &
\multicolumn{1}{|c|}{\underline{\textbf{98.3}}} & \underline{\textbf{3.1}} &
\multicolumn{1}{|c|}{\underline{\textbf{98.3}}} & \underline{\textbf{3.1}} &
\multicolumn{1}{|c|}{\underline{\textbf{98.3}}} & \underline{\textbf{3.1}} &
\multicolumn{1}{|c|}{\underline{\textbf{98.3}}} & \underline{\textbf{3.1}}
 \\
$\varepsilon$ = 40 &
\multicolumn{1}{|c|}{\textbf{98.4}} & \textbf{3.1} &
\multicolumn{1}{|c|}{\underline{\textbf{98.4}}} & \underline{\textbf{3.1}} &
\multicolumn{1}{|c|}{\underline{\textbf{98.4}}} & \underline{\textbf{3.1}} &
\multicolumn{1}{|c|}{\underline{\textbf{98.4}}} & \underline{\textbf{3.1}} &
\multicolumn{1}{|c|}{\underline{\textbf{98.4}}} & \underline{\textbf{3.1}}
\\
\hline
L$_{1}$ Average & 
    \multicolumn{1}{|c|}{\textbf{96.8}} & \textbf{9.4} & 
    \multicolumn{1}{c|}{\underline{97.0}} & 8.2 &
    \multicolumn{1}{c|}{97.1} & \underline{8.6} &
    \multicolumn{1}{c|}{97.4} & 7.0 &
    \multicolumn{1}{c|}{97.1} & 8.1
    \\  \hline
\midrule
  
  
 \textbf{Norm L$_2$}  &
  \multicolumn{1}{|c|}{\auc\%} &
  \fpr\% &
  \multicolumn{1}{c|}{\auc\%} &
  \fpr\% &
  \multicolumn{1}{c|}{\auc\%} &
  \fpr\% &
  \multicolumn{1}{c|}{\auc\%} &
  \fpr\% &
  \multicolumn{1}{c|}{\auc\%} &
  \fpr\%\\ \midrule
  \underline{PGD2}  & \multicolumn{1}{|c|}{ } & \multicolumn{1}{c||}{ } & \multicolumn{1}{c|}{ } & \multicolumn{1}{c||}{ } & \multicolumn{1}{c|}{ } & \multicolumn{1}{c||}{ } & \multicolumn{1}{c|}{ } & \multicolumn{1}{c||}{ } & \multicolumn{1}{c|}{ } & \multicolumn{1}{c}{ }\\
  $\varepsilon$ = 0.125 &
\multicolumn{1}{|c|}{\textbf{80.4}} & 55.3 &
\multicolumn{1}{|c|}{81.3} & 55.2 &
\multicolumn{1}{|c|}{81.3} & 49.7 &
\multicolumn{1}{|c|}{82.3} & 51.4 &
\multicolumn{1}{|c|}{\underline{81.2}} & \underline{\textbf{56.2}}
 \\
$\varepsilon$ = 0.25 &
\multicolumn{1}{|c|}{\textbf{86.4}} & \textbf{42.0} &
\multicolumn{1}{|c|}{87.6} & 38.4 &
\multicolumn{1}{|c|}{87.9} & 39.7 &
\multicolumn{1}{|c|}{89.0} & 40.3 &
\multicolumn{1}{|c|}{\underline{86.9}} & \underline{41.8}
 \\
$\varepsilon$ = 0.3125 &
\multicolumn{1}{|c|}{\textbf{88.7}} & 33.8 &
\multicolumn{1}{|c|}{89.8} & 33.1 &
\multicolumn{1}{|c|}{90.1} & 32.5 &
\multicolumn{1}{|c|}{90.9} & 32.7 &
\multicolumn{1}{|c|}{\underline{89.1}} & \underline{\textbf{37.6}}
 \\
$\varepsilon$ = 0.5 &
\multicolumn{1}{|c|}{\textbf{92.6}} & 21.7 &
\multicolumn{1}{|c|}{\underline{92.9}} & 20.6 &
\multicolumn{1}{|c|}{93.1} & \underline{\textbf{22.2}} &
\multicolumn{1}{|c|}{94.7} & 17.5 &
\multicolumn{1}{|c|}{93.2} & 20.9
 \\
$\varepsilon$ = 1 &
\multicolumn{1}{|c|}{\textbf{96.8}} & \textbf{10.0} &
\multicolumn{1}{|c|}{\underline{96.9}} & \underline{9.2} &
\multicolumn{1}{|c|}{97.0} & 9.0 &
\multicolumn{1}{|c|}{97.5} & 7.0 &
\multicolumn{1}{|c|}{97.2} & 8.1
 \\
$\varepsilon$ = 1.5 &
\multicolumn{1}{|c|}{\textbf{97.5}} & \textbf{8.1} &
\multicolumn{1}{|c|}{\underline{97.6}} & \underline{7.5} &
\multicolumn{1}{|c|}{\underline{97.6}} & 7.1 &
\multicolumn{1}{|c|}{98.0} & 4.5 &
\multicolumn{1}{|c|}{\underline{97.6}} & 7.3
 \\
$\varepsilon$ = 2 &
\multicolumn{1}{|c|}{\textbf{98.0}} & \textbf{4.6} &
\multicolumn{1}{|c|}{\underline{98.1}} & \underline{4.1} &
\multicolumn{1}{|c|}{\underline{98.1}} & 3.5 &
\multicolumn{1}{|c|}{\underline{98.1}} & 3.7 &
\multicolumn{1}{|c|}{\underline{98.1}} & 3.3
\\

  \multicolumn{1}{c||}{\underline{DeepFool}} & \multicolumn{1}{|c|}{ } & \multicolumn{1}{c||}{ } & \multicolumn{1}{c|}{ } & \multicolumn{1}{c||}{ } & \multicolumn{1}{c|}{ } & \multicolumn{1}{c||}{ } & \multicolumn{1}{c|}{ } & \multicolumn{1}{c||}{ } & \multicolumn{1}{c|}{ } & \multicolumn{1}{c}{ } \\ 
  No $\varepsilon$ & 
    \multicolumn{1}{|c|}{\textbf{97.8}} & \textbf{4.8} &
    \multicolumn{1}{|c|}{\underline{\textbf{97.8}}} & \underline{\textbf{4.8}} &
    \multicolumn{1}{|c|}{\underline{\textbf{97.8}}} & \underline{\textbf{4.8}} &
    \multicolumn{1}{|c|}{\underline{\textbf{97.8}}} & \underline{\textbf{4.8}} &
    \multicolumn{1}{|c|}{\underline{\textbf{97.8}}} & \underline{\textbf{4.8}}
    \\

  \multicolumn{1}{c||}{\underline{CW2}} & \multicolumn{1}{|c|}{ } & \multicolumn{1}{c||}{ } & \multicolumn{1}{c|}{ } & \multicolumn{1}{c||}{ } & \multicolumn{1}{c|}{ } & \multicolumn{1}{c||}{ } & \multicolumn{1}{c|}{ } & \multicolumn{1}{c||}{ } & \multicolumn{1}{c|}{ } & \multicolumn{1}{c}{ }  \\
  $\varepsilon$ = 0.01 &
    \multicolumn{1}{|c|}{\textbf{66.9}} & \textbf{81.9} &
    \multicolumn{1}{|c|}{\underline{\textbf{66.9}}} & \underline{\textbf{81.9}} &
    \multicolumn{1}{|c|}{\underline{\textbf{66.9}}} & \underline{\textbf{81.9}} &
    \multicolumn{1}{|c|}{\underline{\textbf{66.9}}} & \underline{\textbf{81.9}} &
    \multicolumn{1}{|c|}{\underline{\textbf{66.9}}} & \underline{\textbf{81.9}}
     \\

  \multicolumn{1}{c||}{\underline{HOP}}  & \multicolumn{1}{|c|}{ } & \multicolumn{1}{c||}{ } & \multicolumn{1}{c|}{ } & \multicolumn{1}{c||}{ } & \multicolumn{1}{c|}{ } & \multicolumn{1}{c||}{ } & \multicolumn{1}{c|}{ } & \multicolumn{1}{c||}{ } & \multicolumn{1}{c|}{ } & \multicolumn{1}{c}{ }   \\
  $\varepsilon$ = 0.1 &
    \multicolumn{1}{|c|}{\textbf{98.3}} & \textbf{3.1} &
    \multicolumn{1}{|c|}{\underline{\textbf{98.3}}} & \underline{\textbf{3.1}} &
    \multicolumn{1}{|c|}{\underline{\textbf{98.3}}} & \underline{\textbf{3.1}} &
    \multicolumn{1}{|c|}{\underline{\textbf{98.3}}} & \underline{\textbf{3.1}} &
    \multicolumn{1}{|c|}{\underline{\textbf{98.3}}} & \underline{\textbf{3.1}}

  \\
\hline
L$_2$ Average & \multicolumn{1}{|c|}{\textbf{90.3}} & \textbf{26.5} &
    \multicolumn{1}{c|}{90.7} & 25.8 &
    \multicolumn{1}{c|}{90.8} & 25.4 &
    \multicolumn{1}{c|}{91.4} & 23.7 &
    \multicolumn{1}{c|}{\underline{90.6}} & \underline{\textbf{26.5}}
    \\ \hline
\midrule

 \textbf{Norm L$_\infty$} &
  \multicolumn{1}{|c|}{\auc\%} &
  \fpr\% &
  \multicolumn{1}{c|}{\auc\%} &
  \fpr\% &
  \multicolumn{1}{c|}{\auc\%} &
  \fpr\% &
  \multicolumn{1}{c|}{\auc\%} &
  \fpr\% &
  \multicolumn{1}{c|}{\auc\%} &
  \fpr\%\\ \midrule
   
  \underline{PGDi, FGSM, BIM} & \multicolumn{1}{|c|}{ } & \multicolumn{1}{c||}{ } & \multicolumn{1}{c|}{ } & \multicolumn{1}{c||}{ } & \multicolumn{1}{c|}{ } & \multicolumn{1}{c||}{ } & \multicolumn{1}{c|}{ } & \multicolumn{1}{c||}{ } & \multicolumn{1}{c|}{ } & \multicolumn{1}{c}{ }\\
    $\varepsilon$ = 0.03125 &
    \multicolumn{1}{|c|}{\textbf{93.4}} & \textbf{9.6} &
    \multicolumn{1}{|c|}{\underline{94.5}} & \underline{9.5} &
    \multicolumn{1}{|c|}{94.7} & \underline{9.5} &
    \multicolumn{1}{|c|}{95.3} & \underline{9.5} &
    \multicolumn{1}{|c|}{\underline{94.5}} & \underline{9.5}
     \\
    $\varepsilon$ = 0.0625 &
    \multicolumn{1}{|c|}{\textbf{92.5}} & \textbf{9.6} &
    \multicolumn{1}{|c|}{\underline{93.2}} & \underline{\textbf{9.6}} &
    \multicolumn{1}{|c|}{93.6} & \underline{\textbf{9.6}} &
    \multicolumn{1}{|c|}{93.5} & \underline{\textbf{9.6}} &
    \multicolumn{1}{|c|}{93.3} & \underline{\textbf{9.6}}
     \\
    $\varepsilon$ = 0.25 &
    \multicolumn{1}{|c|}{\textbf{92.2}} & \textbf{9.6} &
    \multicolumn{1}{|c|}{\underline{93.1}} & \underline{\textbf{9.6}} &
    \multicolumn{1}{|c|}{93.2} & \underline{\textbf{9.6}} &
    \multicolumn{1}{|c|}{93.7} & \underline{\textbf{9.6}} &
    \multicolumn{1}{|c|}{93.5} & \underline{\textbf{9.6}}
     \\
    $\varepsilon$ = 0.5 &
    \multicolumn{1}{|c|}{\textbf{91.5}} & \textbf{9.6} &
    \multicolumn{1}{|c|}{92.5} & \underline{\textbf{9.6}} &
    \multicolumn{1}{|c|}{\underline{92.3}} & \underline{\textbf{9.6}}&
    \multicolumn{1}{|c|}{93.8} & \underline{\textbf{9.6}} &
    \multicolumn{1}{|c|}{93.0} & \underline{\textbf{9.6}}
     \\

    
    \multicolumn{1}{c||}{\underline{PGDi, FGSM, BIM, CWi, SA}} & \multicolumn{1}{|c|}{ } & \multicolumn{1}{c||}{ } & \multicolumn{1}{c|}{ } & \multicolumn{1}{c||}{ } & \multicolumn{1}{c|}{ } & \multicolumn{1}{c||}{ } & \multicolumn{1}{c|}{ } & \multicolumn{1}{c||}{ } & \multicolumn{1}{c|}{ } & \multicolumn{1}{c}{ }  \\ 
$\varepsilon$ = 0.3125 &
\multicolumn{1}{|c|}{73.9} & 79.0 &
\multicolumn{1}{|c|}{74.2} & 79.1 &
\multicolumn{1}{|c|}{\textbf{\underline{73.5}}} & \textbf{\underline{79.6}} &
\multicolumn{1}{|c|}{73.7} & 79.4 &
\multicolumn{1}{|c|}{74.3} & 79.2

  \\
\hline
L$_{\infty}$ Average & 
    \multicolumn{1}{|c|}{\textbf{88.7}} & 23.5 &
    \multicolumn{1}{c|}{\underline{89.5}} & 23.5 &
    \multicolumn{1}{c|}{\underline{89.5}} & \underline{\textbf{23.6}} &
    \multicolumn{1}{c|}{90.0} & \underline{\textbf{23.6}} &
    \multicolumn{1}{c|}{89.8} & 23.5
    \\ \hline
\midrule

  \textbf{No norm} &
  \multicolumn{1}{|c|}{\auc\%} &
  \fpr\% &
  \multicolumn{1}{c|}{\auc\%} &
  \fpr\% &
  \multicolumn{1}{c|}{\auc\%} &
  \fpr\% &
  \multicolumn{1}{c|}{\auc\%} &
  \fpr\% &
  \multicolumn{1}{c|}{\auc\%} &
  \fpr\%\\ \midrule
  \underline{STA}  & \multicolumn{1}{|c|}{ } & \multicolumn{1}{c||}{ } & \multicolumn{1}{c|}{ } & \multicolumn{1}{c||}{ } & \multicolumn{1}{c|}{ } & \multicolumn{1}{c||}{ } & \multicolumn{1}{c|}{ } & \multicolumn{1}{c||}{ } & \multicolumn{1}{c|}{ } & \multicolumn{1}{c}{ } \\
No $\varepsilon$ & 
    \multicolumn{1}{|c|}{\textbf{87.1}} & \textbf{57.8} &
    \multicolumn{1}{|c|}{\underline{\textbf{87.1}}} & \underline{\textbf{57.8}} &
    \multicolumn{1}{|c|}{\underline{\textbf{87.1}}} & \underline{\textbf{57.8}} &
    \multicolumn{1}{|c|}{\underline{\textbf{87.1}}} & \underline{\textbf{57.8}} &
    \multicolumn{1}{|c|}{\underline{\textbf{87.1}}} & \underline{\textbf{57.8}}
     \\ \hline
     No norm Average & \multicolumn{1}{|c|}{\textbf{87.1}} & \textbf{57.8} &
    \multicolumn{1}{|c|}{\underline{\textbf{87.1}}} & \underline{\textbf{57.8}} &
    \multicolumn{1}{|c|}{\underline{\textbf{87.1}}} & \underline{\textbf{57.8}} &
    \multicolumn{1}{|c|}{\underline{\textbf{87.1}}} & \underline{\textbf{57.8}} &
    \multicolumn{1}{|c|}{\underline{\textbf{87.1}}} & \underline{\textbf{57.8}} 
    \\ \hline
  \bottomrule
\end{tabular}
}
\caption{Performances on \method{NSS} per objective and in \mead~on MNIST. The worst results among all the settings 
is in \textbf{bold}; the ones in the single-armed setting is \underline{underlined}. No norm denotes the group of attacks that do not depend on the norm constraint.}
\label{tab:mnist_nss_per_loss}
\end{table*}

\method{NSS} is effective on MNIST, in particular when considering L$_{\infty}$ threats. However, in this case, when $\varepsilon = 0.3125$, the performances decrease. This is not surprising as CWi and SA have different attack schemes from the ones in PGD/FGSM/BIM. 
\method{NSS} also loses some of its effectiveness with L$_{1}$ and L$_{2}$ threats. Note that all the single-armed settings behave quite similarly: this is probably due to the computation of the \textit{Natural Scene Statistics}, which are not meaningful for perturbed images. Therefore, it is not surprising that the decrease in \auc~considering \mead~is less than one percentage point compared to the worst single-armed setting.

\newpage
\subsubsection{\method{KD-BU}}\label{appendix:mnist-kd-bu}
In~\cref{tab:mnist_kd_bu_per_loss}, we show the result of our \mead~framework on MNIST, evaluated on \method{KD-BU}.
\begin{table*}[!htb]
\centering
\resizebox{\columnwidth}{!}{%
\begin{tabular}{c||cc||cc||cc||cc||cc}
\toprule 
    \textbf{\method{KD-BU}} &
  \multicolumn{2}{c||}{\mead} & \multicolumn{2}{c||}{ACE}& \multicolumn{2}{c||}{KL}& \multicolumn{2}{c||}{Gini} &
  \multicolumn{2}{c}{FR}
   \\ \hline\midrule
  \textbf{Norm L$_1$}&
  \multicolumn{1}{|c|}{\auc\%} &
   \fpr\%  &
  \multicolumn{1}{c|}{\auc\%} &
   \fpr\% 
   &
  \multicolumn{1}{c|}{\auc\%} &
  \fpr\% &
  \multicolumn{1}{c|}{\auc\%} &
   \fpr\% &
  \multicolumn{1}{c|}{\auc\%} &
   \fpr\% \\ \midrule
   \underline{PGD1}  & \multicolumn{1}{|c|}{ } & \multicolumn{1}{c||}{ } & \multicolumn{1}{c|}{ } & \multicolumn{1}{c||}{ } & \multicolumn{1}{c|}{ } & \multicolumn{1}{c||}{ } & \multicolumn{1}{c|}{ } & \multicolumn{1}{c||}{ } & \multicolumn{1}{c|}{ } & \multicolumn{1}{c}{ } \\
       $\varepsilon$ = 5 &
    \multicolumn{1}{|c|}{\textbf{45.2}} & \textbf{95.7} &
    \multicolumn{1}{|c|}{60.9} & 92.5 &
    \multicolumn{1}{|c|}{56.6} & 93.8 &
    \multicolumn{1}{|c|}{61.5} & 92.4 &
    \multicolumn{1}{|c|}{\underline{52.8}} & \underline{94.6}
     \\
    $\varepsilon$ = 10 &
    \multicolumn{1}{|c|}{\textbf{46.4}} & \textbf{95.6} &
    \multicolumn{1}{|c|}{\underline{58.3}} & \underline{93.3} &
    \multicolumn{1}{|c|}{59.9} & 92.9 &
    \multicolumn{1}{|c|}{59.2} & 93.2 &
    \multicolumn{1}{|c|}{59.4} & 93.0
     \\
    $\varepsilon$ = 15 &
    \multicolumn{1}{|c|}{\textbf{45.5}} & \textbf{95.7} &
    \multicolumn{1}{|c|}{58.3} & \underline{93.4} &
    \multicolumn{1}{|c|}{59.3} & 93.0 &
    \multicolumn{1}{|c|}{59.7} & 93.0 &
    \multicolumn{1}{|c|}{\underline{58.2}} & \underline{93.4}
     \\
    $\varepsilon$ = 20 &
    \multicolumn{1}{|c|}{\textbf{45.5}} & \textbf{95.7} &
    \multicolumn{1}{|c|}{59.7} & 93.1 &
    \multicolumn{1}{|c|}{\underline{58.7}} & \underline{93.3} &
    \multicolumn{1}{|c|}{61.3} & 92.7 &
    \multicolumn{1}{|c|}{59.6} & 93.1
     \\
    $\varepsilon$ = 25 &
    \multicolumn{1}{|c|}{\textbf{45.8}} & \textbf{95.7} &
    \multicolumn{1}{|c|}{60.3} & \underline{93.0} &
    \multicolumn{1}{|c|}{\underline{60.2}} & 92.9 &
    \multicolumn{1}{|c|}{61.7} & 92.7 &
    \multicolumn{1}{|c|}{60.5} & 92.9
     \\
    $\varepsilon$ = 30 &
    \multicolumn{1}{|c|}{\textbf{45.7}} & \textbf{95.7} &
    \multicolumn{1}{|c|}{60.8} & 92.9 &
    \multicolumn{1}{|c|}{60.4} & 92.9 &
    \multicolumn{1}{|c|}{62.9} & 92.3 &
    \multicolumn{1}{|c|}{\underline{60.3}} & \underline{93.0}
     \\
    $\varepsilon$ = 40 &
    \multicolumn{1}{|c|}{\textbf{44.8}} & \textbf{95.8} &
    \multicolumn{1}{|c|}{61.2} & 92.8 &
    \multicolumn{1}{|c|}{\underline{60.2}} & \underline{93.0} &
    \multicolumn{1}{|c|}{63.5} & 92.3 &
    \multicolumn{1}{|c|}{61.2} & 92.8
     \\
   \hline
   L$_{1}$ Average & 
   \multicolumn{1}{|c|}{\textbf{45.6}} & \textbf{95.7} &
    \multicolumn{1}{c|}{59.9} & 93.0 &
    \multicolumn{1}{c|}{59.3} & 93.1 &
    \multicolumn{1}{c|}{61.4} & 92.7 &
    \multicolumn{1}{c|}{\underline{58.9}} & \underline{93.3}
    \\ \hline
   \midrule
  
  
\textbf{Norm L$_2$} &
  \multicolumn{1}{|c|}{\auc\%} &
   \fpr\% &
  \multicolumn{1}{|c|}{\auc\%} &
   \fpr\% &
  \multicolumn{1}{|c|}{\auc\%} &
   \fpr\% &
  \multicolumn{1}{|c|}{\auc\%} &
   \fpr\% &
  \multicolumn{1}{|c|}{\auc\%} &
   \fpr\%\\ \midrule
   \underline{PGD2}  & \multicolumn{1}{|c|}{ } & \multicolumn{1}{c||}{ } & \multicolumn{1}{c|}{ } & \multicolumn{1}{c||}{ } & \multicolumn{1}{c|}{ } & \multicolumn{1}{c||}{ } & \multicolumn{1}{c|}{ } & \multicolumn{1}{c||}{ } & \multicolumn{1}{c|}{ } & \multicolumn{1}{c}{ }\\
   $\varepsilon$ = 0.125 &
    \multicolumn{1}{|c|}{\textbf{43.0}} & \textbf{96.0} &
    \multicolumn{1}{|c|}{59.8} & 92.8 &
    \multicolumn{1}{|c|}{58.1} & 93.4 &
    \multicolumn{1}{|c|}{\underline{55.8}} & \underline{94.0} &
    \multicolumn{1}{|c|}{56.0} & \underline{94.0}
     \\
    $\varepsilon$ = 0.25 &
    \multicolumn{1}{|c|}{\textbf{44.1}} & \textbf{95.9} &
    \multicolumn{1}{|c|}{60.4} & 92.6 &
    \multicolumn{1}{|c|}{57.7} & 93.5 &
    \multicolumn{1}{|c|}{\underline{56.0}} & \underline{94.0} &
    \multicolumn{1}{|c|}{60.6} & 92.6
     \\
    $\varepsilon$ = 0.3125 &
    \multicolumn{1}{|c|}{\textbf{45.3}} & \textbf{95.7} &
    \multicolumn{1}{|c|}{\underline{55.6}} & \underline{94.0} &
    \multicolumn{1}{|c|}{59.2} & 93.0 &
    \multicolumn{1}{|c|}{55.7} & \underline{94.0} &
    \multicolumn{1}{|c|}{59.3} & 93.0
     \\
    $\varepsilon$ = 0.5 &
    \multicolumn{1}{|c|}{\textbf{43.9}} & \textbf{95.9} &
    \multicolumn{1}{|c|}{57.0} & 93.7 &
    \multicolumn{1}{|c|}{55.8} & \underline{94.0} &
    \multicolumn{1}{|c|}{58.3} & 93.4 &
    \multicolumn{1}{|c|}{\underline{55.5}} & \underline{94.0}
     \\
    $\varepsilon$ = 1 &
    \multicolumn{1}{|c|}{\textbf{46.6}} & \textbf{95.6} &
    \multicolumn{1}{|c|}{59.2} & 93.1 &
    \multicolumn{1}{|c|}{\underline{58.1}} & \underline{93.4} &
    \multicolumn{1}{|c|}{59.4} & 93.0 &
    \multicolumn{1}{|c|}{58.4} & 93.3
     \\
    $\varepsilon$ = 1.5 &
    \multicolumn{1}{|c|}{\textbf{45.8}} & \textbf{95.7} &
    \multicolumn{1}{|c|}{\underline{58.8}} & \underline{93.3} &
    \multicolumn{1}{|c|}{60.0} & 92.9 &
    \multicolumn{1}{|c|}{58.9} & \underline{93.3} &
    \multicolumn{1}{|c|}{\underline{58.8}} & 93.2
     \\
    $\varepsilon$ = 2 &
    \multicolumn{1}{|c|}{\textbf{46.7}} & \textbf{95.6} &
    \multicolumn{1}{|c|}{60.0} & 93.0 &
    \multicolumn{1}{|c|}{\underline{59.5}} & \underline{93.1} &
    \multicolumn{1}{|c|}{61.0} & 92.8 &
    \multicolumn{1}{|c|}{60.9} & 92.7
       \\

   \underline{DeepFool} & \multicolumn{1}{|c|}{ } & \multicolumn{1}{c||}{ } & \multicolumn{1}{c|}{ } & \multicolumn{1}{c||}{ } & \multicolumn{1}{c|}{ } & \multicolumn{1}{c||}{ } & \multicolumn{1}{c|}{ } & \multicolumn{1}{c||}{ } & \multicolumn{1}{c|}{ } & \multicolumn{1}{c}{ }\\ 
    No $\varepsilon$ & 
    \multicolumn{1}{|c|}{\textbf{62.9}} & \textbf{92.4} &
    \multicolumn{1}{|c|}{\underline{\textbf{62.9}}} & \underline{\textbf{92.4}} &
    \multicolumn{1}{|c|}{\underline{\textbf{62.9}}} & \underline{\textbf{92.4}} &
    \multicolumn{1}{|c|}{\underline{\textbf{62.9}}} & \underline{\textbf{92.4}} &
    \multicolumn{1}{|c|}{\underline{\textbf{62.9}}} & \underline{\textbf{92.4}}
     \\
   \underline{CW2} & \multicolumn{1}{|c|}{ } & \multicolumn{1}{c||}{ } & \multicolumn{1}{c|}{ } & \multicolumn{1}{c||}{ } & \multicolumn{1}{c|}{ } & \multicolumn{1}{c||}{ } & \multicolumn{1}{c|}{ } & \multicolumn{1}{c||}{ } & \multicolumn{1}{c|}{ } & \multicolumn{1}{c}{ }\\
   $\varepsilon$ = 0.01 &
    \multicolumn{1}{|c|}{\textbf{62.5}} & \textbf{92.5} &
    \multicolumn{1}{|c|}{\underline{\textbf{62.5}}} & \underline{\textbf{92.5}} &
    \multicolumn{1}{|c|}{\underline{\textbf{62.5}}} & \underline{\textbf{92.5}} &
    \multicolumn{1}{|c|}{\underline{\textbf{62.5}}} & \underline{\textbf{92.5}} &
    \multicolumn{1}{|c|}{\underline{\textbf{62.5}}} & \underline{\textbf{92.5}}
     \\

   \underline{HOP} & \multicolumn{1}{|c|}{ } & \multicolumn{1}{c||}{ } & \multicolumn{1}{c|}{ } & \multicolumn{1}{c||}{ } & \multicolumn{1}{c|}{ } & \multicolumn{1}{c||}{ } & \multicolumn{1}{c|}{ } & \multicolumn{1}{c||}{ } & \multicolumn{1}{c|}{ } & \multicolumn{1}{c}{ }\\  
    $\varepsilon$ = 0.1 &
    \multicolumn{1}{|c|}{\textbf{62.6}} & \textbf{92.5} &
    \multicolumn{1}{|c|}{\underline{\textbf{62.6}}} & \underline{\textbf{92.5}} &
    \multicolumn{1}{|c|}{\underline{\textbf{62.6}}} & \underline{\textbf{92.5}} &
    \multicolumn{1}{|c|}{\underline{\textbf{62.6}}} & \underline{\textbf{92.5}} &
    \multicolumn{1}{|c|}{\underline{\textbf{62.6}}} & \underline{\textbf{92.5}}
     \\
    \hline
    L$_{2}$ Average & \multicolumn{1}{|c|}{\textbf{50.3}} & \textbf{94.8} &
    \multicolumn{1}{c|}{59.9} & \underline{93.0} &
    \multicolumn{1}{c|}{59.7} & 93.1 &
    \multicolumn{1}{c|}{\underline{59.3}} & 93.2 &
    \multicolumn{1}{c|}{59.8} & \underline{93.0}
    \\ \hline
    \midrule
  

\textbf{Norm L$_\infty$}  &
  \multicolumn{1}{|c|}{\auc\%} &
   \fpr\% &
  \multicolumn{1}{|c|}{\auc\%} &
   \fpr\% &
  \multicolumn{1}{|c|}{\auc\%} &
   \fpr\% &
  \multicolumn{1}{|c|}{\auc\%} &
   \fpr\% &
  \multicolumn{1}{|c|}{\auc\%} &
   \fpr\%\\ \midrule
   
   \underline{PGDi, FGSM, BIM}  & \multicolumn{1}{|c|}{ } & \multicolumn{1}{c||}{ } & \multicolumn{1}{c|}{ } & \multicolumn{1}{c||}{ } & \multicolumn{1}{c|}{ } & \multicolumn{1}{c||}{ } & \multicolumn{1}{c|}{ } & \multicolumn{1}{c||}{ } & \multicolumn{1}{c|}{ } & \multicolumn{1}{c}{ }\\
   $\varepsilon$ = 0.03125 &
    \multicolumn{1}{|c|}{\textbf{34.5}} & \textbf{96.7} &
    \multicolumn{1}{|c|}{\underline{42.0}} & 96.1 &
    \multicolumn{1}{|c|}{44.0} & 95.9 &
    \multicolumn{1}{|c|}{48.1} & 95.4 &
    \multicolumn{1}{|c|}{42.9} & \underline{96.0}
     \\
    $\varepsilon$ = 0.0625 &
    \multicolumn{1}{|c|}{\textbf{33.6}} & \textbf{96.8} &
    \multicolumn{1}{|c|}{\underline{41.0}} & \underline{96.2} &
    \multicolumn{1}{|c|}{44.2} & 95.9 &
    \multicolumn{1}{|c|}{47.6} & 95.5 &
    \multicolumn{1}{|c|}{44.1} & 95.9
     \\
    $\varepsilon$ = 0.25 &
    \multicolumn{1}{|c|}{\textbf{34.2}} & \textbf{96.7} &
    \multicolumn{1}{|c|}{44.6} & \underline{95.8} &
    \multicolumn{1}{|c|}{\underline{44.5}} & \underline{95.8} &
    \multicolumn{1}{|c|}{52.2} & 94.8 &
    \multicolumn{1}{|c|}{45.9} & 95.6
     \\
    $\varepsilon$ = 0.5 &
    \multicolumn{1}{|c|}{\textbf{34.0}} & \textbf{96.6} &
    \multicolumn{1}{|c|}{\underline{44.5}} & \underline{95.7} &
    \multicolumn{1}{|c|}{44.8} & \underline{95.7} &
    \multicolumn{1}{|c|}{51.2} & 94.9 &
    \multicolumn{1}{|c|}{46.0} & 95.6
     \\


  
 \underline{PGDi, FGSM, BIM, CWi, SA} & \multicolumn{1}{|c|}{ } & \multicolumn{1}{c||}{ } & \multicolumn{1}{c|}{ } & \multicolumn{1}{c||}{ } & \multicolumn{1}{c|}{ } & \multicolumn{1}{c||}{ } & \multicolumn{1}{c|}{ } & \multicolumn{1}{c||}{ } & \multicolumn{1}{c|}{ } & \multicolumn{1}{c}{ }\\
    $\varepsilon$ = 0.3125 &
\multicolumn{1}{|c|}{\textbf{34.2}} & \textbf{96.7} &
\multicolumn{1}{|c|}{\underline{41.7}} & \underline{96.1} &
\multicolumn{1}{|c|}{45.8} & 95.7 &
\multicolumn{1}{|c|}{44.0} & 95.8 &
\multicolumn{1}{|c|}{45.6} & 95.7
 \\\hline
     L$_{\infty}$ Average & \multicolumn{1}{|c|}{\textbf{34.1}} & \textbf{96.7} &
    \multicolumn{1}{c|}{\underline{42.8}} & \underline{96.0}  &
    \multicolumn{1}{c|}{44.7} & 95.8 &
    \multicolumn{1}{c|}{48.6} & 95.3 &
    \multicolumn{1}{c|}{44.9} & 95.8
    \\ \hline
  \midrule
  
  \textbf{No norm}  &
  \multicolumn{1}{|c|}{\auc\%} &
   \fpr\% &
  \multicolumn{1}{|c|}{\auc\%} &
   \fpr\% &
  \multicolumn{1}{|c|}{\auc\%} &
   \fpr\% &
  \multicolumn{1}{|c|}{\auc\%} &
   \fpr\% &
  \multicolumn{1}{|c|}{\auc\%} &
   \fpr\%\\ \midrule
   \underline{STA}  & \multicolumn{1}{|c|}{ } & \multicolumn{1}{c||}{ } & \multicolumn{1}{c|}{ } & \multicolumn{1}{c||}{ } & \multicolumn{1}{c|}{ } & \multicolumn{1}{c||}{ } & \multicolumn{1}{c|}{ } & \multicolumn{1}{c||}{ } & \multicolumn{1}{c|}{ } & \multicolumn{1}{c}{ }\\
No $\varepsilon$ & 
\multicolumn{1}{|c|}{\textbf{76.0}} & \textbf{88.2} &
\multicolumn{1}{|c|}{\underline{\textbf{76.0}}} & \underline{\textbf{88.2}} &
\multicolumn{1}{|c|}{\underline{\textbf{76.0}}} & \underline{\textbf{88.2}} &
\multicolumn{1}{|c|}{\underline{\textbf{76.0}}} & \underline{\textbf{88.2}} &
\multicolumn{1}{|c|}{\underline{\textbf{76.0}}} & \underline{\textbf{88.2}}
 \\ \hline
 No norm Average & \multicolumn{1}{|c|}{\textbf{76.0}} & \textbf{88.2} &
\multicolumn{1}{|c|}{\underline{\textbf{76.0}}} & \underline{\textbf{88.2}} &
\multicolumn{1}{|c|}{\underline{\textbf{76.0}}} & \underline{\textbf{88.2}} &
\multicolumn{1}{|c|}{\underline{\textbf{76.0}}} & \underline{\textbf{88.2}} &
\multicolumn{1}{|c|}{\underline{\textbf{76.0}}} & \underline{\textbf{88.2}}
    \\ \hline
   \bottomrule
\end{tabular}
}
\caption{Performances on \method{KD-BU} per objective and in \mead~on MNIST. The worst results among all the settings 
are shown in \textbf{bold}; the ones in the single-armed setting is \underline{underlined}. No norm denotes the group of attacks that do not depend on the norm constraint.}
\label{tab:mnist_kd_bu_per_loss}
\end{table*}

The \method{KD-BU} method is the least effective one at detecting the adversarial samples under the \mead~framework. The decrease of \auc~can go up to 23 percentage points. Fisher-Rao and Adversarial Cross-Entropy-based attacks seem to be the toughest to detect for \method{KD-BU} detectors. 

\newpage
\subsubsection{\method{LID}}\label{appendix:mnist-lid}
In~\cref{tab:mnist_lid_per_loss}, we show the result of our \mead~framework on MNIST, evaluated on \method{LID}.
\begin{table*}[!htb]
\centering
\resizebox{\columnwidth}{!}{%
\begin{tabular}{c||cc||cc||cc||cc||cc}
\toprule 
    \textbf{\method{LID}} &
  \multicolumn{2}{c||}{\mead} & \multicolumn{2}{c||}{ACE}& \multicolumn{2}{c||}{KL}& \multicolumn{2}{c||}{Gini} &
  \multicolumn{2}{c}{FR}
   \\ \hline\midrule
  \textbf{Norm L$_1$}&
  \multicolumn{1}{|c|}{\auc\%} &
   {\fpr\% } &
  \multicolumn{1}{c|}{\auc\%} &
   \fpr\% 
   &
  \multicolumn{1}{c|}{\auc\%} &
  \fpr\% &
  \multicolumn{1}{c|}{\auc\%} &
   \fpr\% &
  \multicolumn{1}{c|}{\auc\%} &
   \fpr\% \\ \midrule
   \underline{PGD1}  & \multicolumn{1}{|c|}{ } & \multicolumn{1}{c||}{ } & \multicolumn{1}{c|}{ } & \multicolumn{1}{c||}{ } & \multicolumn{1}{c|}{ } & \multicolumn{1}{c||}{ } & \multicolumn{1}{c|}{ } & \multicolumn{1}{c||}{ } & \multicolumn{1}{c|}{ } & \multicolumn{1}{c}{ } \\
 $\varepsilon$ = 5 &
\multicolumn{1}{|c|}{88.1} & 41.5 &
\multicolumn{1}{|c|}{89.6} & 36.1 &
\multicolumn{1}{|c|}{88.6} & 41.7 &
\multicolumn{1}{|c|}{88.4} & 39.3 &
\multicolumn{1}{|c|}{\underline{\textbf{87.5}}} & \underline{\textbf{46.9}}
   \\ 
$\varepsilon$ = 10 &
\multicolumn{1}{|c|}{\textbf{83.2}} & \textbf{48.8} &
\multicolumn{1}{|c|}{\underline{86.8}} & 41.9 &
\multicolumn{1}{|c|}{87.1} & 41.3 &
\multicolumn{1}{|c|}{89.4} & 35.7 &
\multicolumn{1}{|c|}{86.9} & \underline{42.1}
   \\ 
$\varepsilon$ = 15 &
\multicolumn{1}{|c|}{\textbf{83.1}} & 48.8 &
\multicolumn{1}{|c|}{84.3} & 41.9 &
\multicolumn{1}{|c|}{84.2} & 46.5 &
\multicolumn{1}{|c|}{87.1} & 42.3 &
\multicolumn{1}{|c|}{\underline{83.7}} & \underline{\textbf{49.8}} 
   \\
$\varepsilon$ = 20 &
\multicolumn{1}{|c|}{\textbf{77.7}} & \textbf{57.7} &
\multicolumn{1}{|c|}{\underline{83.8}} & 47.2 &
\multicolumn{1}{|c|}{84.6} & 46.6 &
\multicolumn{1}{|c|}{90.1} & 35.5 &
\multicolumn{1}{|c|}{84.5} & \underline{47.8}
   \\
$\varepsilon$ = 25 &
\multicolumn{1}{|c|}{\textbf{78.5}} & \textbf{58.7} &
\multicolumn{1}{|c|}{\underline{83.0}} & 51.9 &
\multicolumn{1}{|c|}{83.5} & \underline{56.8} &
\multicolumn{1}{|c|}{91.4} & 32.4 &
\multicolumn{1}{|c|}{83.6} & 51.2
   \\
$\varepsilon$ = 30 &
\multicolumn{1}{|c|}{\textbf{74.3}} & \textbf{65.6} &
\multicolumn{1}{|c|}{\underline{80.8}} & \underline{57.4} &
\multicolumn{1}{|c|}{81.8} & 56.8 &
\multicolumn{1}{|c|}{92.7} & 29.1 &
\multicolumn{1}{|c|}{82.7} & 54.9
   \\ 
$\varepsilon$ = 40 &
\multicolumn{1}{|c|}{\textbf{74.5}} & \textbf{63.5} &
\multicolumn{1}{|c|}{\underline{77.5}} & \underline{60.8} &
\multicolumn{1}{|c|}{78.4} & 60.1 &
\multicolumn{1}{|c|}{93.8} & 26.5 &
\multicolumn{1}{|c|}{80.1} & 58.9
   \\ \hline
  L$_1$ Average & 
\multicolumn{1}{|c|}{\textbf{79.9}} & \textbf{54.9} &
\multicolumn{1}{|c|}{\underline{83.7}} & 48.2 &
\multicolumn{1}{|c|}{84.0} & 50.0 &
\multicolumn{1}{|c|}{90.4} & \underline{52.1} &
\multicolumn{1}{|c|}{84.1} & 50.2
     \\ \hline
 \midrule
  
\textbf{Norm L$_2$} &
  \multicolumn{1}{|c|}{\auc\%} &
   \fpr\% &
  \multicolumn{1}{|c|}{\auc\%} &
   \fpr\% &
  \multicolumn{1}{|c|}{\auc\%} &
   \fpr\% &
  \multicolumn{1}{|c|}{\auc\%} &
   \fpr\% &
  \multicolumn{1}{|c|}{\auc\%} &
   \fpr\%\\ \midrule
   \underline{PGD2}  & \multicolumn{1}{|c|}{ } & \multicolumn{1}{c||}{ } & \multicolumn{1}{c|}{ } & \multicolumn{1}{c||}{ } & \multicolumn{1}{c|}{ } & \multicolumn{1}{c||}{ } & \multicolumn{1}{c|}{ } & \multicolumn{1}{c||}{ } & \multicolumn{1}{c|}{ } & \multicolumn{1}{c}{ } \\
   $\varepsilon$ = 0.125 &
\multicolumn{1}{|c|}{87.7} & 47.7 &
\multicolumn{1}{|c|}{88.0} & 47.5 &
\multicolumn{1}{|c|}{87.6} & 47.4 &
\multicolumn{1}{|c|}{\underline{\textbf{86.8}}} & 46.7 &
\multicolumn{1}{|c|}{\underline{\textbf{86.8}}} & \underline{\textbf{49.1}}
   \\
$\varepsilon$ = 0.25 &
\multicolumn{1}{|c|}{88.0} & 40.5 &
\multicolumn{1}{|c|}{89.1} & 45.2 &
\multicolumn{1}{|c|}{\underline{\textbf{87.9}}} & \underline{\textbf{49.2}} &
\multicolumn{1}{|c|}{88.1} & 45.2 &
\multicolumn{1}{|c|}{88.0} & 44.6
   \\ 
$\varepsilon$ = 0.3125 &
\multicolumn{1}{|c|}{88.4} & 39.7 &
\multicolumn{1}{|c|}{89.6} & 44.0 &
\multicolumn{1}{|c|}{87.9} & 46.0 &
\multicolumn{1}{|c|}{\underline{\textbf{87.6}}} & \underline{\textbf{53.1}} &
\multicolumn{1}{|c|}{88.1} & 45.4
   \\
$\varepsilon$ = 0.5 &
\multicolumn{1}{|c|}{\textbf{88.0}} & 38.1 &
\multicolumn{1}{|c|}{90.0} & 33.8 &
\multicolumn{1}{|c|}{88.9} & 35.6 &
\multicolumn{1}{|c|}{\underline{88.1}} & \underline{\textbf{44.5}} &
\multicolumn{1}{|c|}{\underline{88.1}} & 41.1
   \\
$\varepsilon$ = 1 &
\multicolumn{1}{|c|}{\textbf{80.1}} & \textbf{55.3} &
\multicolumn{1}{|c|}{86.8} & 42.3 &
\multicolumn{1}{|c|}{87.0} & 41.8 &
\multicolumn{1}{|c|}{88.0} & 39.1 &
\multicolumn{1}{|c|}{\underline{86.7}} & \underline{43.2}
   \\
$\varepsilon$ = 1.5 &
\multicolumn{1}{|c|}{\textbf{81.9}} & \textbf{51.3} &
\multicolumn{1}{|c|}{84.8} & 46.0 &
\multicolumn{1}{|c|}{84.5} & 47.2 &
\multicolumn{1}{|c|}{87.4} & 42.0 &
\multicolumn{1}{|c|}{\underline{84.0}} & \underline{48.7}
   \\
$\varepsilon$ = 2 &
\multicolumn{1}{|c|}{\textbf{81.1}} & \textbf{53.6} &
\multicolumn{1}{|c|}{85.2} & 46.4 &
\multicolumn{1}{|c|}{\underline{84.8}} & \underline{47.2} &
\multicolumn{1}{|c|}{89.2} & 38.1 &
\multicolumn{1}{|c|}{85.6} & 46.2
   \\

   \underline{DeepFool} & \multicolumn{1}{|c|}{ } & \multicolumn{1}{c||}{ } & \multicolumn{1}{c|}{ } & \multicolumn{1}{c||}{ } & \multicolumn{1}{c|}{ } & \multicolumn{1}{c||}{ } & \multicolumn{1}{c|}{ } & \multicolumn{1}{c||}{ } & \multicolumn{1}{c|}{ } & \multicolumn{1}{c}{ } \\ 
No $\varepsilon$  & 
\multicolumn{1}{|c|}{\textbf{87.9}} & \textbf{42.1} &
\multicolumn{1}{|c|}{\underline{\textbf{87.9}}} & \underline{\textbf{42.1}} &
\multicolumn{1}{|c|}{\underline{\textbf{87.9}}} & \underline{\textbf{42.1}} &
\multicolumn{1}{|c|}{\underline{\textbf{87.9}}} & \underline{\textbf{42.1}} &
\multicolumn{1}{|c|}{\underline{\textbf{87.9}}} & \underline{\textbf{42.1}}
   \\

   \underline{CW2}  & \multicolumn{1}{|c|}{ } & \multicolumn{1}{c||}{ } & \multicolumn{1}{c|}{ } & \multicolumn{1}{c||}{ } & \multicolumn{1}{c|}{ } & \multicolumn{1}{c||}{ } & \multicolumn{1}{c|}{ } & \multicolumn{1}{c||}{ } & \multicolumn{1}{c|}{ } & \multicolumn{1}{c}{ } \\
$\varepsilon$ = 0.01 &
\multicolumn{1}{|c|}{\textbf{83.6}} & \textbf{52.6} &
\multicolumn{1}{|c|}{\underline{\textbf{83.6}}} & \underline{\textbf{52.6}} &
\multicolumn{1}{|c|}{\underline{\textbf{83.6}}} & \underline{\textbf{52.6}} &
\multicolumn{1}{|c|}{\underline{\textbf{83.6}}} & \underline{\textbf{52.6}} &
\multicolumn{1}{|c|}{\underline{\textbf{83.6}}} & \underline{\textbf{52.6}}
   \\ 
  
   \underline{HOP}  & \multicolumn{1}{|c|}{ } & \multicolumn{1}{c||}{ } & \multicolumn{1}{c|}{ } & \multicolumn{1}{c||}{ } & \multicolumn{1}{c|}{ } & \multicolumn{1}{c||}{ } & \multicolumn{1}{c|}{ } & \multicolumn{1}{c||}{ } & \multicolumn{1}{c|}{ } & \multicolumn{1}{c}{ } \\
$\varepsilon$ = 0.1 &
\multicolumn{1}{|c|}{\textbf{89.3}} & \textbf{41.0} &
\multicolumn{1}{|c|}{\underline{\textbf{89.3}}} & \underline{\textbf{41.0}} &
\multicolumn{1}{|c|}{\underline{\textbf{89.3}}} & \underline{\textbf{41.0}} &
\multicolumn{1}{|c|}{\underline{\textbf{89.3}}} & \underline{\textbf{41.0}} &
\multicolumn{1}{|c|}{\underline{\textbf{89.3}}} & \underline{\textbf{41.0}}
   \\ \hline
L$_2$ Average & 
\multicolumn{1}{|c|}{\textbf{85.6}} & \textbf{46.2} &
\multicolumn{1}{|c|}{87.4} & 44.1 &
\multicolumn{1}{|c|}{87.0} & 45.1 &
\multicolumn{1}{|c|}{87.6} & 44.4 &
\multicolumn{1}{|c|}{\underline{86.1}} & \underline{45.4}
     \\ \hline
 \midrule

 \textbf{Norm L$_\infty$} &
  \multicolumn{1}{|c|}{\auc\%} &
   \fpr\% &
  \multicolumn{1}{|c|}{\auc\%} &
   \fpr\% &
  \multicolumn{1}{|c|}{\auc\%} &
   \fpr\% &
  \multicolumn{1}{|c|}{\auc\%} &
   \fpr\% &
  \multicolumn{1}{|c|}{\auc\%} &
   \fpr\%\\ \midrule
   
   \underline{PGDi, FGSM, BIM} & \multicolumn{1}{|c|}{ } & \multicolumn{1}{c||}{ } & \multicolumn{1}{c|}{ } & \multicolumn{1}{c||}{ } & \multicolumn{1}{c|}{ } & \multicolumn{1}{c||}{ } & \multicolumn{1}{c|}{ } & \multicolumn{1}{c||}{ } & \multicolumn{1}{c|}{ } & \multicolumn{1}{c}{ } \\
    $\varepsilon$ = 0.03125 &
\multicolumn{1}{|c|}{87.5} & 44.1 &
\multicolumn{1}{|c|}{89.8} & 34.5 &
\multicolumn{1}{|c|}{87.6} & 45.3 &
\multicolumn{1}{|c|}{89.5} & 40.0 &
\multicolumn{1}{|c|}{\underline{\textbf{86.9}}} & \underline{\textbf{46.7}}
   \\ 
$\varepsilon$ = 0.0625 &
\multicolumn{1}{|c|}{\textbf{84.7}} & \textbf{45.5} &
\multicolumn{1}{|c|}{88.3} & 37.6 &
\multicolumn{1}{|c|}{88.1} & 36.7 &
\multicolumn{1}{|c|}{88.3} & 38.0 &
\multicolumn{1}{|c|}{\underline{87.7}} & \underline{42.4}
   \\
$\varepsilon$ = 0.125 &
\multicolumn{1}{|c|}{\textbf{80.6}} & \textbf{52.1} &
\multicolumn{1}{|c|}{85.3} & 43.1 &
\multicolumn{1}{|c|}{85.6} & 43.9 &
\multicolumn{1}{|c|}{\underline{84.2}} & 45.3 &
\multicolumn{1}{|c|}{84.4} & \underline{46.7}
   \\
$\varepsilon$ = 0.25 &
\multicolumn{1}{|c|}{\textbf{74.1}} & \textbf{63.0} &
\multicolumn{1}{|c|}{\underline{83.7}} & 48.9 &
\multicolumn{1}{|c|}{\underline{83.7}} & \underline{49.5} &
\multicolumn{1}{|c|}{91.5} & 32.9 &
\multicolumn{1}{|c|}{85.6} & 47.6
   \\ 
$\varepsilon$ = 0.5 &
\multicolumn{1}{|c|}{\textbf{65.4}} & \textbf{66.8} &
\multicolumn{1}{|c|}{74.4} & 58.8 &
\multicolumn{1}{|c|}{75.2} & 58.8 &
\multicolumn{1}{|c|}{92.3} & 32.9 &
\multicolumn{1}{|c|}{\underline{72.3}} & \underline{61.4}
   \\

  
 \underline{PGDi, FGSM, BIM, CWi, SA}  & \multicolumn{1}{|c|}{ } & \multicolumn{1}{c||}{ } & \multicolumn{1}{c|}{ } & \multicolumn{1}{c||}{ } & \multicolumn{1}{c|}{ } & \multicolumn{1}{c||}{ } & \multicolumn{1}{c|}{ } & \multicolumn{1}{c||}{ } & \multicolumn{1}{c|}{ } & \multicolumn{1}{c}{ } \\
$\varepsilon$ = 0.3125 &
\multicolumn{1}{|c|}{\textbf{74.8}} & \textbf{59.1} &
\multicolumn{1}{|c|}{\underline{78.0}} & \underline{55.1} &
\multicolumn{1}{|c|}{81.3} & 52.5 &
\multicolumn{1}{|c|}{86.2} & 43.9 &
\multicolumn{1}{|c|}{80.9} & 52.4
   \\ \hline

L$_\infty$ Average & 
\multicolumn{1}{|c|}{\textbf{77.9}} & \textbf{55.1} &
\multicolumn{1}{|c|}{83.3} & 46.3 &
\multicolumn{1}{|c|}{83.6} & 47.8 &
\multicolumn{1}{|c|}{88.7} & 38.8 &
\multicolumn{1}{|c|}{\underline{83.0}} & \underline{49.5}
     \\ \hline
 \midrule
\textbf{No norm}&
  \multicolumn{1}{|c|}{\auc\%} &
   \fpr\% &
  \multicolumn{1}{|c|}{\auc\%} &
   \fpr\% &
  \multicolumn{1}{|c|}{\auc\%} &
   \fpr\% &
  \multicolumn{1}{|c|}{\auc\%} &
   \fpr\% &
  \multicolumn{1}{|c|}{\auc\%} &
   \fpr\%\\ \midrule
   
   \underline{STA} & \multicolumn{1}{|c|}{ } & \multicolumn{1}{c||}{ } & \multicolumn{1}{c|}{ } & \multicolumn{1}{c||}{ } & \multicolumn{1}{c|}{ } & \multicolumn{1}{c||}{ } & \multicolumn{1}{c|}{ } & \multicolumn{1}{c||}{ } & \multicolumn{1}{c|}{ } & \multicolumn{1}{c}{ } \\
No $\varepsilon$ & 
\multicolumn{1}{|c|}{\textbf{98.1}} & \textbf{8.2} &
\multicolumn{1}{|c|}{\underline{\textbf{98.1}}} & \underline{\textbf{8.2}} &
\multicolumn{1}{|c|}{\underline{\textbf{98.1}}} & \underline{\textbf{8.2}} &
\multicolumn{1}{|c|}{\underline{\textbf{98.1}}} & \underline{\textbf{8.2}} &
\multicolumn{1}{|c|}{\underline{\textbf{98.1}}} & \underline{\textbf{8.2}}
   \\ \hline

No norm Average & 
\multicolumn{1}{|c|}{\textbf{98.1}} & \textbf{8.2} &
\multicolumn{1}{|c|}{\underline{\textbf{98.1}}} & \underline{\textbf{8.2}} &
\multicolumn{1}{|c|}{\underline{\textbf{98.1}}} & \underline{\textbf{8.2}} &
\multicolumn{1}{|c|}{\underline{\textbf{98.1}}} & \underline{\textbf{8.2}} &
\multicolumn{1}{|c|}{\underline{\textbf{98.1}}} & \underline{\textbf{8.2}}
    \\ \hline
   \bottomrule
\end{tabular}
}
\caption{Performances on \method{LID} per objective and in \mead~on MNIST. The worst results among all the settings 
are shown in \textbf{bold}; the ones in the single-armed setting is \underline{underlined}. No norm denotes the group of attacks that do not depend on the norm constraint.}
\label{tab:mnist_lid_per_loss}
\end{table*}

\method{LID} is quite effective in detecting STA. Contrary to the other methods, \method{LID} has more difficulty detecting attacks with significant perturbations. The maximum decrease in \auc~considering \mead~is slightly higher than 8 percentage points. Even if the results seem to be quite similar among the single-armed, the \method{LID}-based detector trained on MNIST seems more vulnerable to the attacks generated thanks to the Kullback-Leibler divergence on L$_{1}$-norm-based attacks and sensitive to the Fisher-Rao distance under L$_{2}$ threats.

\newpage
\subsubsection{\method{FS}}\label{appendix:mnist-fs}
In~\cref{tab:mnist_fs_per_loss}, we show the result of our \mead~framework on MNIST, evaluated on \method{FS}.
\begin{table*}[!htb]
\centering
\resizebox{\columnwidth}{!}{%
\begin{tabular}{c||cc||cc||cc||cc||cc}
\toprule 
    \textbf{\method{FS}} &
  \multicolumn{2}{c||}{\mead} & \multicolumn{2}{c||}{ACE}& \multicolumn{2}{c||}{KL}& \multicolumn{2}{c||}{Gini} &
  \multicolumn{2}{c}{FR}
   \\ \hline\midrule
  \textbf{Norm L$_1$}&
  \multicolumn{1}{|c|}{\auc\%} &
   \fpr\%  &
  \multicolumn{1}{c|}{\auc\%} &
   \fpr\% 
   &
  \multicolumn{1}{c|}{\auc\%} &
  \fpr\% &
  \multicolumn{1}{c|}{\auc\%} &
   \fpr\% &
  \multicolumn{1}{c|}{\auc\%} &
   \fpr\% \\ \midrule
   \underline{PGD1}  & \multicolumn{1}{|c|}{ } & \multicolumn{1}{c||}{ } & \multicolumn{1}{c|}{ } & \multicolumn{1}{c||}{ } & \multicolumn{1}{c|}{ } & \multicolumn{1}{c||}{ } & \multicolumn{1}{c|}{ } & \multicolumn{1}{c||}{ } & \multicolumn{1}{c|}{ } & \multicolumn{1}{c}{ } \\
     $\varepsilon$ = 5 &
    \multicolumn{1}{|c|}{\textbf{59.2}} & 88.1 &
    \multicolumn{1}{|c|}{63.3} & 85.4 &
    \multicolumn{1}{|c|}{62.6} & 87.1 &
    \multicolumn{1}{|c|}{\underline{\textbf{59.2}}} & 84.7 &
    \multicolumn{1}{|c|}{59.8} & \underline{\textbf{89.5}}
     \\
    $\varepsilon$ = 10 &
    \multicolumn{1}{|c|}{\textbf{72.0}} & \textbf{76.3} &
    \multicolumn{1}{|c|}{76.8} & 68.6 &
    \multicolumn{1}{|c|}{76.9} & 67.8 &
    \multicolumn{1}{|c|}{\underline{73.0}} & \underline{70.0} &
    \multicolumn{1}{|c|}{75.6} & 69.8
     \\
    $\varepsilon$ = 15 &
    \multicolumn{1}{|c|}{86.5} & \textbf{56.4} &
    \multicolumn{1}{|c|}{90.7} & 45.2 &
    \multicolumn{1}{|c|}{90.9} & 44.3 &
    \multicolumn{1}{|c|}{\underline{\textbf{84.9}}} & \underline{54.2} &
    \multicolumn{1}{|c|}{90.7} & 44.5
     \\
    $\varepsilon$ = 20 &
    \multicolumn{1}{|c|}{\textbf{87.9}} & \textbf{54.1} &
    \multicolumn{1}{|c|}{92.3} & 41.3 &
    \multicolumn{1}{|c|}{92.5} & 40.0 &
    \multicolumn{1}{|c|}{\underline{90.8}} & \underline{42.3} &
    \multicolumn{1}{|c|}{92.5} & 40.5
     \\
    $\varepsilon$ = 25 &
    \multicolumn{1}{|c|}{\textbf{86.6}} & \textbf{56.0} &
    \multicolumn{1}{|c|}{\underline{90.4}} & 46.1 &
    \multicolumn{1}{|c|}{90.5} & 45.9 &
    \multicolumn{1}{|c|}{92.0} & 39.0 &
    \multicolumn{1}{|c|}{90.9} & \underline{46.5}
     \\
    $\varepsilon$ = 30 &
    \multicolumn{1}{|c|}{\textbf{83.9}} & \textbf{60.4} &
    \multicolumn{1}{|c|}{\underline{88.0}} & \underline{53.0} &
    \multicolumn{1}{|c|}{88.3} & 50.9 &
    \multicolumn{1}{|c|}{92.0} & 39.3 &
    \multicolumn{1}{|c|}{89.5} & 49.9
     \\
    $\varepsilon$ = 40 &
    \multicolumn{1}{|c|}{\textbf{72.3}} & \textbf{76.0} &
    \multicolumn{1}{|c|}{\underline{82.5}} & \underline{63.8} &
    \multicolumn{1}{|c|}{82.7} & \underline{63.8} &
    \multicolumn{1}{|c|}{90.8} & 41.8 &
    \multicolumn{1}{|c|}{84.6} & 60.8
     \\ 
 \hline
 L$_{1}$ Average & \multicolumn{1}{|c|}{\textbf{79.8}} & \textbf{66.8} &
    \multicolumn{1}{c|}{83.4} & \underline{57.6} &
    \multicolumn{1}{c|}{83.5} & 57.1 &
    \multicolumn{1}{c|}{\underline{83.2}} & 53.0 &
    \multicolumn{1}{c|}{83.4} & 57.4
    \\ \hline
 \midrule
  
  
 \textbf{Norm L$_2$} &
  \multicolumn{1}{|c|}{\auc\%} &
  \fpr\% &
  \multicolumn{1}{|c|}{\auc\%} &
  \fpr\% &
  \multicolumn{1}{|c|}{\auc\%} &
  \fpr\% &
  \multicolumn{1}{|c|}{\auc\%} &
  \fpr\% &
  \multicolumn{1}{|c|}{\auc\%} &
  \fpr\%\\ \midrule
  \underline{PGD2}  & \multicolumn{1}{|c|}{ } & \multicolumn{1}{c||}{ } & \multicolumn{1}{c|}{ } & \multicolumn{1}{c||}{ } & \multicolumn{1}{c|}{ } & \multicolumn{1}{c||}{ } & \multicolumn{1}{c|}{ } & \multicolumn{1}{c||}{ } & \multicolumn{1}{c|}{ } & \multicolumn{1}{c}{ } \\
     $\varepsilon$ = 0.125 &
    \multicolumn{1}{|c|}{\textbf{56.6}} & 90.5 &
    \multicolumn{1}{|c|}{58.3} & 89.2 &
    \multicolumn{1}{|c|}{57.7} & \underline{\textbf{91.4}} &
    \multicolumn{1}{|c|}{\underline{57.0}} & 89.4 &
    \multicolumn{1}{|c|}{57.4} & 87.9
     \\
    $\varepsilon$ = 0.25 &
    \multicolumn{1}{|c|}{56.2} & \textbf{90.3} &
    \multicolumn{1}{|c|}{58.3} & 85.4 &
    \multicolumn{1}{|c|}{58.5} & 85.8 &
    \multicolumn{1}{|c|}{58.0} & \underline{89.3} &
    \multicolumn{1}{|c|}{\underline{\textbf{55.9}}} & 89.1
     \\
    $\varepsilon$ = 0.3125 &
    \multicolumn{1}{|c|}{\textbf{57.1}} & 88.1 &
    \multicolumn{1}{|c|}{60.3} & 85.6 &
    \multicolumn{1}{|c|}{60.1} & 86.6 &
    \multicolumn{1}{|c|}{60.8} & 86.2 &
    \multicolumn{1}{|c|}{\underline{57.7}} & \underline{\textbf{91.3}}
     \\
    $\varepsilon$ = 0.5 &
    \multicolumn{1}{|c|}{\textbf{59.5}} & 85.4 &
    \multicolumn{1}{|c|}{62.9} & 82.7 &
    \multicolumn{1}{|c|}{62.4} & 85.4 &
    \multicolumn{1}{|c|}{64.0} & 81.7 &
    \multicolumn{1}{|c|}{\underline{60.5}} & \underline{\textbf{85.9}}
     \\
    $\varepsilon$ = 1 &
    \multicolumn{1}{|c|}{76.9} & \textbf{67.1} &
    \multicolumn{1}{|c|}{80.2} & 62.8 &
    \multicolumn{1}{|c|}{80.2} & 62.5 &
    \multicolumn{1}{|c|}{\underline{\textbf{76.4}}} & \underline{64.5} &
    \multicolumn{1}{|c|}{79.6} & 61.0
     \\
    $\varepsilon$ = 1.5 &
    \multicolumn{1}{|c|}{89.2} & 47.0 &
    \multicolumn{1}{|c|}{92.9} & 33.3 &
    \multicolumn{1}{|c|}{93.1} & 33.2 &
    \multicolumn{1}{|c|}{\underline{\textbf{86.9}}} & \underline{\textbf{47.2}} &
    \multicolumn{1}{|c|}{92.8} & 35.5
     \\
    $\varepsilon$ = 2 &
    \multicolumn{1}{|c|}{\textbf{88.8}} & \textbf{50.7} &
    \multicolumn{1}{|c|}{92.4} & 40.5 &
    \multicolumn{1}{|c|}{92.7} & 39.1 &
    \multicolumn{1}{|c|}{\underline{91.2}} & \underline{41.7} &
    \multicolumn{1}{|c|}{93.2} & 36.3 
       \\

  \underline{DeepFool} & \multicolumn{1}{|c|}{ } & \multicolumn{1}{c||}{ } & \multicolumn{1}{c|}{ } & \multicolumn{1}{c||}{ } & \multicolumn{1}{c|}{ } & \multicolumn{1}{c||}{ } & \multicolumn{1}{c|}{ } & \multicolumn{1}{c||}{ } & \multicolumn{1}{c|}{ } & \multicolumn{1}{c}{ } \\ 
    No $\varepsilon$ &
    \multicolumn{1}{|c|}{\textbf{88.3}} & \textbf{52.0} &
    \multicolumn{1}{|c|}{\underline{\textbf{88.3}}} & \underline{\textbf{52.0}} &
    \multicolumn{1}{|c|}{\underline{\textbf{88.3}}} & \underline{\textbf{52.0}} &
    \multicolumn{1}{|c|}{\underline{\textbf{88.3}}} & \underline{\textbf{52.0}} &
    \multicolumn{1}{|c|}{\underline{\textbf{88.3}}} & \underline{\textbf{52.0}}
     \\
  
  \underline{CW2}  & \multicolumn{1}{|c|}{ } & \multicolumn{1}{c||}{ } & \multicolumn{1}{c|}{ } & \multicolumn{1}{c||}{ } & \multicolumn{1}{c|}{ } & \multicolumn{1}{c||}{ } & \multicolumn{1}{c|}{ } & \multicolumn{1}{c||}{ } & \multicolumn{1}{c|}{ } & \multicolumn{1}{c}{ } \\
  $\varepsilon$ = 0.01 &
    \multicolumn{1}{|c|}{\textbf{68.6}} & \textbf{81.7} &
    \multicolumn{1}{|c|}{\underline{\textbf{68.6}}} & \underline{\textbf{81.7}} &
    \multicolumn{1}{|c|}{\underline{\textbf{68.6}}} & \underline{\textbf{81.7}} &
    \multicolumn{1}{|c|}{\underline{\textbf{68.6}}} & \underline{\textbf{81.7}} &
    \multicolumn{1}{|c|}{\underline{\textbf{68.6}}} & \underline{\textbf{81.7}}
     \\

  \underline{HOP} & \multicolumn{1}{|c|}{ } & \multicolumn{1}{c||}{ } & \multicolumn{1}{c|}{ } & \multicolumn{1}{c||}{ } & \multicolumn{1}{c|}{ } & \multicolumn{1}{c||}{ } & \multicolumn{1}{c|}{ } & \multicolumn{1}{c||}{ } & \multicolumn{1}{c|}{ } & \multicolumn{1}{c}{ } \\ 
  $\varepsilon$ = 0.1 &
    \multicolumn{1}{|c|}{\textbf{93.4}} & \textbf{36.6} &
    \multicolumn{1}{|c|}{\underline{\textbf{93.4}}} & \underline{\textbf{36.6}} &
    \multicolumn{1}{|c|}{\underline{\textbf{93.4}}} & \underline{\textbf{36.6}} &
    \multicolumn{1}{|c|}{\underline{\textbf{93.4}}} & \underline{\textbf{36.6}} &
    \multicolumn{1}{|c|}{\underline{\textbf{93.4}}} & \underline{\textbf{36.6}}
     \\ \hline
  L$_{2}$ Average & \multicolumn{1}{|c|}{\textbf{73.5}} & \textbf{69.0} &
    \multicolumn{1}{c|}{75.6} & 65.0 &
    \multicolumn{1}{c|}{75.5} & 65.4 &
    \multicolumn{1}{c|}{\underline{74.5}} & \underline{67.0} &
    \multicolumn{1}{c|}{74.7} & 65.7
    \\ \hline
  \midrule
  

\textbf{Norm L$_\infty$} &
  \multicolumn{1}{|c|}{\auc\%} &
  \fpr\% &
  \multicolumn{1}{|c|}{\auc\%} &
  \fpr\% &
  \multicolumn{1}{|c|}{\auc\%} &
  \fpr\% &
  \multicolumn{1}{|c|}{\auc\%} &
  \fpr\% &
  \multicolumn{1}{|c|}{\auc\%} &
  \fpr\%\\ \midrule
   
    \underline{PGDi, FGSM, BIM} & \multicolumn{1}{|c|}{ } & \multicolumn{1}{c||}{ } & \multicolumn{1}{c|}{ } & \multicolumn{1}{c||}{ } & \multicolumn{1}{c|}{ } & \multicolumn{1}{c||}{ } & \multicolumn{1}{c|}{ } & \multicolumn{1}{c||}{ } & \multicolumn{1}{c|}{ } & \multicolumn{1}{c}{ } \\
    $\varepsilon$ = 0.03125 &
    \multicolumn{1}{|c|}{\textbf{55.0}} & \textbf{90.4} &
    \multicolumn{1}{|c|}{59.6} & 87.8 &
    \multicolumn{1}{|c|}{58.1} & \underline{88.7} &
    \multicolumn{1}{|c|}{58.9} & 85.6 &
    \multicolumn{1}{|c|}{\underline{57.0}} & 88.4
     \\
    $\varepsilon$ = 0.0625 &
    \multicolumn{1}{|c|}{\textbf{62.8}} & \textbf{83.5} &
    \multicolumn{1}{|c|}{68.4} & 76.0 &
    \multicolumn{1}{|c|}{67.4} & 75.9 &
    \multicolumn{1}{|c|}{\underline{64.3}} & \underline{80.5} &
    \multicolumn{1}{|c|}{67.2} & 76.0
     \\
    $\varepsilon$ = 0.25 &
    \multicolumn{1}{|c|}{\textbf{96.7}} & \textbf{17.9} &
    \multicolumn{1}{|c|}{98.6} & 6.7 &
    \multicolumn{1}{|c|}{98.7} & 5.7 &
    \multicolumn{1}{|c|}{\underline{\textbf{96.7}}} & \underline{17.4} &
    \multicolumn{1}{|c|}{99.2} & 3.4
     \\
    $\varepsilon$ = 0.5 &
    \multicolumn{1}{|c|}{\textbf{82.2}} & \textbf{60.0} &
    \multicolumn{1}{|c|}{91.9} & 37.7 &
    \multicolumn{1}{|c|}{91.9} & 37.1 &
    \multicolumn{1}{|c|}{\underline{90.2}} & \underline{43.9} &
    \multicolumn{1}{|c|}{93.0} & 34.4
     \\
  


 \underline{PGDi, FGSM, BIM, CWi, SA}  & \multicolumn{1}{|c|}{ } & \multicolumn{1}{c||}{ } & \multicolumn{1}{c|}{ } & \multicolumn{1}{c||}{ } & \multicolumn{1}{c|}{ } & \multicolumn{1}{c||}{ } & \multicolumn{1}{c|}{ } & \multicolumn{1}{c||}{ } & \multicolumn{1}{c|}{ } & \multicolumn{1}{c}{ }\\ 
$\varepsilon$ = 0.3125 &
\multicolumn{1}{|c|}{85.1} & 65.8 &
\multicolumn{1}{|c|}{85.7} & 64.8 &
\multicolumn{1}{|c|}{85.0} & 65.7 &
\multicolumn{1}{|c|}{\textbf{\underline{84.9}}} & \textbf{\underline{66.1}} &
\multicolumn{1}{|c|}{85.7} & 64.9
 \\ \hline

  L$_{\infty}$ Average & \multicolumn{1}{|c|}{\textbf{76.4}} & \textbf{63.5} &
    \multicolumn{1}{c|}{80.8} & 54.6 &
    \multicolumn{1}{c|}{80.2} & 54.6 &
    \multicolumn{1}{c|}{\underline{79.0}} & \underline{58.7} &
    \multicolumn{1}{c|}{80.4} & 58.2 
    \\ \hline
  \midrule
  
  \textbf{No norm} &
  \multicolumn{1}{|c|}{\auc\%} &
  \fpr\% &
  \multicolumn{1}{|c|}{\auc\%} &
  \fpr\% &
  \multicolumn{1}{|c|}{\auc\%} &
  \fpr\% &
  \multicolumn{1}{|c|}{\auc\%} &
  \fpr\% &
  \multicolumn{1}{|c|}{\auc\%} &
  \fpr\%\\ \midrule
  \underline{STA}  & \multicolumn{1}{|c|}{ } & \multicolumn{1}{c||}{ } & \multicolumn{1}{c|}{ } & \multicolumn{1}{c||}{ } & \multicolumn{1}{c|}{ } & \multicolumn{1}{c||}{ } & \multicolumn{1}{c|}{ } & \multicolumn{1}{c||}{ } & \multicolumn{1}{c|}{ } & \multicolumn{1}{c}{ }\\
No $\varepsilon$ & 
\multicolumn{1}{|c|}{\textbf{61.5}} & \textbf{85.9} &
\multicolumn{1}{|c|}{\underline{\textbf{61.5}}} & \underline{\textbf{85.9}} &
\multicolumn{1}{|c|}{\underline{\textbf{61.5}}} & \underline{\textbf{85.9}} &
\multicolumn{1}{|c|}{\underline{\textbf{61.5}}} & \underline{\textbf{85.9}} &
\multicolumn{1}{|c|}{\underline{\textbf{61.5}}} & \underline{\textbf{85.9}}
 \\ \hline
 No norm Average & 
 \multicolumn{1}{|c|}{\textbf{61.5}} & \textbf{85.9} &
\multicolumn{1}{|c|}{\underline{\textbf{61.5}}} & \underline{\textbf{85.9}} &
\multicolumn{1}{|c|}{\underline{\textbf{61.5}}} & \underline{\textbf{85.9}} &
\multicolumn{1}{|c|}{\underline{\textbf{61.5}}} & \underline{\textbf{85.9}} &
\multicolumn{1}{|c|}{\underline{\textbf{61.5}}} & \underline{\textbf{85.9}}
    \\ \hline
  \bottomrule
\end{tabular}
}
\caption{Performances on \method{FS} per objective and in \mead~on MNIST. The worst results among all the settings 
is in \textbf{bold}; the ones in the single-armed setting is \underline{underlined}. No norm denotes the group of attacks that do not depend on the norm constraint.}
\label{tab:mnist_fs_per_loss}
\end{table*}

Despite having trouble detecting attacks with a small maximal perturbation $\varepsilon$, \method{FS} detectors are not that bad at detecting adversarial examples. The attacks based on the Gini Impurity Score are the least detected ones among all the single-armed settings. The decrease in terms of \auc~is, in that case, 8 percentage points at most.
\newpage
\subsubsection{\method{MagNet}}\label{appendix:mnist-magnet}
In~\cref{tab:mnist_magnet_per_loss}, we show the result of our \mead~framework on MNIST, evaluated on \method{MagNet}.
\begin{table*}[!htb]
\centering
\resizebox{\columnwidth}{!}{%
\begin{tabular}{c||cc||cc||cc||cc||cc}
\toprule 
    \textbf{\method{MagNet}} &
  \multicolumn{2}{c||}{\mead} & \multicolumn{2}{c||}{ACE}& \multicolumn{2}{c||}{KL}& \multicolumn{2}{c||}{Gini} &
  \multicolumn{2}{c}{FR}
   \\ \hline\midrule
  \textbf{Norm L$_1$}&
  \multicolumn{1}{|c|}{\auc\%} &
   {\fpr\% } &
  \multicolumn{1}{c|}{\auc\%} &
   \fpr\% 
   &
  \multicolumn{1}{c|}{\auc\%} &
  \fpr\% &
  \multicolumn{1}{c|}{\auc\%} &
   \fpr\% &
  \multicolumn{1}{c|}{\auc\%} &
   \fpr\% \\ \midrule
   \underline{PGD1}  & \multicolumn{1}{|c|}{ } & \multicolumn{1}{c||}{ } & \multicolumn{1}{c|}{ } & \multicolumn{1}{c||}{ } & \multicolumn{1}{c|}{ } & \multicolumn{1}{c||}{ } & \multicolumn{1}{c|}{ } & \multicolumn{1}{c||}{ } & \multicolumn{1}{c|}{ } & \multicolumn{1}{c}{ } \\
   $\varepsilon$ = 5 &
\multicolumn{1}{|c|}{\textbf{87.6}} & \textbf{37.1} &
\multicolumn{1}{|c|}{88.5} & 35.1 &
\multicolumn{1}{|c|}{88.8} & \underline{36.8} &
\multicolumn{1}{|c|}{88.7} & 34.6 &
\multicolumn{1}{|c|}{\underline{87.7}} & 36.6
 \\
$\varepsilon$ = 10 &
\multicolumn{1}{|c|}{\textbf{99.2}} & \textbf{2.7} &
\multicolumn{1}{|c|}{99.3} & 2.3 &
\multicolumn{1}{|c|}{\textbf{\underline{99.2}}} & \underline{2.5} &
\multicolumn{1}{|c|}{99.4} & 1.9 &
\multicolumn{1}{|c|}{\textbf{\underline{99.2}}} & 2.3
 \\
$\varepsilon$ = 15 &
\multicolumn{1}{|c|}{\textbf{99.9}} & \textbf{0.2} &
\multicolumn{1}{|c|}{\textbf{\underline{99.9}}} & 0.1 &
\multicolumn{1}{|c|}{\textbf{\underline{99.9}}} & \textbf{\underline{0.2}} &
\multicolumn{1}{|c|}{100.0} & 0.1 &
\multicolumn{1}{|c|}{\textbf{\underline{99.9}}} & 0.1
 \\
$\varepsilon$ = 20 &
\multicolumn{1}{|c|}{\textbf{100.0}} & \textbf{0.0} &
\multicolumn{1}{|c|}{\textbf{\underline{100.0}}} & \textbf{\underline{0.0}} &
\multicolumn{1}{|c|}{\textbf{\underline{100.0}}} & \textbf{\underline{0.0}} & \multicolumn{1}{|c|}{\textbf{\underline{100.0}}} & \textbf{\underline{0.0}} &
\multicolumn{1}{|c|}{\textbf{\underline{100.0}}} & \textbf{\underline{0.0}} \\
$\varepsilon$ = 25 &
\multicolumn{1}{|c|}{\textbf{100.0}} & \textbf{0.0} &
\multicolumn{1}{|c|}{\textbf{\underline{100.0}}} & \textbf{\underline{0.0}} &
\multicolumn{1}{|c|}{\textbf{\underline{100.0}}} & \textbf{\underline{0.0}} & \multicolumn{1}{|c|}{\textbf{\underline{100.0}}} & \textbf{\underline{0.0}} &
\multicolumn{1}{|c|}{\textbf{\underline{100.0}}} & \textbf{\underline{0.0}} 
 \\
$\varepsilon$ = 30 &
\multicolumn{1}{|c|}{\textbf{100.0}} & \textbf{0.0} &
\multicolumn{1}{|c|}{\textbf{\underline{100.0}}} & \textbf{\underline{0.0}} &
\multicolumn{1}{|c|}{\textbf{\underline{100.0}}} & \textbf{\underline{0.0}} & \multicolumn{1}{|c|}{\textbf{\underline{100.0}}} & \textbf{\underline{0.0}} &
\multicolumn{1}{|c|}{\textbf{\underline{100.0}}} & \textbf{\underline{0.0}} 
 \\
$\varepsilon$ = 40 &
\multicolumn{1}{|c|}{\textbf{100.0}} & \textbf{0.0} &
\multicolumn{1}{|c|}{\textbf{\underline{100.0}}} & \textbf{\underline{0.0}} &
\multicolumn{1}{|c|}{\textbf{\underline{100.0}}} & \textbf{\underline{0.0}} & \multicolumn{1}{|c|}{\textbf{\underline{100.0}}} & \textbf{\underline{0.0}} &
\multicolumn{1}{|c|}{\textbf{\underline{100.0}}} & \textbf{\underline{0.0}} 
 \\
\hline
L$_{1}$ Average & \multicolumn{1}{|c|}{\textbf{98.1}} & 5.7 &
    \multicolumn{1}{c|}{98.2} & 5.4 &
    \multicolumn{1}{c|}{98.3} & \textbf{\underline{5.6}} &
    \multicolumn{1}{c|}{98.3} & 5.2  &
    \multicolumn{1}{c|}{\textbf{\underline{98.1}}} & \textbf{\underline{5.6}}
    \\ \hline
 \midrule
  
  
\textbf{Norm L$_2$} &
  \multicolumn{1}{|c|}{\auc\%} &
   \fpr\% &
  \multicolumn{1}{|c|}{\auc\%} &
   \fpr\% &
  \multicolumn{1}{|c|}{\auc\%} &
   \fpr\% &
  \multicolumn{1}{|c|}{\auc\%} &
   \fpr\% &
  \multicolumn{1}{|c|}{\auc\%} &
   \fpr\%\\ \midrule
   \underline{PGD2}  & \multicolumn{1}{|c|}{ } & \multicolumn{1}{c||}{ } & \multicolumn{1}{c|}{ } & \multicolumn{1}{c||}{ } & \multicolumn{1}{c|}{ } & \multicolumn{1}{c||}{ } & \multicolumn{1}{c|}{ } & \multicolumn{1}{c||}{ } & \multicolumn{1}{c|}{ } & \multicolumn{1}{c}{ } \\
   $\varepsilon$ = 0.125 &
\multicolumn{1}{|c|}{64.3} & 82.4 &
\multicolumn{1}{|c|}{65.6} & 80.4 &
\multicolumn{1}{|c|}{65.5} & 82.5 &
\multicolumn{1}{|c|}{65.7} & 80.1 &
\multicolumn{1}{|c|}{\textbf{\underline{63.0}}} & \textbf{\underline{84.0}}
 \\
$\varepsilon$ = 0.25 &
\multicolumn{1}{|c|}{74.4} & \textbf{68.3} &
\multicolumn{1}{|c|}{75.5} & 66.8 &
\multicolumn{1}{|c|}{76.6} & 66.4 &
\multicolumn{1}{|c|}{77.2} & \underline{68.0} &
\multicolumn{1}{|c|}{\textbf{\underline{73.4}}} & 67.0
 \\
$\varepsilon$ = 0.3125 &
\multicolumn{1}{|c|}{79.7} & 61.1 &
\multicolumn{1}{|c|}{81.0} & 56.7 &
\multicolumn{1}{|c|}{82.0} & 56.5 &
\multicolumn{1}{|c|}{81.9} & \textbf{\underline{61.3}} &
\multicolumn{1}{|c|}{\textbf{\underline{78.4}}} & 57.1
 \\
$\varepsilon$ = 0.5 &
\multicolumn{1}{|c|}{\textbf{90.8}} & \textbf{32.2} &
\multicolumn{1}{|c|}{91.9} & 30.8 &
\multicolumn{1}{|c|}{91.9} & 31.0 &
\multicolumn{1}{|c|}{91.7} & \textbf{\underline{32.2}} &
\multicolumn{1}{|c|}{\underline{91.0}} & 31.1
 \\
$\varepsilon$ = 1 &
\multicolumn{1}{|c|}{99.1} & 3.1 &
\multicolumn{1}{|c|}{99.6} & 1.6 &
\multicolumn{1}{|c|}{99.5} & 1.7 &
\multicolumn{1}{|c|}{\textbf{\underline{98.2}}} & \textbf{\underline{8.4}} &
\multicolumn{1}{|c|}{99.5} & 1.8
 \\
$\varepsilon$ = 1.5 &
\multicolumn{1}{|c|}{99.8} & 0.3 &
\multicolumn{1}{|c|}{100.0} & 0.1 &
\multicolumn{1}{|c|}{100.0} & 0.1 &
\multicolumn{1}{|c|}{\textbf{\underline{99.5}}} & \textbf{\underline{1.6}} &
\multicolumn{1}{|c|}{100.0} & 0.1
 \\
$\varepsilon$ = 2 &
\multicolumn{1}{|c|}{99.9} & 0.1 &
\multicolumn{1}{|c|}{100.0} & 0.0 &
\multicolumn{1}{|c|}{100.0} & 0.0 &
\multicolumn{1}{|c|}{\textbf{\underline{99.7}}} & \textbf{\underline{0.4}} &
\multicolumn{1}{|c|}{100.0} & 0.0
 \\
  
   \underline{DeepFool} & \multicolumn{1}{|c|}{ } & \multicolumn{1}{c||}{ } & \multicolumn{1}{c|}{ } & \multicolumn{1}{c||}{ } & \multicolumn{1}{c|}{ } & \multicolumn{1}{c||}{ } & \multicolumn{1}{c|}{ } & \multicolumn{1}{c||}{ } & \multicolumn{1}{c|}{ } & \multicolumn{1}{c}{ } \\
No $\varepsilon$  &
\multicolumn{1}{|c|}{\textbf{99.4}} & \textbf{1.1} &
\multicolumn{1}{|c|}{\textbf{\underline{99.4}}} & \textbf{\underline{1.1}} &
\multicolumn{1}{|c|}{\textbf{\underline{99.4}}} & \textbf{\underline{1.1}} &
\multicolumn{1}{|c|}{\textbf{\underline{99.4}}} & \textbf{\underline{1.1}} &
\multicolumn{1}{|c|}{\textbf{\underline{99.4}}} & \textbf{\underline{1.1}} 
 \\
 
   \underline{CW2}  & \multicolumn{1}{|c|}{ } & \multicolumn{1}{c||}{ } & \multicolumn{1}{c|}{ } & \multicolumn{1}{c||}{ } & \multicolumn{1}{c|}{ } & \multicolumn{1}{c||}{ } & \multicolumn{1}{c|}{ } & \multicolumn{1}{c||}{ } & \multicolumn{1}{c|}{ } & \multicolumn{1}{c}{ } \\
   $\varepsilon$ = 0.01 &
\multicolumn{1}{|c|}{\textbf{92.8}} & \textbf{38.3} &
\multicolumn{1}{|c|}{\textbf{\underline{92.8}}} & \textbf{\underline{38.3}} &
\multicolumn{1}{|c|}{\textbf{\underline{92.8}}} & \textbf{\underline{38.3}} &
\multicolumn{1}{|c|}{\textbf{\underline{92.8}}} & \textbf{\underline{38.3}} &
\multicolumn{1}{|c|}{\textbf{\underline{92.8}}} & \textbf{\underline{38.3}}  \\

   \underline{HOP} & \multicolumn{1}{|c|}{ } & \multicolumn{1}{c||}{ } & \multicolumn{1}{c|}{ } & \multicolumn{1}{c||}{ } & \multicolumn{1}{c|}{ } & \multicolumn{1}{c||}{ } & \multicolumn{1}{c|}{ } & \multicolumn{1}{c||}{ } & \multicolumn{1}{c|}{ } & \multicolumn{1}{c}{ } \\
$\varepsilon$ = 0.1 &
\multicolumn{1}{|c|}{\textbf{99.9}} & \textbf{0.0} &
\multicolumn{1}{|c|}{\textbf{\underline{99.9}}} & \textbf{\underline{0.0}} &
\multicolumn{1}{|c|}{\textbf{\underline{99.9}}} & \textbf{\underline{0.0}} &
\multicolumn{1}{|c|}{\textbf{\underline{99.9}}} & \textbf{\underline{0.0}} &
\multicolumn{1}{|c|}{\textbf{\underline{99.9}}} & \textbf{\underline{0.0}} 
 \\
\hline
L$_{2}$ Average & \multicolumn{1}{|c|}{90.0} & 28.7 &
    \multicolumn{1}{c|}{90.6} & 27.6 &
    \multicolumn{1}{c|}{90.8} & 27.8 &
    \multicolumn{1}{c|}{90.6} & \textbf{\underline{29.1}} &
    \multicolumn{1}{c|}{\textbf{\underline{89.7}}} & 28.1
    \\ \hline\midrule
  

\textbf{Norm L$_\infty$}  &
  \multicolumn{1}{|c|}{\auc\%} &
   \fpr\% &
  \multicolumn{1}{|c|}{\auc\%} &
   \fpr\% &
  \multicolumn{1}{|c|}{\auc\%} &
   \fpr\% &
  \multicolumn{1}{|c|}{\auc\%} &
   \fpr\% &
  \multicolumn{1}{|c|}{\auc\%} &
   \fpr\%\\ \midrule
   
   \underline{PGDi, FGSM, BIM}  & \multicolumn{1}{|c|}{ } & \multicolumn{1}{c||}{ } & \multicolumn{1}{c|}{ } & \multicolumn{1}{c||}{ } & \multicolumn{1}{c|}{ } & \multicolumn{1}{c||}{ } & \multicolumn{1}{c|}{ } & \multicolumn{1}{c||}{ } & \multicolumn{1}{c|}{ } & \multicolumn{1}{c}{ } \\
   $\varepsilon$ = 0.03125 &
\multicolumn{1}{|c|}{\textbf{100.0}} & \textbf{0.1} &
\multicolumn{1}{|c|}{\textbf{\underline{100.0}}} & 0.0 &
\multicolumn{1}{|c|}{\textbf{\underline{100.0}}} & 0.0 &
\multicolumn{1}{|c|}{\textbf{\underline{100.0}}} & \textbf{\underline{0.1}} &
\multicolumn{1}{|c|}{\textbf{\underline{100.0}}} & 0.0
 \\
$\varepsilon$ = 0.0625 &
\multicolumn{1}{|c|}{\textbf{100.0}} & \textbf{0.0} &
\multicolumn{1}{|c|}{\textbf{\underline{100.0}}} & \textbf{\underline{0.0}} &
\multicolumn{1}{|c|}{\textbf{\underline{100.0}}} & \textbf{\underline{0.0}} & \multicolumn{1}{|c|}{\textbf{\underline{100.0}}} & \textbf{\underline{0.0}} &
\multicolumn{1}{|c|}{\textbf{\underline{100.0}}} & \textbf{\underline{0.0}}
 \\
$\varepsilon$ = 0.25 &
\multicolumn{1}{|c|}{\textbf{100.0}} & \textbf{0.0} &
\multicolumn{1}{|c|}{\textbf{\underline{100.0}}} & \textbf{\underline{0.0}} &
\multicolumn{1}{|c|}{\textbf{\underline{100.0}}} & \textbf{\underline{0.0}} & \multicolumn{1}{|c|}{\textbf{\underline{100.0}}} & \textbf{\underline{0.0}} &
\multicolumn{1}{|c|}{\textbf{\underline{100.0}}} & \textbf{\underline{0.0}}
 \\
$\varepsilon$ = 0.5 &
\multicolumn{1}{|c|}{\textbf{100.0}} & \textbf{0.0} &
\multicolumn{1}{|c|}{\textbf{\underline{100.0}}} & \textbf{\underline{0.0}} &
\multicolumn{1}{|c|}{\textbf{\underline{100.0}}} & \textbf{\underline{0.0}} & \multicolumn{1}{|c|}{\textbf{\underline{100.0}}} & \textbf{\underline{0.0}} &
\multicolumn{1}{|c|}{\textbf{\underline{100.0}}} & \textbf{\underline{0.0}}
 \\
    
  
  
  
 \underline{PGDi, FGSM, BIM, CWi, SA}   & \multicolumn{1}{|c|}{ } & \multicolumn{1}{c||}{ } & \multicolumn{1}{c|}{ } & \multicolumn{1}{c||}{ } & \multicolumn{1}{c|}{ } & \multicolumn{1}{c||}{ } & \multicolumn{1}{c|}{ } & \multicolumn{1}{c||}{ } & \multicolumn{1}{c|}{ } & \multicolumn{1}{c}{ } \\
 $\varepsilon$ = 0.3125 &
\multicolumn{1}{|c|}{92.6} & 51.4 &
\multicolumn{1}{|c|}{92.6} & 51.4 &
\multicolumn{1}{|c|}{\textbf{\underline{92.2}}} & \textbf{\underline{53.1}} &
\multicolumn{1}{|c|}{92.3} & 52.5 &
\multicolumn{1}{|c|}{92.5} & 51.8
 \\
  \hline
  L$_{\infty}$ Average & \multicolumn{1}{|c|}{98.5} & 10.3 &
    \multicolumn{1}{c|}{98.5} & 10.3 &
    \multicolumn{1}{c|}{\textbf{\underline{98.4}}} & \textbf{\underline{10.6}} &
    \multicolumn{1}{c|}{98.5} & 10.5 &
    \multicolumn{1}{c|}{98.5} & 10.4
    \\ \hline
  \midrule
  
\textbf{No norm} &
  \multicolumn{1}{|c|}{\auc\%} &
   \fpr\% &
  \multicolumn{1}{|c|}{\auc\%} &
   \fpr\% &
  \multicolumn{1}{|c|}{\auc\%} &
   \fpr\% &
  \multicolumn{1}{|c|}{\auc\%} &
   \fpr\% &
  \multicolumn{1}{|c|}{\auc\%} &
   \fpr\%\\ \midrule
   
   \underline{STA}  & \multicolumn{1}{|c|}{ } & \multicolumn{1}{c||}{ } & \multicolumn{1}{c|}{ } & \multicolumn{1}{c||}{ } & \multicolumn{1}{c|}{ } & \multicolumn{1}{c||}{ } & \multicolumn{1}{c|}{ } & \multicolumn{1}{c||}{ } & \multicolumn{1}{c|}{ } & \multicolumn{1}{c}{ } \\
No $\varepsilon$ & 
\multicolumn{1}{|c|}{\textbf{86.9}} & \textbf{74.3} &
\multicolumn{1}{|c|}{\underline{\textbf{86.9}}} & \underline{\textbf{74.3}} &
\multicolumn{1}{|c|}{\underline{\textbf{86.9}}} & \underline{\textbf{74.3}} &
\multicolumn{1}{|c|}{\underline{\textbf{86.9}}} & \underline{\textbf{74.3}}&
\multicolumn{1}{|c|}{\underline{\textbf{86.9}}} & \underline{\textbf{74.3}}
 \\
   \hline
   No norm Average & 
\multicolumn{1}{|c|}{\textbf{86.9}} & \textbf{74.3} &
\multicolumn{1}{|c|}{\underline{\textbf{86.9}}} & \underline{\textbf{74.3}} &
\multicolumn{1}{|c|}{\underline{\textbf{86.9}}} & \underline{\textbf{74.3}} &
\multicolumn{1}{|c|}{\underline{\textbf{86.9}}} & \underline{\textbf{74.3}}&
\multicolumn{1}{|c|}{\underline{\textbf{86.9}}} & \underline{\textbf{74.3}}
    \\ \hline
   \bottomrule
\end{tabular}
}
\caption{Performances on \method{MagNet} per objective and in \mead~on MNIST. The worst results among all the settings 
are shown in \textbf{bold}; the ones in the single-armed setting is \underline{underlined}. No norm denotes the group of attacks that do not depend on the norm constraint.}
\label{tab:mnist_magnet_per_loss}
\end{table*}

\method{MagNet} is effective on MNIST. It is pretty close to the perfect detector for L$_{\infty}$ and L$_{1}$ attacks. Anyway, the Fisher-Rao-based attacks are the most disruptive ones for such detectors. Similar to the CIFAR10 case, the results using \mead~are quite close to the worst single-armed setting case. The decrease in \auc~between \mead~and the worst single-armed setting is at most 0.5 percentage points.

\end{document}